\documentclass{article}

\PassOptionsToPackage{numbers, compress}{natbib}
\usepackage[main, final, nonatbib]{neurips_2026}

\usepackage[utf8]{inputenc} 
\usepackage[T1]{fontenc}    
\usepackage{hyperref}       
\usepackage{url}            
\usepackage{booktabs}       
\usepackage{amsfonts}       
\usepackage{nicefrac}       
\usepackage{xcolor}         
\usepackage{lipsum} 
\usepackage{enumitem} 
\usepackage{amsmath}
\usepackage{graphicx}
\usepackage{amssymb}
\usepackage{algorithm}
\usepackage{algorithmic}
\usepackage{amsthm}
\usepackage{multirow}
\usepackage{makecell}
\usepackage{wrapfig}
\usepackage{microtype}      
\usepackage{mathtools}
\usepackage{tcolorbox}
\usepackage{bm}

\usepackage[normalem]{ulem}
\usepackage{xcolor}

\usepackage[table]{xcolor}
\definecolor{monaPurple}{RGB}{231,230,254}
\definecolor{monaBlue}{RGB}{224,240,255}

\newtheorem{proposition}{Proposition}
\newtheorem{definition}{Definition}
\newtheorem{corollary}{Corollary}
\newtheorem{theorem}{Theorem}

\usepackage{tcolorbox}

\usepackage{xcolor}
\colorlet{lightblue}{blue!75}

\hypersetup{
    colorlinks=true,
    citecolor=lightblue
}

\definecolor{propbg}{RGB}{247,246,252}
\definecolor{propborder}{RGB}{88,76,168}

\newtcolorbox{propositionbox}{
  colback=propbg,
  colframe=propborder,
  boxrule=1.2pt,
  arc=2.5mm,
  left=0.5mm,
  right=0.5mm,
  top=0.5mm,
  bottom=0.5mm,
}

\definecolor{propbg2}{RGB}{248,254,252}
\definecolor{propborder2}{RGB}{76,136,120}

\newtcolorbox{theorembox}{
  colback=propbg2,
  colframe=propborder2,
  boxrule=1.2pt,
  arc=2.5mm,
  left=0.5mm,
  right=0.5mm,
  top=0.5mm,
  bottom=0.5mm,
}

\usepackage{tabularx}

\usepackage[numbers,sort&compress]{natbib}

\title{\raisebox{-0.05em}{\includegraphics[height=0.76em]{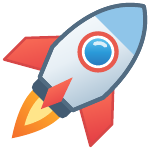}\hspace{0.15em}}NADS: Navigator-Guided Data Selection for Mitigating Catastrophic Forgetting in Fine-Tuning}

\author{
  \textbf{Jianhao Zhang}\textsuperscript{1} \quad
    \textbf{Ou Wu}\textsuperscript{1} \quad
    \textbf{Yi Du}\textsuperscript{1,2,*}
    \\[0.05in]
    \textsuperscript{1}Hangzhou Institute for Advanced Study, \\
    University of Chinese Academy of Sciences, Hangzhou, China,
    \\
    \textsuperscript{2}Computer Network Information Center,
    Chinese Academy of Sciences, Beijing, China,
    \\
    \texttt{zhangjianhao24@mails.ucas.ac.cn}, \quad
    \texttt{wuou@ucas.ac.cn}, \quad
    \texttt{duyi@cnic.cn}
}

\begin{document}

\maketitle
\renewcommand{\thefootnote}{\fnsymbol{footnote}}
\footnotetext[1]{Corresponding author.}

\begin{abstract}
Catastrophic forgetting is a central obstacle in single-stage fine-tuning of large language models (LLMs) on a new task, especially when the original pretraining data are unavailable. Synthetic data offers a practical mitigation path, but candidate pools are often noisy and redundant, making effective selection essential. Existing methods often rely on indirect proxy signals whose alignment with actual training benefit can be limited, while utility--diversity selection over large candidate pools remains computationally demanding. To address these challenges, we propose \textbf{NA}vigator-guided \textbf{D}ata \textbf{S}election (\textbf{NADS}), a fine-tuning framework for LLMs. NADS first follows the target fine-tuning recipe without preservation regularization to obtain a navigator model, whose deviation from the pretrained model exposes recipe-induced capability drift. It then uses the predictive divergence between the navigator and the pretrained model to score forgetting-aware utility for each synthetic candidate, and constructs a constraint set through an efficient utility--diversity selection strategy. Finally, NADS fine-tunes on the new-task data while distilling pretrained behavior on the selected constraints to preserve general capabilities. Experiments show that NADS delivers a stronger balance between new-task performance and general capability preservation, while reducing utility--diversity selection overhead. Code at: \url{https://github.com/zjh-vinky/NADS}.
\end{abstract}

\section{Introduction}
Large language models (LLMs) have achieved remarkable performance across a wide range of natural language processing tasks and are increasingly deployed in real-world applications~\citep{shi2025continual,guan2025survey}. As practical demands evolve, these models are often fine-tuned on new target data to adapt to emerging tasks~\citep{ke2025adaptation, lu2025fine,hui2025hft}. However, recent studies indicate that adapting to new data can also erode previously acquired capabilities~\citep{bethunescaling,hacohenpredicting}, even after a single fine-tuning stage, resulting in catastrophic forgetting~\citep{wu2025mitigating, wang2025on, lin2026sft, sanyalupweighting, shen2025dont}.

Recent advances have explored diverse strategies for mitigating catastrophic forgetting during fine-tuning. Rehearsal-based methods revisit historical samples to preserve prior capabilities~\citep{dengunlocking,abbes2025revisiting}, while representation-based methods mitigate forgetting through representational constraints~\citep{ye2024distilling,he2024seekr}. However, these methods generally assume access to pretraining data, which is often unavailable in realistic deployment settings. This limitation has motivated alternative approaches that do not rely on prior data, such as model merging~\citep{yang2025continual}, selective parameter updates~\citep{bidermanlora}, and importance weighting~\citep{sanyalupweighting}. Recently, work has begun to leverage synthetic data for forgetting mitigation~\citep{huang2024mitigating,wangself}.

Synthetic data offers a promising avenue for mitigating forgetting, yet the resulting candidate pools are often noisy and highly redundant, making effective data selection crucial~\citep{wangself,ding2025improved}. Many existing methods typically rely on indirect proxy signals such as perplexity to estimate sample utility, but these signals are not always well aligned with the actual training benefit~\citep{liu2025take,ivison2025larges,wang2025nice}. Consequently, selected samples may appear plausible while contributing limited practical value~\citep{yu2024mates,sam2025analyzing,cai2025low}. To alleviate redundancy, prior work often incorporates diversity-promoting strategies~\citep{li2025mdit,kessy2018optimal}. However, such strategies are often decoupled from utility estimation, making it difficult to jointly optimize sample quality and diversity. While determinantal point processes~(DPPs)~\citep{kulesza2012determinantal} provide an effective approach, their direct optimization is often computationally expensive for large-scale candidate pools, requiring $O(KNd)~$\citep{chen2018fast} or $O(KN(d+\log N))$~\citep{hemmi2022lazy} to select $K$ samples from a pool of size $N$. These limitations leave an unresolved challenge of how to perform training-aware and diversity-aware selection efficiently over large-scale synthetic candidate pools.

To tackle these challenges, we propose \textbf{NA}vigator-guided \textbf{D}ata \textbf{S}election~(\textbf{NADS}), a novel fine-tuning framework for LLMs. NADS first follows the target fine-tuning recipe without preservation regularization to obtain a navigator model, whose deviation from the pretrained model deliberately exposes recipe-induced drift in pretrained behavior. This drift is then converted into a forgetting-aware utility for selecting synthetic samples as preservation constraints. NADS further combines this utility with diversity-aware subset construction through a scalable optimization procedure with iterative cost $O(TN(E+\log K))$. Finally, NADS fine-tunes on the new-task data while distilling pretrained behavior on the selected constraint set through knowledge distillation (KD), preserving general capabilities during task adaptation. 

Our key contributions are summarized as follows:

\begin{itemize}[leftmargin=*]
    \item We propose NADS, a novel fine-tuning framework for LLMs that uses recipe-conditioned navigator drift to guide data selection for forgetting mitigation. NADS trains a navigator under the target recipe without preservation regularization to deliberately expose drift in pretrained capabilities and then uses this drift to estimate the forgetting-aware utility of candidate samples.

    \item We formulate constraint set construction as a utility--diversity selection problem, and propose an efficient optimization strategy that renders this otherwise costly DPP-based formulation practical at scale. The resulting method reduces utility--diversity selection cost while preserving sample quality and diversity. We further provide theoretical analysis that motivates the proposed framework.

    \item We conduct extensive experiments across two benchmarks and three model families spanning 3B--9B parameters. Results show that NADS delivers a stronger balance between new-task adaptation and general capability preservation, while achieving this with lower utility--diversity selection cost.
\end{itemize}

\section{Related Work}
\subsection{Mitigating Catastrophic Forgetting in LLM Fine-Tuning}
Fine-tuning is the dominant paradigm for adapting large language models to evolving tasks and feedback, yet it often remains difficult to acquire new capabilities without degrading existing ones~\citep{wang2024comprehensive,li2024revisiting,wang2025see}. To mitigate forgetting, various approaches have been proposed. Regularization-based methods augment the training objective with importance-weighted penalties~\citep{wu2024meta,wang2024model,zhao2024a,sarkar2025adapt, chen2026algorithmgoalsdualscoring, qing2026alphatokendecouplingadaptationstability}. Rehearsal-based methods preserve prior knowledge by interleaving stored or synthesized samples during subsequent training~\citep{yin2022contintin,huang2024mitigating,wangself,borsos2020coresets,maekawa2023generative}. Architecture-based methods reduce interference by isolating task-specific capacity, from expandable or modular designs to parameter-efficient tuning that restricts updates to lightweight components~\citep{wang2023incorporating,zhao2024sapt,ren2024analy,lu2025controlled}. Sample-space methods instead modulate the effective training signal via example reweighting, offering a lightweight way to balance stability and plasticity~\citep{sanyalupweighting,chen2025vul,yang2025data,sow2025dynamic}. While some replay methods use post-fine-tuning model-behavior changes to identify forgotten samples~\citep{bai2025efficient}, NADS uses an unconstrained navigator before final constrained training to estimate forgetting-aware utility, enabling proactive synthetic-constraint selection from large candidate pools.

\subsection{Data Selection in LLM Fine-Tuning}
Data selection aims to filter training data that meet desired quality requirements, improving efficiency while maintaining comparable or near-lossless performance. Existing methods often do so using indirect proxy signals (e.g., reliability, consistency, or difficulty scores) to estimate sample usefulness for target adaptation~\citep{liu2024selectit,li2024selective,li2025scar,shen2024rethink,li2024quantity,kang2024get}. Another line of work uses optimization-related proxy signals (e.g., gradients or influence scores)~\citep{xia2024less,nikdan2025efficient,chen2026influencepreserving}. Recently, cross-model discrepancies have been explored for data selection~\citep{zhou2025davir,li2026tokenlevel,taheri2025forgettingnewmechanismbetter}. Additionally, many methods encourage coverage to prevent the selected subset from collapsing into near duplicates, either through diversity diagnostics or explicit quality-diversity trade-offs~\citep{bukharin2024data,yang2025measuring,chen2025mig,zhang2025d3diversity}, or through structured objectives that directly favor non-overlapping evidence~\citep{li2025selection,zhao2025beyond,hoang2026difficultydiversity}. Although these methods improve data selection for fine-tuning, their selection objectives are mostly designed to improve target-task performance or reduce fine-tuning cost, rather than to select preservation constraints against update-induced forgetting.

\begin{figure}[t]
    \centering
    \includegraphics[width=\linewidth]{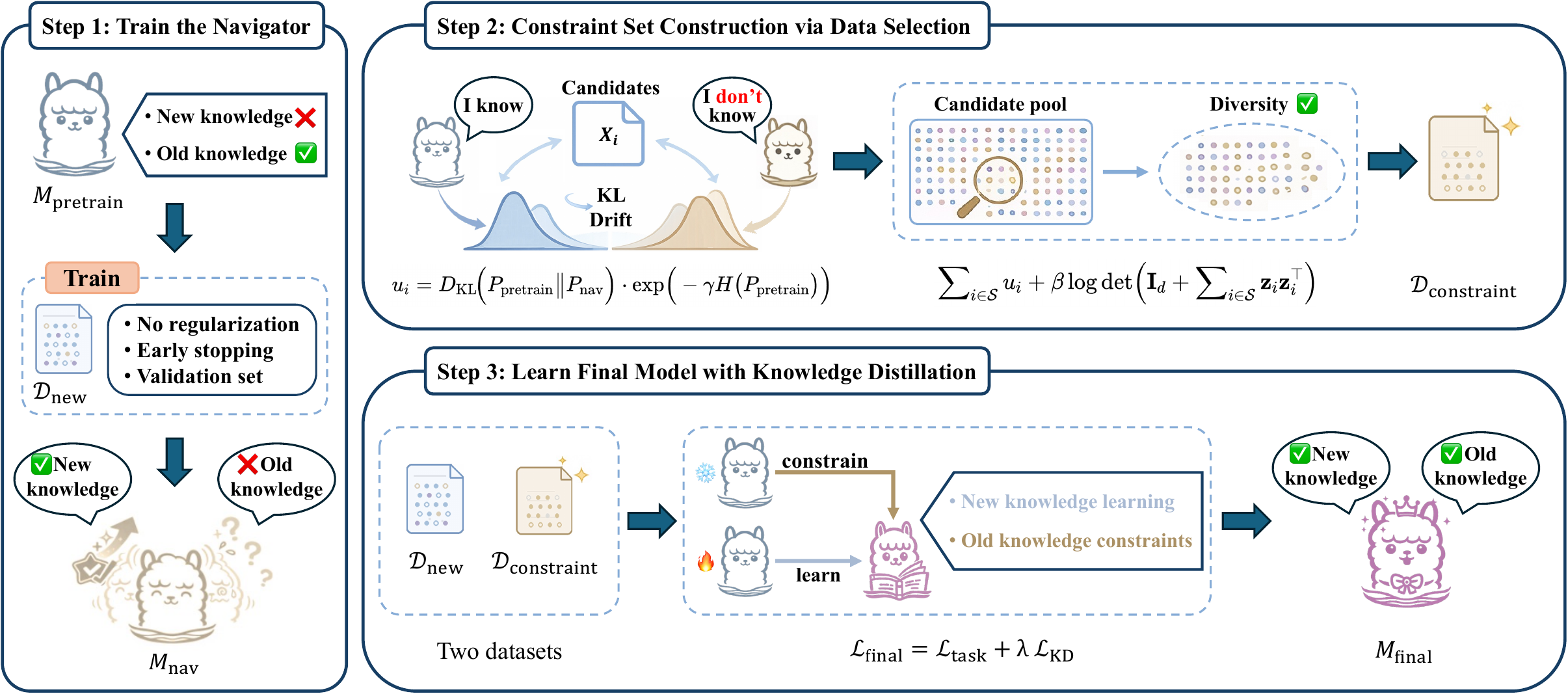}\vspace{-0.1in}
    \caption{The overall framework of NADS. First, we train a navigator $M_{\text{nav}}$ from $M_{\text{pretrain}}$ on $\mathcal{D}_{\text{new}}$ without regularization. Next, we score sample utility from the predictive divergence between $M_{\text{nav}}$ and $M_{\text{pretrain}}$, and construct $\mathcal{D}_{\text{constraint}}$ via utility--diversity selection with DPPs. Finally, we train $M_{\text{final}}$ on $\mathcal{D}_{\text{new}}$ with distillation on $\mathcal{D}_{\text{constraint}}$ to preserve general capabilities during new-task adaptation.}
    \label{framework}
\end{figure}

\section{Methodology}

\subsection{Problem Definition}

Let $\bm{\bm{\theta}}$ denote the model parameters. We consider a single-stage fine-tuning setting with a new-task dataset $\mathcal{D}_{\text{new}}=\{(x_i,y_i)\}_{i=1}^{M}$, where the original pretraining data is unavailable but an auxiliary synthetic dataset $\mathcal{C}=\{x_j\}_{j=1}^{N}$ is accessible. NADS first constructs a navigator model $M_{\text{nav}}$ from the pretrained model $M_{\text{pretrain}}$ to select a constraint set $\mathcal{D}_{\text{constraint}}\subset\mathcal{C}$. It then trains $M_{\text{final}}$ on $\mathcal{D}_{\text{new}}$ with KD regularization on $\mathcal{D}_{\text{constraint}}$, improving new-task performance while preserving the capabilities of $M_{\text{pretrain}}$. The overall framework is illustrated in Fig.~\ref{framework}.

\subsection{Navigator Model Construction}
Existing studies~\citep{zhou2025davir,jain2026train,li2026tokenlevel,taheri2025forgettingnewmechanismbetter} have explored model-centric data selection that measures the learning value of data through model change before and after training, thereby linking sample evaluation to its actual benefit for task adaptation. However, they primarily use the model to facilitate new-task adaptation, without exploring its potential to reveal update-induced forgetting during fine-tuning. Further discussion of the distinctions between these methods and NADS in Appendix~\ref{appendix_method_discussion}.

Motivated by this observation, we construct a model to trace the unconstrained adaptation path under the target recipe. This model is fine-tuned from $M_{\text{pretrain}}$ on $\mathcal{D}_{\text{new}}$ without extra regularization, and therefore follows the realized training trajectory while naturally accumulating forgetting-related drift. Because it reveals how pretrained behavior drifts along this path and provides the guidance signal for subsequent forgetting-aware utility estimation, we refer to it as the \textit{navigator model} $M_{\text{nav}}$.

\textbf{Checkpoint Selection and Local Deviation of the Navigator.}
The navigator serves as a reference along the unconstrained training trajectory. We select its checkpoint by validation-based early stopping, for which the following proposition gives a standard guarantee.

\begin{propositionbox}
\begin{proposition}[Validation selection as a near-best trajectory reference]
\label{prop:nav_near_opt}
Let $\{\bm{\theta}_t\}_{t=0}^T$ denote the unconstrained adaptation trajectory induced by the target recipe, and define
$R(t)\coloneqq \mathcal{L}_{\text{pop}}(\bm{\theta}_t)$ and
$\hat R(t)\coloneqq \mathcal{L}_{\text{val}}(\bm{\theta}_t)$.
Let
$t_{\text{opt}}\in \arg\min_{t\in\{0,\dots,T\}} R(t)$ and
$t^*\in \arg\min_{t\in\{0,\dots,T\}} \hat R(t)$.
Assume that there exists $\varepsilon>0$ such that
$|\hat R(t)-R(t)|\le \varepsilon$ for all $t\in\{0,\dots,T\}$.
Then
$R(t^*)-R(t_{\text{opt}})\le 2\varepsilon$.
Therefore, the validation-selected checkpoint is a near-best target-adapted reference state on the unconstrained trajectory in population new-task risk. The proof is given in Appendix~\ref{Proposition_1}.
\end{proposition}
\end{propositionbox}
Because $M_{\text{nav}}$ is trained without constraint data or anti-forgetting regularization, it follows the unconstrained adaptation path and exhibits a more aggressive drift away from $M_{\text{pretrain}}$. The following proposition locally controls deviation from this reference under bounded updates.

\begin{propositionbox}
\begin{proposition}[Local deviation recursion around the navigator trajectory]
\label{prop:anchoring_navigator}
Let $\{\bar{\bm{\theta}}_t\}_{t=0}^T$ denote the unconstrained trajectory, and let $\{\bm{\theta}_t\}_{t=0}^T$ denote the trajectory of $M_{\text{final}}$ satisfying $\bm{\theta}_{t+1}=\bm{\theta}_t-\eta\bigl(\nabla L_{\text{task}}(\bm{\theta}_t)+\xi_t\bigr)$, where $\xi_t$ is the preservation-gradient component (Section~\ref{kd}) and $\|\xi_t\|\le \kappa$. Assume that $\nabla L_{\text{task}}$ is $L$-Lipschitz. Then, for any step $t$, $\|\bm{\theta}_{t+1}-\bar{\bm{\theta}}_{t+1}\|\le (1+\eta L)\|\bm{\theta}_t-\bar{\bm{\theta}}_t\|+\eta\kappa$. Consequently, for any local horizon $\tau\ge 0$, $\sup_{0\le s\le \tau}\|\bm{\theta}_{t+s}-\bar{\bm{\theta}}_{t+s}\|\le \|\bm{\theta}_t-\bar{\bm{\theta}}_t\|+\sum_{j=0}^{\tau-1}\eta\bigl(L\|\bm{\theta}_{t+j}-\bar{\bm{\theta}}_{t+j}\|+\kappa\bigr)$. Thus, bounded preservation updates locally control trajectory mismatch. The proof is given in Appendix~\ref{Proposition_2}.
\end{proposition}
\end{propositionbox}

Appendix~\ref{app:corollary_ranking_stability} gives a conditional local perturbation view of navigator-side forgetting-risk ordering.

Concretely, we obtain $M_{\text{nav}}$ by fine-tuning $M_{\text{pretrain}}$ on $\mathcal{D}_{\text{new}}$ with validation-based early stopping. When validation data are unavailable or unreliable (e.g., data scarcity or label noise), we can use validation-free overfitting detection methods, such as the evolution of weight-matrix spectra~\citep{he2025earlystopp}. We further analyze the performance and stability of $M_{\text{nav}}$ through experiments in Section~\ref{sec:navigator_analysis}.

\subsection{Constraint Set Construction via Data Selection}
When constructing $\mathcal{D}_{\text{constraint}}$, we aim to select samples that expose forgetting-related predictive drift under the target fine-tuning recipe while avoiding redundant selections. Since utility-only scoring can favor highly similar samples and sacrifice coverage, we formulate subset selection on the synthetic dataset $\mathcal{C}$ as a $k$-DPP~\citep{kulesza2011k} to jointly account for utility and diversity under a fixed budget $K$.

\begin{definition}[$k$-DPP]
Let $\mathcal{C}=\{x_i\}_{i=1}^{N}$ indexed by $\mathcal{Y}=\{1,\ldots,N\}$, and let $ S\subseteq\mathcal{Y}$ denote a selected set. For a real symmetric positive semidefinite kernel $\mathbf{L}\in\mathbb{R}^{N\times N}$, the $k$-DPP is defined by
\begin{equation}
\mathbb{P}( S;\mathbf{L}) \propto \det(\mathbf{L}_{ S}),\qquad | S|=K,
\label{eq:kdpp_def}
\end{equation}
where $\mathbf{L}_{ S}$ is the principal submatrix of $\mathbf{L}$ indexed by $ S$. 
\end{definition}
Following~\citep{kulesza2012determinantal}, we adopt the standard $L$-ensemble decomposition: $(\mathbf{L})_{ij}=q_i\,G_{ij}\,q_j$, where $q_i\in\mathbb{R}_+$ is the quality score of sample $x_i$ and $\mathbf{G}\succeq 0$ is a similarity kernel capturing pairwise diversity.

\begin{figure}[t]
    \centering
    \includegraphics[width=\linewidth]{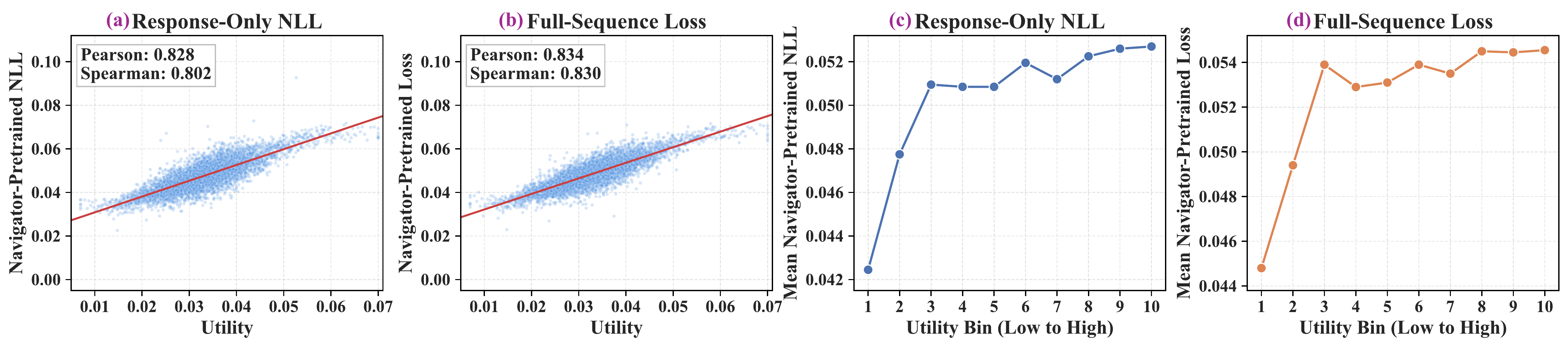}\vspace{-0.2in}
    \caption{(a--b) Utility versus response-only NLL degradation and full-sequence loss increase, with Pearson and Spearman measuring sample correlation and ranking consistency, respectively. (c--d) Mean response-only NLL degradation and full-sequence loss increase across utility deciles. Higher-utility samples exhibit larger increases, indicating strong correlation with samples' forgetting risk.}
    \label{fig2}
\end{figure}

\textbf{Forgetting Utility.} 
$M_{\text{nav}}$ reflects update-induced deviation from $M_{\text{pretrain}}$ during task adaptation. We therefore use the forward KL divergence from $M_{\text{pretrain}}$ to $M_{\text{nav}}$ as a forgetting-aware utility signal. Larger divergence then indicates higher preservation priority for samples that deviate more from pretrained behavior. However, not all large divergences are equally informative, as some may reflect noise or intrinsic uncertainty rather than forgetting-relevant changes. To suppress such spurious utility, we further apply a confidence gate derived from $M_{\text{pretrain}}$. The resulting utility is defined as
\begin{equation}
u_i = \frac{1}{|x_i|}\sum\nolimits_{t=1}^{|x_i|}\bigg[D_{\text{KL}}\!\Big(P_{\text{pretrain}}(\cdot \mid x_{i,<t})\,\big\|\,P_{\text{nav}}(\cdot \mid x_{i,<t})\Big)\cdot\exp\!\Big(-\gamma\, H\!\big(P_{\text{pretrain}}(\cdot\mid x_{i,<t})\big)\Big)\bigg],
\label{utility}
\end{equation}
where $x_i$ denotes the full sequence, $H(p)$ denotes entropy, and $\gamma$ controls the confidence gate. As empirically supported by Fig.~\ref{fig2}, higher-utility samples exhibit higher forgetting risk. To incorporate this utility into the $k$-DPP formulation, we define $q_i \triangleq \exp \left(u_i/(2\beta)\right)$, where $\beta$ controls the scaling. This yields an additive objective in $u_i$, linking utility estimation to subset selection. Appendix~\ref{app:utility_analysis} provides a local utility--KD alignment rationale, and Appendix~\ref{app:utility_constraint_effectiveness} shows that higher-utility candidates form more effective preservation constraints under the same KD objective.

\textbf{Diversity Kernel.}
DPPs capture diversity through the determinant of the similarity submatrix $\mathbf{G}_{ S}$, thereby favoring subsets whose elements are mutually dissimilar under $\mathbf{G}$. To operationalize this property, we construct $\mathbf{G}$ as an inner-product Gram kernel~\citep{belhadji2020determinantal} with $G_{ij}\triangleq \mathbf{z}_i^{\top}\mathbf{z}_j$ and $\mathbf{z}_i\in\mathbb{R}^{d}$, where $d$ is the embedding dimension, and adopt the stabilized diversity term $\log\det(\mathbf{I}_K+\mathbf{G}_{ S})$~\citep{chen2026posttraining}, where $\mathbf{G}_{ S}$ is the principal submatrix indexed by $ S$. Let $\mathbf{Z}_{ S}=[\mathbf{z}_i]_{i\in  S}\in\mathbb{R}^{d\times K}$ so that $\mathbf{G}_{ S}=\mathbf{Z}_{ S}^\top \mathbf{Z}_{ S}$. Since this involves a $K\times K$ determinant, we apply Sylvester's determinant identity~\citep{horn2012matrix} to rewrite it as
\begin{equation}
\det(\mathbf{I}_K+\mathbf{Z}_{ S}^\top \mathbf{Z}_{ S})
=
\det(\mathbf{I}_d+\mathbf{Z}_{ S} \mathbf{Z}_{ S}^\top)
=
\det\!\Big(\mathbf{I}_d+\sum\nolimits_{i\in  S}\mathbf{z}_i\mathbf{z}_i^\top\Big).
\end{equation}

\begin{theorembox}
\begin{theorem}[Utility--diversity form of the stabilized $k$-DPP objective]
\label{thm:kdpp_objective}
Let the $L$-ensemble kernel satisfy $(\mathbf{L})_{ij}=q_i G_{ij} q_j$, where
$q_i=\exp(u_i/(2\beta))$ and $G_{ij}=\mathbf{z}_i^\top \mathbf{z}_j$. Then, under the cardinality constraint $| S|=K$, maximizing the stabilized $k$-DPP objective is equivalent to solving
\begin{equation}
 S^* =
\operatornamewithlimits{arg\,max}_{ S\subseteq\mathcal{Y},\,| S|=K}\
\sum_{i\in S} u_i
\;+\;
\beta \log\det\!\bigg(\mathbf{I}_d+\sum_{i\in S} \mathbf{z}_i \mathbf{z}_i^\top\bigg).
\label{eq:final_obj}
\end{equation}
\end{theorem}
\end{theorembox}
Theorem~\ref{thm:kdpp_objective} gives a unified objective that jointly balances forgetting-related utility and subset diversity. The full derivation is provided in Appendix~\ref{app:kdpp_to_ud}.

\begin{wrapfigure}[14]{r}{0.5\textwidth}
\vspace{-0.48in}
\begin{minipage}{\linewidth}
\begin{algorithm}[H]
\caption{NADS}
\small
\renewcommand{\baselinestretch}{1}\selectfont
\label{NADS}
\begin{algorithmic}[1]
\REQUIRE $M_{\text{pretrain}}$, $\mathcal{D}_{\text{new}}$, $\mathcal{C}=\{x_i\}_{i=1}^{N}$, $\mathbf{w}^{(0)}\leftarrow \mathbf{0}$, $K$, $\gamma,\eta,\nu,T,\lambda$
\ENSURE $M_{\text{final}}$
\STATE $M_{\text{nav}}\leftarrow M_{\text{pretrain}}$ using $\mathcal{D}_{\text{new}}$
\STATE Compute utilities $\{u_i\}_{i=1}^{N}$ by Eq.~\eqref{utility}
\STATE Extract candidate representations $\{z_i\}_{i=1}^{N}$ from $\mathcal{C}$.
\STATE Apply random projection by Eq.~\eqref{eq:rp_center} and whitening to obtain low-dimensional features $\{\tilde z_i\}_{i=1}^{N}$.
\STATE Construct the scalable surrogate objective $\widetilde{F}(\mathbf{w})$ in Eq.~\eqref{eq:Ft_surrogate} using $\{u_i,\tilde z_i\}_{i=1}^{N}$.
\FOR{$t=1$ to $T$}
    \STATE $\mathbf{w}^{(t)} \leftarrow \mathcal{P}_{K}\!\big(\big(\mathbf{w}^{(t-1)}+\eta\nabla\tilde{F}(\mathbf{w}^{(t-1)})\big)_{+}\big)$
\ENDFOR
\STATE $\mathcal{D}_{\text{constraint}}\leftarrow \{x_i\mid i\in\Omega_K(\mathbf{w}^{(T)})\}$
\STATE Train $M_{\text{final}}$ on $\mathcal{D}_{\text{new}}$ and $\mathcal{D}_{\text{constraint}}$ by Eq.~\eqref{eq:dual_overall}
\RETURN $M_{\text{final}}$
\end{algorithmic}
\end{algorithm}
\end{minipage}
\end{wrapfigure}
While Theorem~\ref{thm:kdpp_objective} yields a principled objective, directly applying it to large-scale subset selection remains computationally costly. Moreover, existing greedy solvers~\citep{chen2018fast,hemmi2022lazy} rely on repeated gain evaluation during subset expansion, which quickly becomes a major bottleneck at scale.

\subsection{Efficient Data Selection Algorithm}
\label{Selection_Algorithm}

We propose a scalable subset selection strategy, as summarized in Algorithm~\ref{NADS}. Specifically, we recast Eq.~\eqref{eq:final_obj} into a sparse-weight surrogate that retains its utility--diversity structure while making the optimization tractable. To further enhance optimization stability, we incorporate an $\ell_2$ regularization term with strength $\nu$. The resulting objective is
\begin{equation}
\label{eq:iht_obj}
\!F(\mathbf{w}) \!\triangleq\! \sum\nolimits_{i=1}^{N} w_i u_i + \beta \log\det\!\bigg(
\mathbf{I}_d+\sum\nolimits_{i=1}^{N} w_i\,\mathbf{z}_i\mathbf{z}_i^{\top}\bigg)
-\frac{\nu}{2}\|\mathbf{w}\|_2^2,
\quad \!\!
\mathbf{w}^*=\!\!\operatornamewithlimits{arg\,max}_{\mathbf{w}\succeq \mathbf{0},\,\|\mathbf{w}\|_0=K}\!\! F(\mathbf{w}).
\end{equation}
Further analysis of the discrepancy between Eq.~\eqref{eq:final_obj} and Eq.~\eqref{eq:iht_obj} is provided in Appendix~\ref{app:subset_to_sparse_gap}.

\textbf{Random Projection and Whitening.}
We apply a Johnson--Lindenstrauss random projection~\citep{johnson1984extensions} to improve efficiency while preserving feature geometry. Let $\mathbf{z}_i\in\mathbb{R}^d$ denote the sample-level representation of $x_i$, obtained by mean pooling last-layer hidden states over non-padding tokens. After centering, we map the features into an $E$-dimensional space using $\mathbf{R}\in\mathbb{R}^{d\times E}$ with $E\ll d$:
\begin{equation}
\tilde{\mathbf{z}}_i=\mathbf{R}^{\top}\!\left(\mathbf{z}_i-\boldsymbol{\mu}\right),
\quad
\boldsymbol{\mu}=\frac{1}{N}\textstyle\sum_{i=1}^{N}\mathbf{z}_i.
\label{eq:rp_center}
\end{equation}
To further reduce feature anisotropy and off-diagonal coupling, we apply whitening~\citep{kessy2018optimal} by computing
$\boldsymbol{\Sigma}=\frac{1}{N}\sum_{i=1}^{N}\tilde{\mathbf{z}}_i\tilde{\mathbf{z}}_i^{\top}$
and updating
$\tilde{\mathbf{z}}_i \leftarrow (\boldsymbol{\Sigma}+\rho\mathbf{I}_E)^{-\frac{1}{2}}\tilde{\mathbf{z}}_i$,
where $\rho$ is a regularization constant.

\textbf{Diagonal Approximation.}
Optimizing the log-determinant term remains computationally demanding, as the covariance still couples different feature dimensions. To obtain a more tractable objective, we introduce a diagonal approximation. Specifically, let
$\mathbf{A}(\mathbf{w}) \triangleq \sum_{i=1}^{N} w_i \tilde{\mathbf{z}}_i\tilde{\mathbf{z}}_i^{\top}$ and replace $\log\det(\mathbf I_E+\mathbf A(\mathbf w))$ with $\sum_{e=1}^{E}\log(1+A_{ee}(\mathbf w))$.
Under this approximation, the objective becomes
\begin{equation}
\label{eq:Ft_surrogate}
\tilde{F}(\mathbf{w}) \triangleq \sum\nolimits_{i=1}^{N} w_i u_i + \beta \sum\nolimits_{e=1}^{E}\log\!\bigg(1+\sum\nolimits_{i=1}^{N} w_i\,\tilde{z}_{i,e}^{\,2}\bigg)-\frac{\nu}{2}\|\mathbf{w}\|_2^2,
\quad
\mathbf{w}\succeq \mathbf{0},~\|\mathbf{w}\|_0=K,
\end{equation}
where $\tilde{z}_{i,e}$ denotes the $e$-th coordinate of $\tilde{\mathbf{z}}_i$, with diagonal error analysis in Appendix~\ref{app:diag_error}.

\textbf{Iterative Optimization.}
We adopt Iterative Hard Thresholding (IHT)~\citep{blumensath2009iterative} for scalable optimization under the nonnegativity and $K$-sparsity constraints. Specifically, we perform gradient ascent on $\tilde{F}(\mathbf{w})$, followed by a nonnegative hard-thresholding step that yields the weights at iteration $t$
\begin{equation}
\label{eq:iht_update}
\mathbf{w}^{(t)}
=
\mathcal{P}_{K}\!\Big(\big(\mathbf{w}^{(t-1)}+\eta\,\nabla \tilde{F}(\mathbf{w}^{(t-1)})\big)_{+}\Big),
\end{equation}
where $\eta$ is the step size, $(\cdot)_{+}$ denotes elementwise nonnegativity truncation, and $\mathcal{P}_{K}$ keeps the top-$K$ entries and sets the rest to zero. After $T$ iterations, we set $ S^*=\Omega_{K}(\mathbf{w}^{(T)})$ and construct $\mathcal{D}_{\text{constraint}}=\{x_i\mid i\in S^*\}$, where $\Omega_{K}$ returns the top-$K$ indices. Appendices~\ref{tF_derivation}, \ref{app:iht_update_derivation}, and \ref{app:iht_convergence} provide the derivations of $\nabla \tilde{F}(\mathbf{w})$ and Eq.~\eqref{eq:iht_update}, together with analysis of the iterative procedure.

\textbf{Time Complexity Analysis.}
The selection-stage complexity of NADS is $O(NEd + E^3 + T(NE + N\log K))$. Random projection and whitening incur a one-time preprocessing cost of $O(NEd + E^3)$, where the $O(NE^2)$ whitening cost is dominated by $O(NEd)$ since $E\ll d$. In each iteration, computing $1+\sum_{i=1}^{N} w_i \tilde{z}_{i,e}^2$ takes $O(NE)$ time, and projecting onto the feasible set by retaining the top-$K$ entries costs $O(N\log K)$. By contrast, Fast Greedy~\citep{chen2018fast} evaluates gains over the full candidate pool when adding each element, resulting in $O(KNd)$ complexity. Lazy-Fast variants~\citep{hemmi2022lazy} alleviate redundant gain evaluation via lazy updates, yet their reported overall complexity still remains $O(KN(d+\log N))$. Overall, our optimization yields a markedly more efficient route to optimizing a scalable surrogate of Eq.~\eqref{eq:final_obj} and remains well suited to large candidate pools.

\subsection{Learning with Knowledge Distillation}
\label{kd}
We train $M_{\text{final}}$ on $\mathcal{D}_{\text{new}}$ and $\mathcal{D}_{\text{constraint}}$ via KD, where $\mathcal{D}_{\text{new}}$ supports target-task adaptation and $\mathcal{D}_{\text{constraint}}$ provides capability-preserving supervision from $M_{\text{pretrain}}$. The overall training objective is
\begin{equation}
\mathcal{L}_{\text{final}}
=
\mathcal{L}_{\text{Task}}
+\lambda\,\mathcal{L}_{\text{KD}},
\label{eq:dual_overall}
\end{equation}
where $\lambda$ controls the distillation strength. Here, $\mathcal{L}_{\text{Task}}$ is the negative log-likelihood on $\mathcal{D}_{\text{new}}$, encouraging $M_{\text{final}}$ to fit the target task, while $\mathcal{L}_{\text{KD}}$ uses $M_{\text{pretrain}}$ as a teacher on $\mathcal{D}_{\text{constraint}}$ and penalizes deviations of $M_{\text{final}}$ from pretrained predictive behavior on the selected constraint samples.

\section{Experiments}
We evaluate NADS along four key dimensions: \textbf{Q1:} Does NADS better balance target-task adaptation and general capability preservation than SOTA baselines?  \textbf{Q2:} Does $M_{\text{nav}}$ provide an effective reference under the target recipe? \textbf{Q3:} Can NADS select more effective and diverse data with lower selection-stage overhead than alternative selection methods? \textbf{Q4:} Do the key components of NADS each contribute meaningfully, and does the framework remain robust across hyperparameter settings?

\subsection{Experiment Setup}

\textbf{Datasets \& Evaluation Metrics.}
We conduct experiments on Magicoder~\citep{pmlr-v235-wei24h} and MetaMathQA~\citep{yu2024metamath} to evaluate NADS, using Magpie-Qwen2.5-Pro-300K-Filtered~\citep{xu2025magpie} as the large-scale synthetic data source. Following recent studies~\citep{wang2025on,lin2026sft,sanyalupweighting,liu2026learn,yuan2026differential}, we decouple evaluation into target performance and general capability, with benchmark roles adjusted across scenarios. For Magicoder, HumanEval~(HE)~\citep{chen2021evaluatinglargelanguagemodels} is the target benchmark, while MMLU~\citep{hendrycks2021measuring}, HellaSwag~(HASW)~\citep{zellers2019hellaswag}, GSM8K~\citep{cobbe2021training}, and OpenBookQA~(OBQA)~\citep{mihaylov2018can} assess general capability. For MetaMathQA, GSM8K is the target benchmark, while HumanEval is moved to the general-capability suite. We report general capability (mean of four benchmarks), target performance, and Overall (arithmetic mean of the two).

\textbf{Baselines \& Implementation Details.}
We compare NADS with three baseline groups: 1) standard fine-tuning, including Standard Fine-Tuning (FT) and LoRA~\citep{hu2022lora}; 2) replay-based constraint selection, including Random selection, InsCL~\citep{wang2024inscl}, LER~\citep{m2024leitner}, and Mix-cd~\citep{bai2025efficient}; and 3) target-data selection, including LESS~\citep{xia2024less}, FisherSFT~\citep{deb2025fishersft}, and SPICE~\citep{chang2026spice}. We evaluate all methods on LLaMA3.2-3B~\citep{grattafiori2024llama3herdmodels}, Gemma3-4B~\citep{gemmateam2025gemma3technicalreport}, and Qwen3.5-9B~\citep{qwen3.5}. For each target dataset, we hold out $1{,}000$ examples for validation and use the remaining target-training split for all methods. Target-data selection methods select 10\% of this split, whereas replay-based methods and NADS use the entire split with $5\%$ of $\mathcal{C}$ as constraints. For NADS, we set $\gamma=0.7$, $\nu=3\times10^{-3}$, $\beta=1.0$, $\lambda=5$, $T=50$, and $E=256$. We report mean and standard deviation over three seeds. Details are in Appendix~\ref{app:implementation_details}.

\newcommand{\modelrow}[1]{%
\specialrule{\lightrulewidth}{0pt}{0pt}
\rowcolor{yellow!15}
\multicolumn{8}{c}{\fontsize{7}{9}\selectfont\textbf{\textit{#1}}} \\
}

\newcommand{\grouprow}[1]{%
\multicolumn{8}{l}{\cellcolor{gray!10}\textit{#1}} \\
}

\newcommand{\redunderline}[1]{%
  {\bgroup
  \markoverwith{\textcolor{red}{\rule[-0.5ex]{2pt}{0.4pt}}}%
  \ULon{#1}%
  \egroup}
}

\begin{table}[t]
\centering
\scriptsize
\renewcommand{\arraystretch}{1}
\setlength{\tabcolsep}{2.32mm}
\caption{Performance comparison with methods selecting $\mathcal{D}_{\text{constraint}}$ for replay on Magicoder (\%). The \textcolor{red}{best results} are highlighted \textcolor{red}{in red}, and the \textcolor{blue}{second-best} results are highlighted \textcolor{blue}{in blue}.}
\label{tab:magicoder_main}


\begin{tabular}{clccccc|c|c}
\toprule
\multirow{2}{*}{\textbf{Model}} 
& \multirow{2}{*}{\textbf{Method}} 
& \multicolumn{5}{c|}{\textbf{General Capability} $\uparrow$} 
& \multicolumn{1}{c|}{\textbf{Target} $\uparrow$} 
& \multicolumn{1}{c}{\textbf{Overall} $\uparrow$} \\
\cmidrule(lr){3-7} \cmidrule(lr){8-8} \cmidrule(lr){9-9}
& & \textbf{HASW} & \textbf{OBQA} & \textbf{MMLU} & \textbf{GSM8K} & \textbf{Avg.} & \textbf{HE} & \textbf{Avg.} \\
\midrule


& Pre-trained & $\text{55.83}_{\pm \text{0.00}}$ & $\text{31.20}_{\pm \text{0.00}}$ & $\text{56.92}_{\pm \text{0.00}}$ & $\text{27.90}_{\pm \text{0.00}}$ & $\text{42.96}_{\pm \text{0.00}}$ & $\text{28.16}_{\pm \text{0.00}}$ & $\text{35.56}_{\pm \text{0.00}}$ \\
& Standard FT & $\text{52.12}_{\pm \text{0.52}}$ & $\text{28.40}_{\pm \text{0.69}}$ & $\text{48.61}_{\pm \text{0.40}}$ & $\text{20.75}_{\pm \text{0.74}}$ & $\text{37.47}_{\pm \text{0.30}}$ & $\text{45.13}_{\pm \text{0.32}}$ & $\text{41.30}_{\pm \text{0.22}}$ \\
& LoRA & $\text{53.38}_{\pm \text{0.55}}$ & $\text{29.47}_{\pm \text{0.64}}$ & $\text{51.98}_{\pm \text{0.42}}$ & $\text{21.18}_{\pm \text{0.79}}$ & $\text{39.00}_{\pm \text{0.31}}$ & $\text{41.92}_{\pm \text{0.30}}$ & $\text{40.46}_{\pm \text{0.21}}$ \\
& Random & $\text{52.03}_{\pm \text{0.54}}$ & $\text{28.13}_{\pm \text{0.64}}$ & $\text{49.21}_{\pm \text{0.43}}$ & $\text{21.83}_{\pm \text{0.73}}$ & $\text{37.80}_{\pm \text{0.30}}$ & $\text{41.77}_{\pm \text{0.46}}$ & $\text{39.79}_{\pm \text{0.27}}$ \\

& InsCL & $\text{53.95}_{\pm \text{0.54}}$ & $\text{29.27}_{\pm \text{0.70}}$ & $\textcolor{blue}{\text{54.86}_{\pm \text{0.43}}}$ & $\text{25.40}_{\pm \text{0.76}}$ & $\text{40.87}_{\pm \text{0.31}}$ & $\text{42.02}_{\pm \text{0.82}}$ & $\text{41.45}_{\pm \text{0.44}}$ \\
& LER & $\text{53.77}_{\pm \text{0.49}}$ & $\text{29.47}_{\pm \text{0.58}}$ & $\text{53.89}_{\pm \text{0.40}}$ & $\text{24.82}_{\pm \text{0.73}}$ & $\text{40.49}_{\pm \text{0.28}}$ & $\textcolor{blue}{\text{43.12}_{\pm \text{0.75}}}$ & $\text{41.81}_{\pm \text{0.40}}$ \\
& Mix-cd & $\textcolor{blue}{\text{54.18}_{\pm \text{0.51}}}$ & $\textcolor{blue}{\text{30.47}_{\pm \text{0.61}}}$ & $\text{54.79}_{\pm \text{0.41}}$ & $\textcolor{blue}{\text{26.28}_{\pm \text{0.75}}}$ & $\textcolor{blue}{\text{41.43}_{\pm \text{0.29}}}$ & $\text{42.58}_{\pm \text{0.74}}$ & $\textcolor{blue}{\text{42.01}_{\pm \text{0.40}}}$ \\

\rowcolor{monaPurple}
\cellcolor{white}\multirow{-8}{*}{\rotatebox{90}{\textbf{LLaMA3.2-3B}}}
& NADS (Ours) & $\textcolor{red}{\text{55.17}_{\pm \text{0.49}}}$ & $\textcolor{red}{\text{31.07}_{\pm \text{0.58}}}$ & $\textcolor{red}{\text{56.45}_{\pm \text{0.41}}}$ & $\textcolor{red}{\text{27.37}_{\pm \text{0.67}}}$ & $\textcolor{red}{\text{42.52}_{\pm \text{0.27}}}$ & $\textcolor{red}{\text{44.39}_{\pm \text{0.40}}}$ & $\textcolor{red}{\text{43.46}_{\pm \text{0.24}}}$ \\

\midrule
& Pre-trained & $\text{57.41}_{\pm \text{0.00}}$ & $\text{34.60}_{\pm \text{0.00}}$ & $\text{60.24}_{\pm \text{0.00}}$ & $\text{36.69}_{\pm \text{0.00}}$ & $\text{47.24}_{\pm \text{0.00}}$ & $\text{35.77}_{\pm \text{0.00}}$ & $\text{41.51}_{\pm \text{0.00}}$ \\
& Standard FT & $\text{53.06}_{\pm \text{0.49}}$ & $\text{31.27}_{\pm \text{0.70}}$ & $\text{53.21}_{\pm \text{0.40}}$ & $\text{23.78}_{\pm \text{0.68}}$ & $\text{40.33}_{\pm \text{0.29}}$ & $\text{58.17}_{\pm \text{0.89}}$ & $\text{49.25}_{\pm \text{0.47}}$ \\
& LoRA & $\text{55.81}_{\pm \text{0.50}}$ & $\text{31.80}_{\pm \text{0.60}}$ & $\text{56.74}_{\pm \text{0.39}}$ & $\text{31.61}_{\pm \text{0.82}}$ & $\text{43.99}_{\pm \text{0.30}}$ & $\text{54.69}_{\pm \text{0.91}}$ & $\text{49.34}_{\pm \text{0.48}}$ \\
& Random & $\text{52.83}_{\pm \text{0.51}}$ & $\text{31.27}_{\pm \text{0.64}}$ & $\text{53.67}_{\pm \text{0.40}}$ & $\text{24.11}_{\pm \text{0.77}}$ & $\text{40.47}_{\pm \text{0.30}}$ & $\text{54.11}_{\pm \text{0.86}}$ & $\text{47.29}_{\pm \text{0.46}}$ \\

& InsCL & $\text{55.25}_{\pm \text{0.49}}$ & $\text{32.20}_{\pm \text{0.69}}$ & $\text{58.07}_{\pm \text{0.39}}$ & $\text{34.72}_{\pm \text{0.86}}$ & $\text{45.06}_{\pm \text{0.31}}$ & $\text{55.62}_{\pm \text{0.92}}$ & $\text{50.34}_{\pm \text{0.49}}$ \\
& LER & $\textcolor{red}{\text{56.88}_{\pm \text{0.50}}}$ & $\text{32.27}_{\pm \text{0.64}}$ & $\textcolor{blue}{\text{58.19}_{\pm \text{0.40}}}$ & $\text{34.39}_{\pm \text{0.84}}$ & $\textcolor{blue}{\text{45.43}_{\pm \text{0.31}}}$ & $\textcolor{blue}{\text{56.09}_{\pm \text{0.89}}}$ & $\textcolor{blue}{\text{50.76}_{\pm \text{0.47}}}$ \\
& Mix-cd & $\text{56.41}_{\pm \text{0.49}}$ & $\textcolor{blue}{\text{33.07}_{\pm \text{0.64}}}$ & $\text{57.28}_{\pm \text{0.39}}$ & $\textcolor{blue}{\text{34.75}_{\pm \text{0.83}}}$ & $\text{45.38}_{\pm \text{0.31}}$ & $\text{55.87}_{\pm \text{0.74}}$ & $\text{50.63}_{\pm \text{0.40}}$ \\

\rowcolor{monaPurple}
\cellcolor{white}\multirow{-8}{*}[-0.3ex]{\rotatebox{90}{\textbf{Gemma3-4B}}}
&NADS (Ours) & $\textcolor{blue}{\text{56.83}_{\pm \text{0.50}}}$ & $\textcolor{red}{\text{33.13}_{\pm \text{0.61}}}$ & $\textcolor{red}{\text{59.62}_{\pm \text{0.40}}}$ & $\textcolor{red}{\text{35.99}_{\pm \text{0.74}}}$ & $\textcolor{red}{\text{46.39}_{\pm \text{0.29}}}$ & $\textcolor{red}{\text{57.88}_{\pm \text{0.90}}}$ & $\textcolor{red}{\text{52.14}_{\pm \text{0.47}}}$ \\

\midrule
& Pre-trained & $\text{59.67}_{\pm \text{0.00}}$ & $\text{34.20}_{\pm \text{0.00}}$ & $\text{72.39}_{\pm \text{0.00}}$ & $\text{86.20}_{\pm \text{0.00}}$ & $\text{63.12}_{\pm \text{0.00}}$ & $\text{61.05}_{\pm \text{0.00}}$ & $\text{62.09}_{\pm \text{0.00}}$ \\
& Standard FT & $\text{54.21}_{\pm \text{0.50}}$ & $\text{31.07}_{\pm \text{0.64}}$ & $\text{66.22}_{\pm \text{0.39}}$ & $\text{78.57}_{\pm \text{0.70}}$ & $\text{57.52}_{\pm \text{0.29}}$ & $\text{75.82}_{\pm \text{0.81}}$ & $\text{66.67}_{\pm \text{0.43}}$ \\
& LoRA & $\text{56.73}_{\pm \text{0.49}}$ & $\text{32.27}_{\pm \text{0.61}}$ & $\text{69.44}_{\pm \text{0.37}}$ & $\text{81.88}_{\pm \text{0.55}}$ & $\text{60.08}_{\pm \text{0.26}}$ &$\text{70.33}_{\pm \text{0.64}}$ & $\text{65.21}_{\pm \text{0.34}}$ \\
& Random & $\text{54.78}_{\pm \text{0.50}}$ & $\text{31.33}_{\pm \text{0.64}}$ & $\text{65.29}_{\pm \text{0.40}}$ & $\text{79.23}_{\pm \text{0.72}}$ & $\text{57.66}_{\pm \text{0.29}}$ & $\textcolor{blue}{\text{72.45}_{\pm \text{0.82}}}$ & $\text{65.06}_{\pm \text{0.43}}$ \\

& InsCL          & $\text{56.36}_{\pm \text{0.49}}$ & $\text{32.27}_{\pm \text{0.64}}$ & $\text{70.18}_{\pm \text{0.37}}$ & $\text{82.54}_{\pm \text{0.46}}$ & $\text{60.34}_{\pm \text{0.25}}$ & $\text{71.29}_{\pm \text{0.85}}$ & $\text{65.82}_{\pm \text{0.44}}$ \\
& LER            & $\textcolor{blue}{\text{57.64}_{\pm \text{0.50}}}$ & $\text{32.33}_{\pm \text{0.61}}$ & $\textcolor{blue}{\text{71.27}_{\pm \text{0.39}}}$ & $\textcolor{blue}{\text{83.35}_{\pm \text{0.68}}}$ & $\textcolor{blue}{\text{61.15}_{\pm \text{0.28}}}$ & $\text{72.15}_{\pm \text{0.81}}$ & $\textcolor{blue}{\text{66.65}_{\pm \text{0.43}}}$ \\
& Mix-cd         & $\text{57.25}_{\pm \text{0.49}}$ & $\textcolor{blue}{\text{32.73}_{\pm \text{0.61}}}$ & $\text{70.49}_{\pm \text{0.41}}$ & $\text{82.54}_{\pm \text{0.70}}$ & $\text{60.75}_{\pm \text{0.28}}$ & $\text{71.37}_{\pm \text{0.72}}$ & $\text{66.06}_{\pm \text{0.39}}$ \\

\rowcolor{monaPurple}
\cellcolor{white}\multirow{-8}{*}[-0.3ex]{\rotatebox{90}{\textbf{Qwen3.5-9B}}}
& NADS (Ours)  & $\textcolor{red}{\text{58.49}_{\pm \text{0.49}}}$ & $\textcolor{red}{\text{33.13}_{\pm \text{0.61}}}$ & $\textcolor{red}{\text{71.61}_{\pm \text{0.41}}}$ & $\textcolor{red}{\text{85.62}_{\pm \text{0.69}}}$ & $\textcolor{red}{\text{62.21}_{\pm \text{0.28}}}$ & $\textcolor{red}{\text{74.46}_{\pm \text{0.81}}}$ & $\textcolor{red}{\text{68.34}_{\pm \text{0.43}}}$ \\

\bottomrule
\end{tabular}
\end{table}

\begin{table}[t]
\centering
\scriptsize
\renewcommand{\arraystretch}{1}
\setlength{\tabcolsep}{2.3mm}
\caption{Cross-paradigm reference comparison with methods selecting $\mathcal{D}_{\text{new}}$ on Magicoder (\%).}
\label{tab2}


\begin{tabular}{clccccc|c|c}
\toprule
\multirow{2}{*}{\textbf{Model}} 
& \multirow{2}{*}{\textbf{Method}} 
& \multicolumn{5}{c|}{\textbf{General Capability} $\uparrow$} 
& \multicolumn{1}{c|}{\textbf{Target} $\uparrow$} 
& \multicolumn{1}{c}{\textbf{Overall} $\uparrow$} \\
\cmidrule(lr){3-7} \cmidrule(lr){8-8} \cmidrule(lr){9-9}
& & \textbf{HASW} & \textbf{OBQA} & \textbf{MMLU} & \textbf{GSM8K} & \textbf{Avg.} & \textbf{HE} & \textbf{Avg.} \\
\midrule

&LESS & $\textcolor{blue}{\text{54.12}_{\pm \text{0.49}}}$ & $\textcolor{blue}{\text{30.67}_{\pm \text{0.61}}}$ & $\textcolor{blue}{\text{54.38}_{\pm \text{0.40}}}$ & $\text{23.73}_{\pm \text{0.62}}$ & $\textcolor{blue}{\text{40.73}_{\pm \text{0.27}}}$ & $\textcolor{blue}{\text{43.01}_{\pm \text{0.84}}}$ & $\textcolor{blue}{\text{41.87}_{\pm \text{0.44}}}$ \\
&FisherSFT & $\text{54.03}_{\pm \text{0.49}}$ & $\text{29.80}_{\pm \text{0.60}}$ & $\text{52.89}_{\pm \text{0.41}}$ & $\text{23.15}_{\pm \text{0.62}}$ & $\text{39.97}_{\pm \text{0.27}}$ & $\text{41.85}_{\pm \text{0.74}}$ & $\text{40.91}_{\pm \text{0.39}}$ \\
&SPICE & $\text{53.58}_{\pm \text{0.50}}$ & $\text{29.60}_{\pm \text{0.60}}$ & $\text{53.39}_{\pm \text{0.40}}$ & $\textcolor{blue}{\text{24.92}_{\pm \text{0.73}}}$ & $\text{40.37}_{\pm \text{0.29}}$ & $\text{42.61}_{\pm \text{0.66}}$ & $\text{41.49}_{\pm \text{0.36}}$ \\

\rowcolor{monaPurple}
\cellcolor{white}\multirow{-4}{*}{\rotatebox{90}{\tiny \textbf{LLaMA3.2}}}
& NADS (Ours) & $\textcolor{red}{\text{55.17}_{\pm \text{0.49}}}$ & $\textcolor{red}{\text{31.07}_{\pm \text{0.58}}}$ & $\textcolor{red}{\text{56.45}_{\pm \text{0.41}}}$ & $\textcolor{red}{\text{27.37}_{\pm \text{0.67}}}$ & $\textcolor{red}{\text{42.52}_{\pm \text{0.27}}}$ & $\textcolor{red}{\text{44.39}_{\pm \text{0.40}}}$ & $\textcolor{red}{\text{43.46}_{\pm \text{0.24}}}$ \\

\midrule

&LESS & $\text{55.56}_{\pm \text{0.52}}$ & $\text{31.80}_{\pm \text{0.69}}$ & $\text{57.26}_{\pm \text{0.39}}$ & $\text{35.08}_{\pm \text{0.83}}$ & $\text{44.93}_{\pm \text{0.31}}$ & $\textcolor{blue}{\text{57.07}_{\pm \text{0.91}}}$ & $\textcolor{blue}{\text{51.00}_{\pm \text{0.48}}}$ \\
&FisherSFT & $\textcolor{red}{\text{56.86}_{\pm \text{0.51}}}$ & $\textcolor{blue}{\text{32.20}_{\pm \text{0.60}}}$ & $\text{56.05}_{\pm \text{0.40}}$ & $\text{35.38}_{\pm \text{0.82}}$ & $\text{45.12}_{\pm \text{0.30}}$ & $\text{53.62}_{\pm \text{0.74}}$ & $\text{49.37}_{\pm \text{0.40}}$ \\
&SPICE & $\text{55.95}_{\pm \text{0.49}}$ & $\text{31.60}_{\pm \text{0.60}}$ & $\textcolor{blue}{\text{57.43}_{\pm \text{0.39}}}$ & $\textcolor{blue}{\text{35.61}_{\pm \text{0.85}}}$ & $\textcolor{blue}{\text{45.15}_{\pm \text{0.31}}}$ & $\text{56.46}_{\pm \text{0.76}}$ & $\text{50.81}_{\pm \text{0.41}}$ \\

\rowcolor{monaPurple}
\cellcolor{white}\multirow{-4}{*}[-0.3ex]{\rotatebox{90}{\tiny \textbf{Gemma3}}}
&NADS (Ours) & $\textcolor{blue}{\text{56.83}_{\pm \text{0.50}}}$ & $\textcolor{red}{\text{33.13}_{\pm \text{0.61}}}$ & $\textcolor{red}{\text{59.62}_{\pm \text{0.40}}}$ & $\textcolor{red}{\text{35.99}_{\pm \text{0.74}}}$ & $\textcolor{red}{\text{46.39}_{\pm \text{0.29}}}$ & $\textcolor{red}{\text{57.88}_{\pm \text{0.90}}}$ & $\textcolor{red}{\text{52.14}_{\pm \text{0.47}}}$ \\
\midrule

&LESS          & $\text{57.06}_{\pm \text{0.49}}$ & $\textcolor{blue}{\text{32.87}_{\pm \text{0.64}}}$ & $\textcolor{blue}{\text{69.68}_{\pm \text{0.40}}}$ & $\text{81.20}_{\pm \text{0.80}}$ & $\text{60.20}_{\pm \text{0.30}}$ & $\textcolor{blue}{\text{72.81}_{\pm \text{0.61}}}$ & $\textcolor{blue}{\text{66.51}_{\pm \text{0.34}}}$ \\
&FisherSFT     & $\textcolor{blue}{\text{57.69}_{\pm \text{0.50}}}$ & $\text{32.07}_{\pm \text{0.61}}$ & $\text{68.55}_{\pm \text{0.37}}$ & $\text{81.30}_{\pm \text{0.74}}$ & $\text{59.90}_{\pm \text{0.29}}$ & $\text{71.80}_{\pm \text{0.83}}$ & $\text{65.85}_{\pm \text{0.44}}$ \\
&SPICE         & $\text{56.92}_{\pm \text{0.49}}$ & $\text{32.13}_{\pm \text{0.61}}$ & $\text{69.31}_{\pm \text{0.40}}$ & $\textcolor{blue}{\text{83.42}_{\pm \text{0.65}}}$ & $\textcolor{blue}{\text{60.45}_{\pm \text{0.27}}}$ & $\text{72.35}_{\pm \text{0.78}}$ & $\text{66.40}_{\pm \text{0.41}}$ \\

\rowcolor{monaPurple}
\cellcolor{white}\multirow{-4}{*}[-0.3ex]{\rotatebox{90}{\tiny \textbf{Qwen3.5}}}
& NADS (Ours)  & $\textcolor{red}{\text{58.49}_{\pm \text{0.49}}}$ & $\textcolor{red}{\text{33.13}_{\pm \text{0.61}}}$ & $\textcolor{red}{\text{71.61}_{\pm \text{0.41}}}$ & $\textcolor{red}{\text{85.62}_{\pm \text{0.69}}}$ & $\textcolor{red}{\text{62.21}_{\pm \text{0.28}}}$ & $\textcolor{red}{\text{74.46}_{\pm \text{0.81}}}$ & $\textcolor{red}{\text{68.34}_{\pm \text{0.43}}}$ \\

\bottomrule
\end{tabular}\vspace{-0.05in}
\end{table}

\subsection{Main Results~(Q1)}

\textbf{Comparison Results.}
Table~\ref{tab:magicoder_main} reports the main Magicoder comparison with replay-based constraint-selection methods. On LLaMA3.2, NADS achieves 42.52\% general-capability and 43.46\% overall averages, compared with 41.43\% and 42.01\% for the strongest baseline, while remaining competitive on the target task. Similar trends on Gemma3 and Qwen3.5 indicate stable gains across model scales. Since target-task data selection can also mitigate forgetting~\citep{wu2025mitigating,liu2026learn,zhang2026training}, Table~\ref{tab2} reports LESS, FisherSFT, and SPICE as cross-paradigm references, which select from $\mathcal{D}_{\text{new}}$ rather than constructing synthetic constraints. This reference comparison is not a direct ranking, but indicates that NADS remains competitive under a different selection paradigm. MetaMathQA results are in Appendix~\ref{MetaMathQA_result}.

As shown in Table~\ref{tab:magicoder_main}, existing replay-based baselines partly mitigate the adaptation--preservation trade-off, but limited gains reflect selection signals that do not target update-induced forgetting. In contrast, NADS constructs the constraint set from navigator-induced predictive drift, making selected samples more sensitive to forgetting-related changes under the target recipe. This targeted, less redundant constraint set supports target-task adaptation while maintaining stronger general capability.

\begin{figure}[t]
    \centering
    \includegraphics[width=\linewidth]{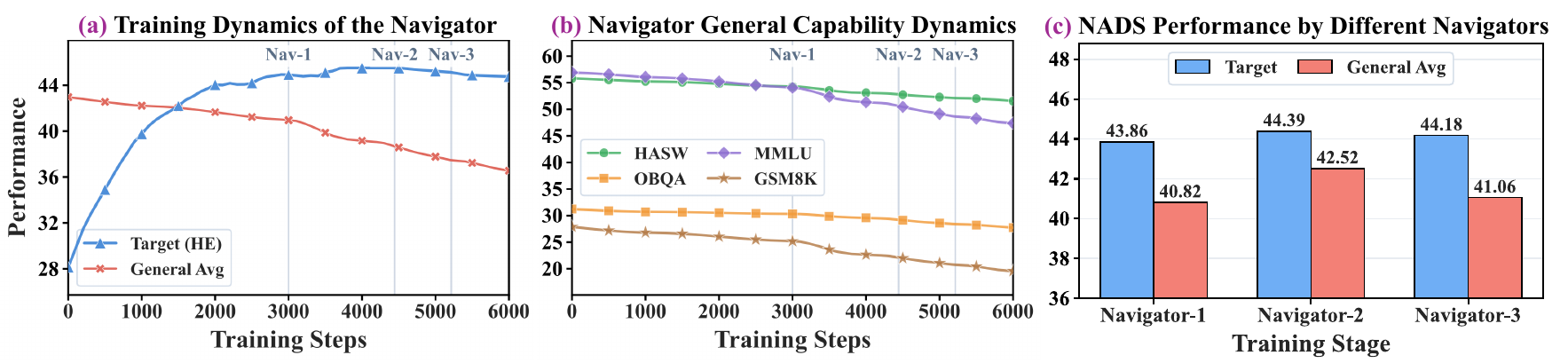}\vspace{-0.2in}
    \caption{(a) Training dynamics of the navigator model on the Magicoder dataset. (b) Dynamics of individual general-capability benchmarks during navigator training. (c) NADS performance with navigators selected at different training stages on Magicoder with LLaMA3.2-3B.}
    \label{fig3}
\end{figure}

\begin{wrapfigure}{r}{0.54\textwidth}\vspace{-0.18in}
    \centering
    \includegraphics[width=\linewidth, trim=2mm 1mm 1mm 0mm, clip]{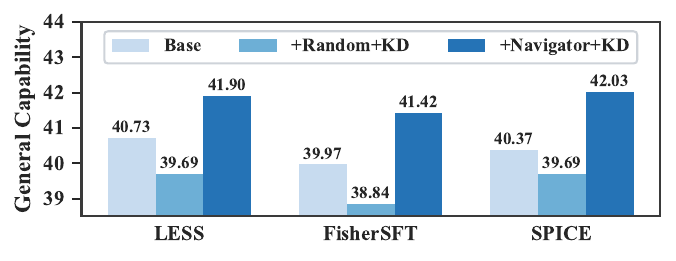}\vspace{-0.32in}
    \caption{Controlled comparison with a fixed KD objective on Magicoder with LLaMA3.2.}\vspace{-0.15in}
    \label{controlled_comparison}
\end{wrapfigure}
\textbf{Controlled comparison with a fixed KD objective.}
A potential concern is that the gains of NADS in Table~\ref{tab2} may come from the additional KD objective rather than the selected constraint set. Since NADS selects synthetic constraints rather than new-task samples, using LESS, FisherSFT, or SPICE-selected data would evaluate a combined method instead of NADS itself. We therefore fix the KD objective and vary only the constraint source. We apply the same KD objective to three baselines with either random or navigator-selected constraints. As shown in Fig.~\ref{controlled_comparison}, random constraints do not reliably preserve general capability and can even hurt retention, whereas navigator-selected constraints consistently improve it across baselines. This suggests that the improvement is not merely due to KD or additional constraint samples, but to the effectiveness of the selected constraints. Details are given in Appendix~\ref{app:training_objective_fairness}.

\subsection{Analysis of the Navigator Model (Q2)}
\label{sec:navigator_analysis}
The navigator $M_{\text{nav}}$ is central to NADS, providing the signal for estimating update-induced forgetting risk. Thus, we examine whether $M_{\text{nav}}$ is useful and stable in practice from the following two aspects.

\begin{wraptable}{r}{0.43\linewidth}
\vspace{-0.22in}
\centering
\small
\renewcommand{\arraystretch}{1.1}
\setlength{\tabcolsep}{0.2mm}
\caption{Performance comparison between $M_{\text{pretrain}}$ and $M_{\text{nav}}$.}
\label{tab:navigator_performance}
\begin{tabular}{lccccc}
\toprule
\textbf{Model} & \textbf{HASW} & \textbf{OBQA} & \textbf{MMLU} & \textbf{GSM8K}  & \textbf{HE}\\
\midrule
$M_{\text{pretrain}}$ & 55.83 & 31.20 & 56.92 & 27.90 & 28.16 \\
$M_{\text{nav}}$      & 52.80 & 29.20 & 50.60 & 22.09 & 45.50 \\
\bottomrule
\end{tabular}\vspace{-0.1in}
\end{wraptable}
\textbf{Unconstrained Drift Behavior.} Table~\ref{tab:navigator_performance} shows the performance of $M_{\text{pretrain}}$ and $M_{\text{nav}}$ on Magicoder with LLaMA3.2. $M_{\text{nav}}$ achieves much stronger new-task performance, indicating that it has effectively adapted to the target data. Meanwhile, because it is trained without any anti-forgetting constraints, this gain is accompanied by noticeable degradation on prior capability.

\textbf{Training Dynamics of the Navigator Model.} Fig.~\ref{fig3}(a) and Fig.~\ref{fig3}(b) depict the dynamics of new-task performance and general capability throughout $M_{\text{nav}}$ training on LLaMA3.2-3B. As training proceeds, $M_{\text{nav}}$ improves on the new task while progressively degrading on prior capability, reflecting increasing update-induced drift. To further examine how the navigator training stage affects constraint-set construction, Fig.~\ref{fig3}(c) compares NADS using $M_{\text{nav}}$-1, $M_{\text{nav}}$-2, and $M_{\text{nav}}$-3 from three training stages. $M_{\text{nav}}$-1 is trained too early and remains too close to $M_{\text{pretrain}}$, resulting in a weak forgetting-risk signal and a less informative constraint set. $M_{\text{nav}}$-3 is trained for too long, causing more aggressive drift that biases forgetting-risk estimation and weakens general capability preservation. In contrast, $M_{\text{nav}}$-2 achieves the best trade-off between target-task performance and general capability preservation. Additional checkpoint comparisons across models and target datasets and sensitivity analyses under validation-set perturbations are provided in Appendix~\ref{sec:checkpoint_generalization} and Appendix~\ref{sec:validation_sensitivity}, respectively.


\subsection{Effectiveness and Efficiency of Data Selection~(Q3)}
We further analyze the data selection strategy of NADS from three perspectives by examining whether it can select higher-quality data, maintain greater diversity, and reduce selection cost.

\begin{wraptable}{r}{0.45\textwidth}
\vspace{-0.28in}
\centering
\small
\caption{Controlled constraint-selection comparison on Magicoder.}
\label{tab:selected_data_quality}
\setlength{\tabcolsep}{6.5pt}
\begin{tabular}{lccc}
\toprule
\textbf{Method} & \textbf{General} & \textbf{Target} & \textbf{Overall} \\
\midrule
Random  & 37.96 & 41.91 & 39.94  \\
PPL-based  & 38.76 & 42.02 & 40.39 \\
DPP-based  & 38.04 & 41.27 & 39.66 \\
\rowcolor{monaBlue}
NADS       & \textbf{42.52} & \textbf{44.39} & \textbf{43.46} \\
\bottomrule
\end{tabular}
\vspace{-0.25in}
\end{wraptable}
\textbf{Quality of the Selected Data.}
We compare NADS with Random, PPL-based, and DPP-based selection on LLaMA3.2-3B under the same final KD objective and $D_{\text{constraint}}$ budget. Selected-data quality is measured by downstream performance after training on each selected set. As shown in Table~\ref{tab:selected_data_quality}, the alternatives incur larger general-capability drops, indicating that NADS identifies higher-quality data.

\begin{figure}[t]
    \centering\
    \includegraphics[width=0.98\linewidth, trim=8mm 0mm 0mm 0mm, clip]{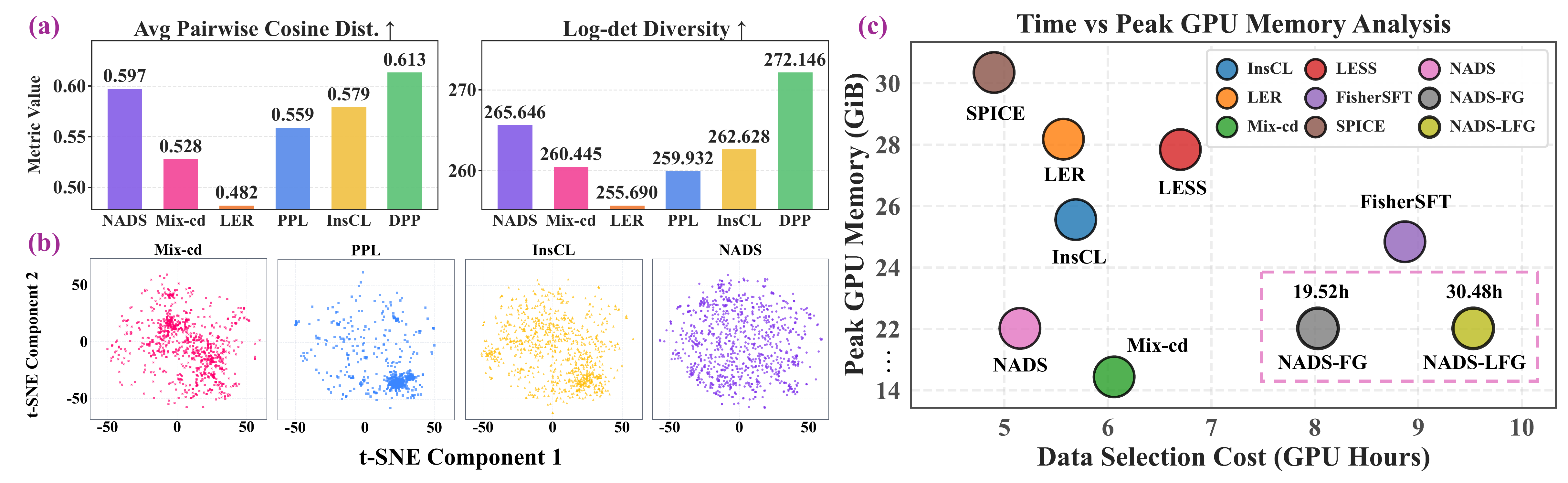}\vspace{-0.1in}
    \caption{(a) Quantitative diversity diagnostics of selected data. (b) t-SNE visualizations of selected data. (c) Data-selection cost in terms of selection time and peak GPU memory usage.}\vspace{-0.1in}
    \label{fig4}
\end{figure}
\textbf{Diversity of the Selected Set.}
We further examine whether the proposed strategy selects more diverse data on Magicoder with LLaMA3.2-3B. As shown in Fig.~\ref{fig4}(a), NADS achieves strong average pairwise cosine distance and log-det diversity among the compared methods, suggesting that it selects less redundant constraint samples. Fig.~\ref{fig4}(b) further visualizes the selected samples using t-SNE~\citep{van2008visualizing} in the representation space, where the data selected by NADS are distributed more broadly than those selected by utility-oriented competing methods. Additional qualitative and quantitative comparisons with more selection methods are provided in Appendix~\ref{diversity}.

\textbf{Data-Selection Efficiency.}
Fig.~\ref{fig4}(c) compares the runtime and peak GPU memory of different data-selection methods for selecting 10\% of the data under the same candidate pool and budget. The reported cost covers the selection stage, including method-specific score or utility computation, related preprocessing, and subset optimization. NADS reduces selection-stage runtime relative to competing methods while maintaining comparable peak GPU memory, demonstrating efficient large-scale data selection. We further compare the overhead after replacing our optimizer with Fast Greedy (NADS-FG)~\citep{chen2018fast} and Lazy-Fast Greedy (NADS-LFG)~\citep{hemmi2022lazy}. The time overhead analysis of $M_{\text{nav}}$ and the performance--cost analysis of the complete training pipeline are provided in Appendix~\ref{time_complexity}. Stage-wise and total computational costs across model sizes, including the cost savings from LoRA-based Navigator training, are reported in Appendix~\ref{sec:cost_scaling}.

\subsection{Ablation Study and Sensitivity Analysis~(Q4)}

\begin{figure}[t]
    \centering
    \includegraphics[width=\linewidth, trim=16mm 0mm 2mm 0mm, clip]{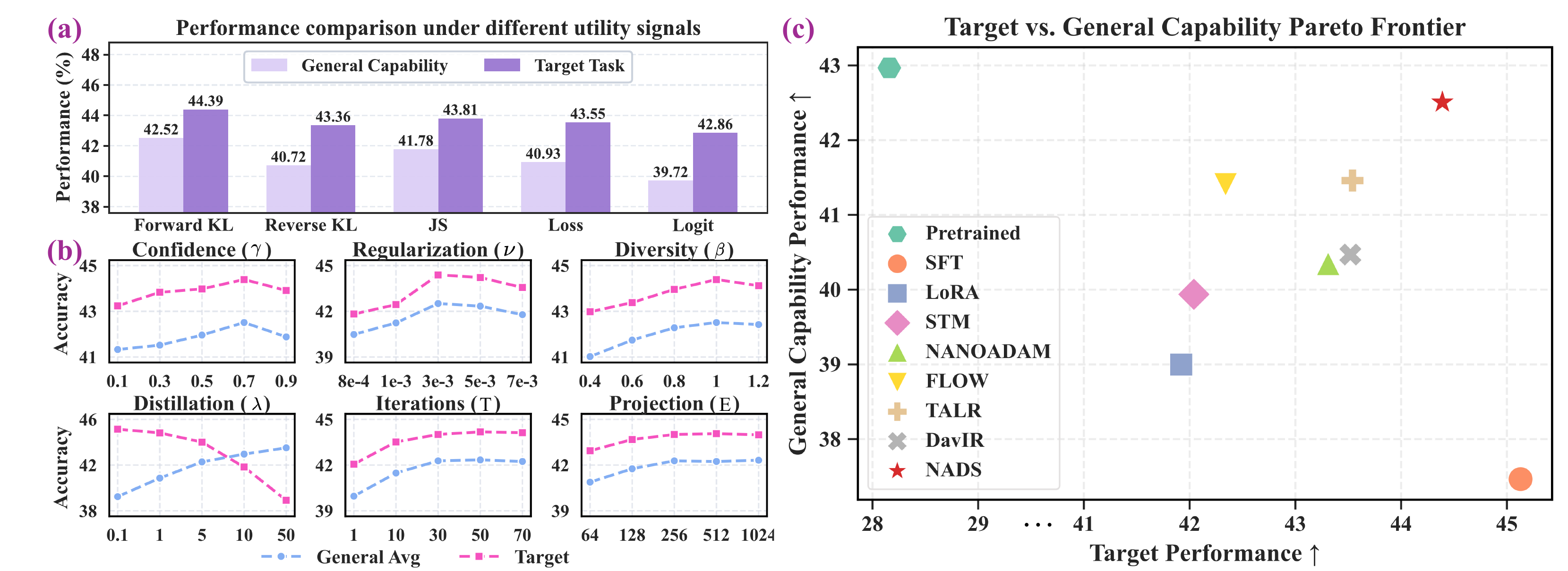}\vspace{-0.2in}
    \caption{(a) Performance comparison under different utility signals. (b) Hyperparameter sensitivity of NADS. (c) Comparison with catastrophic forgetting mitigation methods on the Pareto frontier.}
    \label{fig5}
\end{figure}

\textbf{Ablation Study.}
We conduct ablation studies to assess the key components of NADS. We first ablate the KL-discrepancy factor in the original utility, leaving only the pretrained confidence gate for scoring candidate samples. We also remove the diversity-aware selection term and the confidence gate in utility estimation. As shown in Table~\ref{tab:ablation_nads}, all modifications underperform the full NADS. The gate-only utility causes the largest degradation, indicating that pretrained confidence alone is \begin{wraptable}{r}{0.5\linewidth}
\vspace{-0.1in}
\centering
\small
\renewcommand{\arraystretch}{1}
\setlength{\tabcolsep}{1.3mm}
\caption{Ablation of NADS components.}
\label{tab:ablation_nads}
\begin{tabular}{l|ccc}
\toprule
\textbf{Method} & \textbf{General} & \textbf{Target} & \textbf{Overall} \\
\midrule
NADS                        & \textbf{42.52} & \textbf{44.39} & \textbf{43.46}              \\
\quad - w/o KL drift        & 37.64           & 42.31 & 39.98           \\
\quad - w/o Diversity       & 40.21           & 42.83 & 41.52           \\
\quad - w/o Entropy Gate    & 41.26           & 43.17 & 42.22           \\
\bottomrule
\end{tabular}\vspace{-0.15in}
\end{wraptable}
\noindent insufficient and that navigator-induced predictive discrepancy provides essential forgetting-aware information. Removing the diversity-aware term or the confidence gate also consistently hurts performance, validating the proposed utility and diversity design. We further compare utility signals in Fig.~\ref{fig5}(a). Forward KL performs best, followed by JS divergence. This is consistent with the final KD objective, where the pretrained model serves as the teacher. Thus, measuring drift from the pretrained distribution provides a more aligned signal for selecting preservation constraints.


\textbf{Hyperparameter Analysis.}
We conduct sensitivity analyses for the major hyperparameters in NADS. Fig.~\ref{fig5}(b) shows that NADS remains effective over a broad range of settings. For most hyperparameters, moving from very small values to moderate ones improves performance, after which the curves largely plateau. The main exception is the distillation strength, which exhibits a clearer trade-off between target-task adaptation and general capability preservation. Larger values strengthen preservation, but overly strong distillation begins to hurt target-task performance. The best overall balance is therefore achieved in the intermediate regime.

\textbf{Additional Robustness Analysis.}
We further examine whether NADS is sensitive to the candidate pool or constraint-set size. Results show that NADS remains effective across different candidate pools and constraint-set sizes, with detailed comparisons in Appendices~\ref{constraint_set_size} and~\ref{synthetic}. We also evaluate instruction-tuning performance using the Alpaca-GPT4 dataset~\cite{peng2023instruction} across ten general-capability benchmarks. Detailed experimental settings and results are provided in Appendix~\ref{app:alpaca_gpt4_results}.

\subsection{Comparison with Catastrophic Forgetting Mitigation Methods}

We compare NADS with recent catastrophic-forgetting mitigation methods, including STM~\citep{wu2025mitigating}, NANOADAM~\citep{zhou2025pay}, FLOW~\citep{sanyalupweighting}, DavIR~\citep{zhou2025davir}, and TALR~\citep{lin2026sft}. Fig.~\ref{fig5}(c) reports the comparison on Magicoder with LLaMA3.2-3B, where NADS achieves a favorable overall balance between new-task adaptation and general capability preservation. Detailed analyses and comparisons are in Appendix~\ref{cfmm}.

\section{Conclusion}
We present NADS, a navigator-guided data selection framework for mitigating catastrophic forgetting in LLM fine-tuning without pretraining data. By converting recipe-induced navigator drift into a forgetting-aware signal, NADS selects diverse synthetic constraints that support pretrained-behavior preservation. Experiments across tasks and model families show that NADS improves the adaptation--preservation trade-off with practical selection cost. These results indicate that navigator-induced predictive drift provides a useful bridge between target-task adaptation and preservation-constraint selection. Future work may explore lower-cost navigator construction and adaptive selection strategies.

\section*{Acknowledgments}
We thank the anonymous NeurIPS 2026 reviewers for their constructive feedback. This work is supported by the National Natural Science Foundation of China under Grant No. 62476191 and 42550187, with computing resources provided by our research group.


\medskip

{
\small
\bibliographystyle{IEEEtran}
\bibliography{custom}

@inproceedings{huang2024mitigating,
  title={Mitigating catastrophic forgetting in large language models with self-synthesized rehearsal},
  author={Huang, Jianheng and Cui, Leyang and Wang, Ante and Yang, Chengyi and Liao, Xinting and Song, Linfeng and Yao, Junfeng and Su, Jinsong},
  booktitle={Proceedings of the 62nd Annual Meeting of the Association for Computational Linguistics (Volume 1: Long Papers)},
  pages={1416--1428},
  year={2024}
}

@inproceedings{nikdan2025efficient,
    title={Efficient Data Selection at Scale via Influence Distillation},
    author={Mahdi Nikdan and Vincent Cohen-Addad and Dan Alistarh and Vahab Mirrokni},
    booktitle={The Thirty-ninth Annual Conference on Neural Information Processing Systems},
    year={2025},
}

@inproceedings{chen2026influencepreserving,
    title={Influence-Preserving Proxies for Gradient-Based Data Selection in {LLM} FineTuning},
    author={Sirui Chen and Yunzhe Qi and Mengting Ai and Yifan Sun and Ruizhong Qiu and Jiaru Zou and Jingrui He},
    booktitle={The Fourteenth International Conference on Learning Representations},
    year={2026},
}

@inproceedings{hoang2026difficultydiversity,
    title={Difficulty{\textendash}Diversity Collaborative Filtering for Data-Efficient {LLM} Fine-Tuning},
    author={Long P. Hoang and Wenxuan Zhang and Wei Lu},
    booktitle={The Fourteenth International Conference on Learning Representations},
    year={2026},
}

@inproceedings{wangself,
  title={Self-Evolving Pseudo-Rehearsal for Catastrophic Forgetting with Task Similarity in LLMs},
  author={Wang, Jun and Ding, Liang and Wang, Shuai and Li, Hongyu and Luo, Yong and Zhao, Huangxuan and Hu, Han and Du, Bo},
  booktitle={The Thirty-ninth Annual Conference on Neural Information Processing Systems},
  year={2025}
}

@article{shi2025continual,
  title={Continual learning of large language models: A comprehensive survey},
  author={Shi, Haizhou and Xu, Zihao and Wang, Hengyi and Qin, Weiyi and Wang, Wenyuan and Wang, Yibin and Wang, Zifeng and Ebrahimi, Sayna and Wang, Hao},
  journal={ACM Computing Surveys},
  volume={58},
  number={5},
  pages={1--42},
  year={2025},
  publisher={ACM New York, NY}
}

@inproceedings{wang2024inscl,
  title={InsCL: A Data-efficient Continual Learning Paradigm for Fine-tuning Large Language Models with Instructions},
  author={Wang, Yifan and Liu, Yafei and Shi, Chufan and Li, Haoling and Chen, Chen and Lu, Haonan and Yang, Yujiu},
  booktitle={Proceedings of the 2024 Conference of the North American Chapter of the Association for Computational Linguistics: Human Language Technologies (Volume 1: Long Papers)},
  pages={663--677},
  year={2024}
}

@inproceedings{sanyalupweighting,
  title={Upweighting Easy Samples in Fine-Tuning Mitigates Forgetting},
  author={Sanyal, Sunny and Prairie, Hayden and Das, Rudrajit and Kavis, Ali and Sanghavi, Sujay},
  booktitle={Forty-second International Conference on Machine Learning},
  year={2025}
}

@inproceedings{bethunescaling,
title={Scaling Laws for Forgetting during Finetuning with Pretraining Data Injection},
author={Louis B{\'e}thune and David Grangier and Dan Busbridge and Eleonora Gualdoni and marco cuturi and Pierre Ablin},
booktitle={Forty-second International Conference on Machine Learning},
year={2025},
}

@inproceedings{hacohenpredicting,
  title={Predicting the Susceptibility of Examples to Catastrophic Forgetting},
  author={Hacohen, Guy and Tuytelaars, Tinne},
  booktitle={International Conference on Machine Learning},
  pages={21546--21569},
  year={2025},
}

@inproceedings{dengunlocking,
    title={Unlocking the Power of Rehearsal in Continual Learning: A Theoretical Perspective},
    author={Junze Deng and Qinhang Wu and Peizhong Ju and Sen Lin and Yingbin Liang and Ness Shroff},
    booktitle={Forty-second International Conference on Machine Learning},
  year={2025}
}

@misc{abbes2025revisiting,
      title={Revisiting Replay and Gradient Alignment for Continual Pre-Training of Large Language Models}, 
      author={Istabrak Abbes and Gopeshh Subbaraj and Matthew Riemer and Nizar Islah and Benjamin Therien and Tsuguchika Tabaru and Hiroaki Kingetsu and Sarath Chandar and Irina Rish},
      year={2025},
      eprint={2508.01908},
      archivePrefix={arXiv},
      primaryClass={cs.LG},
      url={https://arxiv.org/abs/2508.01908}, 
}

@inproceedings{yang2025continual,
  title={Continual Model Merging without Data: Dual Projections for Balancing Stability and Plasticity},
  author={Yang, Enneng and Tang, Anke and Shen, Li and Guo, Guibing and Wang, Xingwei and Cao, Xiaochun and Zhang, Jie},
  booktitle={The Thirty-ninth Annual Conference on Neural Information Processing Systems},
  year={2025}
}

@article{bidermanlora,
    title={Lo{RA} Learns Less and Forgets Less},
    author={Dan Biderman and Jacob Portes and Jose Javier Gonzalez Ortiz and Mansheej Paul and Philip Greengard and Connor Jennings and Daniel King and Sam Havens and Vitaliy Chiley and Jonathan Frankle and Cody Blakeney and John Patrick Cunningham},
    journal={Transactions on Machine Learning Research},
    issn={2835-8856},
    year={2024}
}

@misc{ding2025improved,
      title={Improved Supervised Fine-Tuning for Large Language Models to Mitigate Catastrophic Forgetting}, 
      author={Fei Ding and Baiqiao Wang},
      year={2025},
      eprint={2506.09428},
      archivePrefix={arXiv},
      primaryClass={cs.CL},
      url={https://arxiv.org/abs/2506.09428},  
}

@inproceedings{liu2025take,
  title={Take the essence and discard the dross: A rethinking on data selection for fine-tuning large language models},
  author={Liu, Ziche and Ke, Rui and Liu, Yajiao and Jiang, Feng and Li, Haizhou},
  booktitle={Proceedings of the 2025 Conference of the Nations of the Americas Chapter of the Association for Computational Linguistics: Human Language Technologies (Volume 1: Long Papers)},
  pages={6595--6611},
  year={2025}
}

@misc{ivison2025larges,
      title={Large-Scale Data Selection for Instruction Tuning}, 
      author={Hamish Ivison and Muru Zhang and Faeze Brahman and Pang Wei Koh and Pradeep Dasigi},
      year={2025},
      eprint={2503.01807},
      archivePrefix={arXiv},
      primaryClass={cs.CL},
      url={https://arxiv.org/abs/2503.01807},  
}

@inproceedings{wang2025nice,
    title={{NICE} Data Selection for Instruction Tuning in {LLM}s with Non-differentiable Evaluation Metric},
    author={Jingtan Wang and Xiaoqiang Lin and Rui Qiao and Pang Wei Koh and Chuan-Sheng Foo and Bryan Kian Hsiang Low},
    booktitle={Forty-second International Conference on Machine Learning},
    year={2025},
}

@inproceedings{sam2025analyzing,
  title={Analyzing Similarity Metrics for Data Selection for Language Model Pretraining},
  author={Dylan Sam and Ayan Chakrabarti and Afshin Rostamizadeh and Srikumar Ramalingam and Gui Citovsky and Sanjiv Kumar},
  booktitle={The Thirty-ninth Annual Conference on Neural Information Processing Systems},
  year={2025},
}

@article{yu2024mates,
  title={Mates: Model-aware data selection for efficient pretraining with data influence models},
  author={Yu, Zichun and Das, Spandan and Xiong, Chenyan},
  journal={Advances in Neural Information Processing Systems},
  volume={37},
  pages={108735--108759},
  year={2024}
}

@misc{cai2025low,
      title={Low-Confidence Gold: Refining Low-Confidence Samples for Efficient Instruction Tuning}, 
      author={Hongyi Cai and Jie Li and Mohammad Mahdinur Rahman and Wenzhen Dong},
      year={2026},
      eprint={2502.18978},
      archivePrefix={arXiv},
      primaryClass={cs.CL},
      url={https://arxiv.org/abs/2502.18978}, 
}

@misc{li2025mdit,
      title={MDIT: A Model-free Data Interpolation Method for Diverse Instruction Tuning}, 
      author={Yangning Li and Zihua Lan and Lv Qingsong and Yinghui Li and Hai-Tao Zheng},
      year={2025},
      eprint={2504.07288},
      archivePrefix={arXiv},
      primaryClass={cs.CL},
      url={https://arxiv.org/abs/2504.07288}, 
}

@article{kessy2018optimal,
  title={Optimal whitening and decorrelation},
  author={Kessy, Agnan and Lewin, Alex and Strimmer, Korbinian},
  journal={The American Statistician},
  volume={72},
  number={4},
  pages={309--314},
  year={2018},
  publisher={Taylor \& Francis}
}

@article{kulesza2012determinantal,
  title={Determinantal point processes for machine learning},
  author={Kulesza, Alex and Taskar, Ben and others},
  journal={Foundations and Trends{\textregistered} in Machine Learning},
  volume={5},
  number={2--3},
  pages={123--286},
  year={2012},
  publisher={Now Publishers, Inc.}
}

@article{wang2024comprehensive,
  title={A comprehensive survey of continual learning: Theory, method and application},
  author={Wang, Liyuan and Zhang, Xingxing and Su, Hang and Zhu, Jun},
  journal={IEEE transactions on pattern analysis and machine intelligence},
  volume={46},
  number={8},
  pages={5362--5383},
  year={2024},
  publisher={IEEE}
}

@inproceedings{li2024revisiting,
  title={Revisiting Catastrophic Forgetting in Large Language Model Tuning},
  author={Li, Hongyu and Ding, Liang and Fang, Meng and Tao, Dacheng},
  booktitle={Findings of the Association for Computational Linguistics: EMNLP 2024},
  pages={4297--4308},
  year={2024}
}

@misc{wang2025see,
      title={SEE: Continual Fine-tuning with Sequential Ensemble of Experts}, 
      author={Zhilin Wang and Yafu Li and Xiaoye Qu and Yu Cheng},
      year={2025},
      eprint={2504.06664},
      archivePrefix={arXiv},
      primaryClass={cs.CL},
      url={https://arxiv.org/abs/2504.06664}, 
}

@article{wang2024model,
  title={Model sensitivity aware continual learning},
  author={Wang, Zhenyi and Huang, Heng},
  journal={Advances in Neural Information Processing Systems},
  volume={37},
  pages={132583--132613},
  year={2024}
}

@inproceedings{zhao2024a,
  title={A Statistical Theory of Regularization-Based Continual Learning},
  author={Xuyang Zhao and Huiyuan Wang and Weiran Huang and Wei Lin},
  booktitle={Forty-first International Conference on Machine Learning},
  year={2024}
}

@inproceedings{wu2024meta,
  title={Meta continual learning revisited: Implicitly enhancing online hessian approximation via variance reduction},
  author={Wu, Yichen and Huang, Long-Kai and Wang, Renzhen and Meng, Deyu and Wei, Ying},
  booktitle={The Twelfth international conference on learning representations},
  year={2024}
}

@misc{sarkar2025adapt,
      title={Adaptive Variance-Penalized Continual Learning with Fisher Regularization}, 
      author={Krisanu Sarkar},
      year={2025},
      eprint={2508.16632},
      archivePrefix={arXiv},
      primaryClass={cs.LG},
      url={https://arxiv.org/abs/2508.16632}, 
}

@inproceedings{yin2022contintin,
  title={Contintin: Continual learning from task instructions},
  author={Yin, Wenpeng and Li, Jia and Xiong, Caiming},
  booktitle={Proceedings of the 60th Annual Meeting of the Association for Computational Linguistics (Volume 1: Long Papers)},
  pages={3062--3072},
  year={2022}
}

@article{borsos2020coresets,
  title={Coresets via bilevel optimization for continual learning and streaming},
  author={Borsos, Zal{\'a}n and Mutny, Mojmir and Krause, Andreas},
  journal={Advances in neural information processing systems},
  volume={33},
  pages={14879--14890},
  year={2020}
}

@inproceedings{maekawa2023generative,
  title={Generative replay inspired by hippocampal memory indexing for continual language learning},
  author={Maekawa, Aru and Kamigaito, Hidetaka and Funakoshi, Kotaro and Okumura, Manabu},
  booktitle={Proceedings of the 17th Conference of the European Chapter of the Association for Computational Linguistics},
  pages={930--942},
  year={2023}
}

@article{wang2023incorporating,
  title={Incorporating neuro-inspired adaptability for continual learning in artificial intelligence},
  author={Wang, Liyuan and Zhang, Xingxing and Li, Qian and Zhang, Mingtian and Su, Hang and Zhu, Jun and Zhong, Yi},
  journal={Nature Machine Intelligence},
  volume={5},
  number={12},
  pages={1356--1368},
  year={2023},
  publisher={Nature Publishing Group UK London}
}

@inproceedings{zhao2024sapt,
  title={SAPT: A Shared Attention Framework for Parameter-Efficient Continual Learning of Large Language Models},
  author={Zhao, Weixiang and Wang, Shilong and Hu, Yulin and Zhao, Yanyan and Qin, Bing and Zhang, Xuanyu and Yang, Qing and Xu, Dongliang and Che, Wanxiang},
  booktitle={Proceedings of the 62nd Annual Meeting of the Association for Computational Linguistics (Volume 1: Long Papers)},
  pages={11641--11661},
  year={2024}
}

@misc{ren2024analy,
      title={Analyzing and Reducing Catastrophic Forgetting in Parameter Efficient Tuning}, 
      author={Weijieying Ren and Xinlong Li and Lei Wang and Tianxiang Zhao and Wei Qin},
      year={2024},
      eprint={2402.18865},
      archivePrefix={arXiv},
      primaryClass={cs.LG},
      url={https://arxiv.org/abs/2402.18865}, 
}

@inproceedings{lu2025controlled,
  title={Controlled low-rank adaptation with subspace regularization for continued training on large language models},
  author={Lu, Yuheng and Qian, Bingshuo and Yuan, Caixia and Jiang, Huixing and Wang, Xiaojie},
  booktitle={Proceedings of the 63rd Annual Meeting of the Association for Computational Linguistics (Volume 1: Long Papers)},
  pages={19165--19181},
  year={2025}
}

@inproceedings{sow2025dynamic,
  title={Dynamic Loss-Based Sample Reweighting for Improved Large Language Model Pretraining},
  author={Sow, Daouda and Woisetschl{\"a}ger, Herbert and Bulusu, Sai Kiran and Wang, Shiqiang and Jacobsen, Hans-arno and Liang, Yingbin},
  booktitle={International Conference on Learning Representations},
  year={2025}
}

@misc{chen2025vul,
      title={Vulnerability-Aware Alignment: Mitigating Uneven Forgetting in Harmful Fine-Tuning}, 
      author={Liang Chen and Xueting Han and Li Shen and Jing Bai and Kam-Fai Wong},
      year={2025},
      eprint={2506.03850},
      archivePrefix={arXiv},
      primaryClass={cs.LG},
      url={https://arxiv.org/abs/2506.03850}, 
}

@misc{yang2025data,
      title={Data Mixing Agent: Learning to Re-weight Domains for Continual Pre-training}, 
      author={Kailai Yang and Xiao Liu and Lei Ji and Hao Li and Yeyun Gong and Peng Cheng and Mao Yang},
      year={2025},
      eprint={2507.15640},
      archivePrefix={arXiv},
      primaryClass={cs.LG},
      url={https://arxiv.org/abs/2507.15640}, 
}

@article{liu2024selectit,
  title={Selectit: Selective instruction tuning for llms via uncertainty-aware self-reflection},
  author={Liu, Liangxin and Liu, Xuebo and Wong, Derek F and Li, Dongfang and Wang, Ziyi and Hu, Baotian and Zhang, Min},
  journal={Advances in Neural Information Processing Systems},
  volume={37},
  pages={97800--97825},
  year={2024}
}

@inproceedings{li2024selective,
  title={Selective reflection-tuning: Student-selected data recycling for llm instruction-tuning},
  author={Li, Ming and Chen, Lichang and Chen, Jiuhai and He, Shwai and Gu, Jiuxiang and Zhou, Tianyi},
  booktitle={Findings of the Association for Computational Linguistics ACL 2024},
  pages={16189--16211},
  year={2024}
}

@inproceedings{li2025scar,
  title={Scar: Data selection via style consistency-aware response ranking for efficient instruction-tuning of large language models},
  author={Li, Zhuang and Hua, Yuncheng and Vu, Thuy and Zhan, Haolan and Qu, Lizhen and Haffari, Gholamreza},
  booktitle={Proceedings of the 63rd Annual Meeting of the Association for Computational Linguistics (Volume 1: Long Papers)},
  pages={12756--12790},
  year={2025}
}

@misc{shen2024rethink,
      title={Rethinking Data Selection for Supervised Fine-Tuning}, 
      author={Ming Shen},
      year={2024},
      eprint={2402.06094},
      archivePrefix={arXiv},
      primaryClass={cs.CL},
      url={https://arxiv.org/abs/2402.06094}, 
}

@inproceedings{li2024quantity,
  title={From quantity to quality: Boosting llm performance with self-guided data selection for instruction tuning},
  author={Li, Ming and Zhang, Yong and Li, Zhitao and Chen, Jiuhai and Chen, Lichang and Cheng, Ning and Wang, Jianzong and Zhou, Tianyi and Xiao, Jing},
  booktitle={Proceedings of the 2024 Conference of the North American Chapter of the Association for Computational Linguistics: Human Language Technologies (Volume 1: Long Papers)},
  pages={7602--7635},
  year={2024}
}

@inproceedings{xia2024less,
  title={LESS: selecting influential data for targeted instruction tuning},
  author={Xia, Mengzhou and Malladi, Sadhika and Gururangan, Suchin and Arora, Sanjeev and Chen, Danqi},
  booktitle={Proceedings of the 41st International Conference on Machine Learning},
  pages={54104--54132},
  year={2024}
}

@inproceedings{kang2024get,
  title={Get more for less: Principled Data Selection for Warming Up Fine-Tuning in {LLM}s},
  author={Feiyang Kang and Hoang Anh Just and Yifan Sun and Himanshu Jahagirdar and Yuanzhi Zhang and Rongxing Du and Anit Kumar Sahu and Ruoxi Jia},
  booktitle={The Twelfth International Conference on Learning Representations},
  year={2024}
}

@inproceedings{bukharin2024data,
  title={Data diversity matters for robust instruction tuning},
  author={Bukharin, Alexander and Li, Shiyang and Wang, Zhengyang and Yang, Jingfeng and Yin, Bing and Li, Xian and Zhang, Chao and Zhao, Tuo and Jiang, Haoming},
  booktitle={Findings of the Association for Computational Linguistics: EMNLP 2024},
  pages={3411--3425},
  year={2024}
}

@inproceedings{yang2025measuring,
  title={Measuring data diversity for instruction tuning: A systematic analysis and a reliable metric},
  author={Yang, Yuming and Nan, Yang and Ye, Junjie and Dou, Shihan and Wang, Xiao and Li, Shuo and Lv, Huijie and Gui, Tao and Zhang, Qi and Huang, Xuan-Jing},
  booktitle={Proceedings of the 63rd Annual Meeting of the Association for Computational Linguistics (Volume 1: Long Papers)},
  pages={18530--18549},
  year={2025}
}

@inproceedings{chen2025mig,
  title={Mig: Automatic data selection for instruction tuning by maximizing information gain in semantic space},
  author={Chen, Yicheng and Li, Yining and Hu, Kai and Zerun, Ma and HaochenYe, HaochenYe and Chen, Kai},
  booktitle={Findings of the Association for Computational Linguistics: ACL 2025},
  pages={9902--9915},
  year={2025}
}

@inproceedings{zhang2025d3diversity,
  title={D3: diversity, difficulty, and dependability-aware data selection for sample-efficient llm instruction tuning},
  author={Zhang, Jia and Zhang, Chen-Xi and Liu, Yao and Jin, Yi-Xuan and Yang, Xiao-Wen and Zheng, Bo and Liu, Yi and Guo, Lan-Zhe},
  booktitle={Proceedings of the Thirty-Fourth International Joint Conference on Artificial Intelligence},
  pages={8348--8356},
  year={2025}
}

@inproceedings{li2025selection,
  title={Selection of LLM Fine-Tuning Data Based on Orthogonal Rules},
  author={Li, Xiaomin and Gao, Mingye and Zhang, Zhiwei and Yue, Chang and Hu, Hong},
  booktitle={Proceedings of the AAAI Conference on Artificial Intelligence},
  volume={40},
  number={37},
  pages={31760--31768},
  year={2026}
}

@inproceedings{zhao2025beyond,
  title={Beyond similarity: A gradient-based graph method for instruction tuning data selection},
  author={Zhao, Yang and Du, Li and Ding, Xiao and Ouyang, Yangou and Wang, Hepeng and Xiong, Kai and Gao, Jinglong and Sun, Zhouhao and Xu, Dongliang and Yang, Qing and others},
  booktitle={Proceedings of the 63rd Annual Meeting of the Association for Computational Linguistics (Volume 1: Long Papers)},
  pages={24391--24404},
  year={2025}
}

@inproceedings{kulesza2011k,
  title={k-dpps: Fixed-size determinantal point processes},
  author={Kulesza, Alex and Taskar, Ben},
  booktitle={Proceedings of the 28th International Conference on Machine Learning (ICML-11)},
  pages={1193--1200},
  year={2011}
}

@book{horn2012matrix,
  title={Matrix analysis},
  author={Horn, Roger A and Johnson, Charles R},
  year={2012},
  publisher={Cambridge university press}
}

@article{blumensath2009iterative,
  title={Iterative hard thresholding for compressed sensing},
  author={Blumensath, Thomas and Davies, Mike E},
  journal={Applied and computational harmonic analysis},
  volume={27},
  number={3},
  pages={265--274},
  year={2009},
  publisher={Elsevier}
}

@article{johnson1984extensions,
  title={Extensions of Lipschitz mappings into a Hilbert space},
  author={Johnson, William B and Lindenstrauss, Joram and others},
  journal={Contemporary mathematics},
  volume={26},
  number={189-206},
  pages={1},
  year={1984}
}

@inproceedings{hu2022lora,
    title={LoRA: Low-Rank Adaptation of Large Language Models},
    author={Edward J Hu and yelong shen and Phillip Wallis and Zeyuan Allen-Zhu and Yuanzhi Li and Shean Wang and Lu Wang and Weizhu Chen},
    booktitle={International Conference on Learning Representations},
    year={2022},
}

@inproceedings{chang2026spice,
title={{SPICE}: Submodular Penalized Information{\textendash}Conflict Selection for Efficient Large Language Model Training},
author={Powei Chang and Jinpeng Zhang and Bowen Chen and Chenyu Wang and Chenlu Guo and Yixing Zhang and Yukang Gao and JianXiang Xiang and Yue Gao and Chaoqun Sun and Yiyi Chen and Dongying Kong},
booktitle={The Fourteenth International Conference on Learning Representations},
year={2026},
url={https://openreview.net/forum?id=9rCRy58TPF}
}

@inproceedings{wu2025mitigating,
    title={Mitigating Forgetting in {LLM} Fine-Tuning via Low-Perplexity Token Learning},
    author={Chao-Chung Wu and Zhi Rui Tam and Chieh-Yen Lin and Yun-Nung Chen and Shao-Hua Sun and Hung-yi Lee},
    booktitle={The Thirty-ninth Annual Conference on Neural Information Processing Systems},
    year={2025},
}

@misc{gemmateam2025gemma3technicalreport,
      title={Gemma 3 Technical Report}, 
      author={Gemma Team and Aishwarya Kamath and Johan Ferret and Shreya Pathak and Nino Vieillard and Ramona Merhej et al.},
      year={2025},
      eprint={2503.19786},
      archivePrefix={arXiv},
      primaryClass={cs.CL},
      url={https://arxiv.org/abs/2503.19786}, 
}

@misc{grattafiori2024llama3herdmodels,
      title={The Llama 3 Herd of Models}, 
      author={Aaron Grattafiori and Abhimanyu Dubey and Abhinav Jauhri and Abhinav Pandey and Abhishek Kadian et al.},
      year={2024},
      eprint={2407.21783},
      archivePrefix={arXiv},
      primaryClass={cs.AI},
      url={https://arxiv.org/abs/2407.21783}, 
}

@misc{qwen3.5,
    title  = {Qwen3.5: Towards Native Multimodal Agents},
    author = {{Qwen Team}},
    month  = {February},
    year   = {2026},
    url    = {https://qwen.ai/blog?id=qwen3.5}
}

@inproceedings{hendrycks2021measuring,
    title={Measuring Massive Multitask Language Understanding},
    author={Dan Hendrycks and Collin Burns and Steven Basart and Andy Zou and Mantas Mazeika and Dawn Song and Jacob Steinhardt},
    booktitle={International Conference on Learning Representations},
    year={2021},
}

@misc{chen2021evaluatinglargelanguagemodels,
      title={Evaluating Large Language Models Trained on Code}, 
      author={Mark Chen and Jerry Tworek and Heewoo Jun and Qiming Yuan and Henrique Ponde de Oliveira Pinto and Jared Kaplan et al.},
      year={2021},
      eprint={2107.03374},
      archivePrefix={arXiv},
      primaryClass={cs.LG},
      url={https://arxiv.org/abs/2107.03374}, 
}

@inproceedings{zhou2025davir,
  title={Davir: Data selection via implicit reward for large language models},
  author={Zhou, Haotian and Liu, Tingkai and Ma, Qianli and Zhang, Yufeng and Yuan, Jianbo and Liu, Pengfei and You, Yang and Yang, Hongxia},
  booktitle={Proceedings of the 63rd Annual Meeting of the Association for Computational Linguistics (Volume 1: Long Papers)},
  pages={9220--9237},
  year={2025}
}

@inproceedings{li2026tokenlevel,
    title={Token-level Data Selection for Safe {LLM} Fine-tuning},
    author={Yanping Li and Zhening Liu and Zijian Li and Zehong Lin and Jun Zhang},
    booktitle={The Fourteenth International Conference on Learning Representations},
    year={2026},
}

@misc{cobbe2021training,
      title={Training Verifiers to Solve Math Word Problems}, 
      author={Karl Cobbe and Vineet Kosaraju and Mohammad Bavarian and Mark Chen and Heewoo Jun and Lukasz Kaiser and Matthias Plappert and Jerry Tworek and Jacob Hilton and Reiichiro Nakano and Christopher Hesse and John Schulman},
      year={2021},
      eprint={2110.14168},
      archivePrefix={arXiv},
      primaryClass={cs.LG},
      url={https://arxiv.org/abs/2110.14168}, 
}

@InProceedings{pmlr-v235-wei24h,
  title = 	 {Magicoder: Empowering Code Generation with {OSS}-Instruct},
  author =       {Wei, Yuxiang and Wang, Zhe and Liu, Jiawei and Ding, Yifeng and Zhang, Lingming},
  booktitle = 	 {Proceedings of the 41st International Conference on Machine Learning},
  pages = 	 {52632--52657},
  year = 	 {2024},
  volume = 	 {235},
  series = 	 {Proceedings of Machine Learning Research},
  month = 	 {21--27 Jul},
  publisher =    {PMLR},
}

@inproceedings{zellers2019hellaswag,
  title={Hellaswag: Can a machine really finish your sentence?},
  author={Zellers, Rowan and Holtzman, Ari and Bisk, Yonatan and Farhadi, Ali and Choi, Yejin},
  booktitle={Proceedings of the 57th annual meeting of the association for computational linguistics},
  pages={4791--4800},
  year={2019}
}

@inproceedings{mihaylov2018can,
  title={Can a suit of armor conduct electricity? a new dataset for open book question answering},
  author={Mihaylov, Todor and Clark, Peter and Khot, Tushar and Sabharwal, Ashish},
  booktitle={Proceedings of the 2018 conference on empirical methods in natural language processing},
  pages={2381--2391},
  year={2018}
}

@inproceedings{m2024leitner,
  title={Leitner-guided memory replay for cross-lingual continual learning},
  author={M’hamdi, Meryem and May, Jonathan},
  booktitle={Proceedings of the 2024 Conference of the North American Chapter of the Association for Computational Linguistics: Human Language Technologies (Volume 1: Long Papers)},
  pages={7808--7821},
  year={2024}
}

@inproceedings{bai2025efficient,
  title={An efficient rehearsal scheme for catastrophic forgetting mitigation during multi-stage fine-tuning},
  author={Bai, Andrew and Yeh, Chih-Kuan and Hsieh, Cho-Jui and Taly, Ankur},
  booktitle={Findings of the Association for Computational Linguistics: NAACL 2025},
  pages={2557--2569},
  year={2025}
}

@inproceedings{xu2025magpie,
    title={Magpie: Alignment Data Synthesis from Scratch by Prompting Aligned {LLM}s with Nothing},
    author={Zhangchen Xu and Fengqing Jiang and Luyao Niu and Yuntian Deng and Radha Poovendran and Yejin Choi and Bill Yuchen Lin},
    booktitle={The Thirteenth International Conference on Learning Representations},
    year={2025},
}

@inproceedings{zhou2025pay,
    title={Pay Attention to Small Weights},
    author={Chao Zhou and Tom Jacobs and Advait Gadhikar and Rebekka Burkholz},
    booktitle={The Thirty-ninth Annual Conference on Neural Information Processing Systems},
    year={2025},
}

@misc{eval-harness,
  author       = {Gao, Leo and Tow, Jonathan and Abbasi, Baber and Biderman, Stella and Black, Sid and DiPofi, Anthony and Foster, Charles and Golding, Laurence and Hsu, Jeffrey and Le Noac'h, Alain and Li, Haonan and McDonell, Kyle and Muennighoff, Niklas and Ociepa, Chris and Phang, Jason and Reynolds, Laria and Schoelkopf, Hailey and Skowron, Aviya and Sutawika, Lintang and Tang, Eric and Thite, Anish and Wang, Ben and Wang, Kevin and Zou, Andy},
  title        = {The Language Model Evaluation Harness},
  month        = 07,
  year         = 2024,
  publisher    = {Zenodo},
  version      = {v0.4.3},
  doi          = {10.5281/zenodo.12608602},
  url          = {https://zenodo.org/records/12608602}
}

@inproceedings{lin2026sft,
    title={{SFT} Doesn{\textquoteright}t Always Hurt General Capabilities: Revisiting Domain-Specific Fine-Tuning in {LLM}s},
    author={Jiacheng Lin and Zhongruo Wang and Kun Qian and Tian Wang and Arvind Srinivasan and Hansi Zeng and Ruochen Jiao and Xie Zhou and Jiri Gesi and Dakuo Wang and Yufan Guo and Kai Zhong and Weiqi Zhang and sujay sanghavi and Changyou Chen and Hyokun Yun and Lihong Li},
    booktitle={The Fourteenth International Conference on Learning Representations},
    year={2026},
}

@inproceedings{deb2025fishersft,
    title={Fisher{SFT}: Data-Efficient Supervised Fine-Tuning of Language Models Using Information Gain},
    author={Rohan Deb and Kiran Koshy Thekumparampil and Kousha Kalantari and Gaurush Hiranandani and Shoham Sabach and Branislav Kveton},
    booktitle={Forty-second International Conference on Machine Learning},
    year={2025},
}

@article{van2008visualizing,
  title={Visualizing data using t-SNE.},
  author={Van der Maaten, Laurens and Hinton, Geoffrey},
  journal={Journal of machine learning research},
  volume={9},
  number={11},
  year={2008}
}

@inproceedings{yu2024metamath,
title={MetaMath: Bootstrap Your Own Mathematical Questions for Large Language Models},
author={Longhui Yu and Weisen Jiang and Han Shi and Jincheng YU and Zhengying Liu and Yu Zhang and James Kwok and Zhenguo Li and Adrian Weller and Weiyang Liu},
booktitle={The Twelfth International Conference on Learning Representations},
year={2024},
}

@inproceedings{wang2025on,
title={On the Loss of Context Awareness in General Instruction Fine-tuning},
author={Yihan Wang and Andrew Bai and Nanyun Peng and Cho-Jui Hsieh},
booktitle={The Thirty-ninth Annual Conference on Neural Information Processing Systems},
year={2025},
}

@inproceedings{shen2025dont,
  title={Don’t Make It Up: Preserving Ignorance Awareness in LLM Fine-Tuning},
  author={Shen, William F and Qiu, Xinchi and Cancedda, Nicola and Lane, Nicholas D},
  booktitle={NeurIPS 2025 Workshop: Reliable ML from Unreliable Data},
year={2025},
}

@inproceedings{guan2025survey,
  title={A Survey on Personalized Alignment—The Missing Piece for Large Language Models in Real-World Applications},
  author={Guan, Jian and Wu, Junfei and Li, Jia-Nan and Cheng, Chuanqi and Wu, Wei},
  booktitle={Findings of the Association for Computational Linguistics: ACL 2025},
  pages={5313--5333},
  year={2025}
}

@inproceedings{ke2025adaptation,
  title={Adaptation of large language models},
  author={Ke, Zixuan and Ming, Yifei and Joty, Shafiq},
  booktitle={Proceedings of the 2025 Annual Conference of the Nations of the Americas Chapter of the Association for Computational Linguistics: Human Language Technologies (Volume 5: Tutorial Abstracts)},
  pages={30--37},
  year={2025}
}

@inproceedings{hui2025hft,
  title={Hft: Half fine-tuning for large language models},
  author={Hui, Tingfeng and Zhang, Zhenyu and Wang, Shuohuan and Xu, Weiran and Sun, Yu and Wu, Hua},
  booktitle={Proceedings of the 63rd Annual Meeting of the Association for Computational Linguistics (Volume 1: Long Papers)},
  pages={12791--12819},
  year={2025}
}

@article{lu2025fine,
  title={Fine-tuning large language models for domain adaptation: Exploration of training strategies, scaling, model merging and synergistic capabilities},
  author={Lu, Wei and Luu, Rachel K and Buehler, Markus J},
  journal={npj Computational Materials},
  volume={11},
  number={1},
  pages={84},
  year={2025},
  publisher={Nature Publishing Group UK London}
}

@article{liu2026learn,
  title={Learn More, Forget Less: A Gradient-Aware Data Selection Approach for LLM},
  author={Liu, Zeming and Liu, Yibai and Wang, Shihang and Song, Zheming and Wang, Junzhe and Liu, Jingjing and Liu, Qingjie and Chen, Guangxu and Wang, Yunhong},
  journal={Signal Processing},
  pages={110611},
  year={2026},
  publisher={Elsevier}
}

@inproceedings{yuan2026differential,
title={Differential Fine-Tuning Large Language Models Towards Better Diverse Reasoning Abilities},
author={Xiaosong Yuan and Chen Shen and Shaotian Yan and kaiyuan liu and Xiaofeng Zhang and Sinan Fan and Liang Xie and Wenxiao Wang and Renchu Guan and Ying Wang and Jieping Ye},
booktitle={The Fourteenth International Conference on Learning Representations},
year={2026},
}

@article{hemmi2022lazy,
  title={Lazy and fast greedy map inference for determinantal point process},
  author={Hemmi, Shinichi and Oki, Taihei and Sakaue, Shinsaku and Fujii, Kaito and Iwata, Satoru},
  journal={Advances in Neural Information Processing Systems},
  volume={35},
  pages={2776--2789},
  year={2022}
}

@article{chen2018fast,
  title={Fast greedy map inference for determinantal point process to improve recommendation diversity},
  author={Chen, Laming and Zhang, Guoxin and Zhou, Eric},
  journal={Advances in neural information processing systems},
  volume={31},
  year={2018}
}

@article{belhadji2020determinantal,
  title={A determinantal point process for column subset selection},
  author={Belhadji, Ayoub and Bardenet, R{\'e}mi and Chainais, Pierre},
  journal={Journal of machine learning research},
  volume={21},
  number={197},
  pages={1--62},
  year={2020}
}

@inproceedings{jain2026train,
    title={Train on Validation (ToV): Fast data selection with applications to fine-tuning},
    author={Ayush Jain and Andrea Montanari and Eren Sasoglu},
    booktitle={The Fourteenth International Conference on Learning Representations},
    year={2026},
}

@misc{taheri2025forgettingnewmechanismbetter,
      title={Forgetting: A New Mechanism Towards Better Large Language Model Fine-tuning}, 
      author={Ali Taheri and Alireza Taban and Qizhou Wang and Shanshan Ye and Abdolreza Mirzaei and Tongliang Liu and Bo Han},
      year={2026},
      eprint={2508.04329},
      archivePrefix={arXiv},
      primaryClass={cs.LG},
      url={https://arxiv.org/abs/2508.04329}, 
}

@inproceedings{xu2024wizardlm,
title={Wizard{LM}: Empowering Large Pre-Trained Language Models to Follow Complex Instructions},
author={Can Xu and Qingfeng Sun and Kai Zheng and Xiubo Geng and Pu Zhao and Jiazhan Feng and Chongyang Tao and Qingwei Lin and Daxin Jiang},
booktitle={The Twelfth International Conference on Learning Representations},
year={2024},
}

@inproceedings{ye2024distilling,
  title={Distilling causal effect of data in continual few-shot relation learning},
  author={Ye, Weihang and Zhang, Peng and Zhang, Jing and Gao, Hui and Wang, Moyao},
  booktitle={Proceedings of the 2024 Joint International Conference on Computational Linguistics, Language Resources and Evaluation (LREC-COLING 2024)},
  pages={5041--5051},
  year={2024}
}

@inproceedings{he2024seekr,
  title={Seekr: Selective attention-guided knowledge retention for continual learning of large language models},
  author={He, Jinghan and Guo, Haiyun and Zhu, Kuan and Zhao, Zihan and Tang, Ming and Wang, Jinqiao},
  booktitle={Proceedings of the 2024 Conference on Empirical Methods in Natural Language Processing},
  pages={3254--3266},
  year={2024}
}

@misc{he2025earlystopp,
      title={Early-stopping for Transformer model training}, 
      author={Jing He and Hua Jiang and Cheng Li and Siqian Xin and Shuzhen Yang},
      year={2025},
      eprint={2510.16074},
      archivePrefix={arXiv},
      primaryClass={cs.LG},
      url={https://arxiv.org/abs/2510.16074}, 
}

@misc{zhang2026training,
      title={Training Data Selection with Gradient Orthogonality for Efficient Domain Adaptation}, 
      author={Xiyang Zhang and Yuanhe Tian and Hongzhi Wang and Yan Song},
      year={2026},
      eprint={2602.06359},
      archivePrefix={arXiv},
      primaryClass={cs.LG},
      url={https://arxiv.org/abs/2602.06359}, 
}

@inproceedings{chen2026posttraining,
title={Post-training Large Language Models for Diverse High-Quality Responses},
author={Yilei Chen and Souradip Chakraborty and Lorenz Wolf and Ioannis Paschalidis and Aldo Pacchiano},
booktitle={The Fourteenth International Conference on Learning Representations},
year={2026},
}

@misc{clark2018think,
    title         = {Think you have Solved Question Answering? Try ARC, the AI2 Reasoning Challenge},
    author        = {Peter Clark and Isaac Cowhey and Oren Etzioni and Tushar Khot and Ashish Sabharwal and Carissa Schoenick and Oyvind Tafjord},
    year          = {2018},
    eprint        = {1803.05457},
    archiveprefix = {arXiv},
    primaryclass  = {cs.AI},
    url           = {https://arxiv.org/abs/1803.05457},
}

@inproceedings{bisk2020piqa,
    title     = {PIQA: Reasoning about physical commonsense in natural language},
    author    = {Bisk, Yonatan and Zellers, Rowan and Le Bras, Ronan and Gao, Jianfeng and Choi, Yejin},
    booktitle = {Proceedings of the AAAI conference on artificial intelligence},
    volume    = {34},
    pages     = {7432--7439},
    year      = {2020},
}

@inproceedings{sakaguchi2020winogrande,
    title     = {WinoGrande: An Adversarial Winograd Schema Challenge at Scale},
    author    = {Sakaguchi, Keisuke and Le Bras, Ronan and Bhagavatula, Chandra and Choi, Yejin},
    booktitle = {Proceedings of the AAAI Conference on Artificial Intelligence},
    volume    = {34},
    pages     = {8732--8740},
    year      = {2020},
}

@inproceedings{lin2022truthfulqa,
    title     = {Truthfulqa: Measuring how models mimic human falsehoods},
    author    = {Lin, Stephanie and Hilton, Jacob and Evans, Owain},
    booktitle = {Proceedings of the 60th annual meeting of the association for computational linguistics (volume 1: long papers)},
    pages     = {3214--3252},
    year      = {2022},
}

@inproceedings{li2025ruler,
    title     = {{R}ule{R}: Improving llm controllability by rule-based data recycling},
    author    = {Li, Ming and Chen, Han and Wang, Chenguang and Nguyen, Dang and Li, Dianqi and Zhou, Tianyi},
    booktitle = {Proceedings of the 2025 Conference of the Nations of the Americas Chapter of the Association for Computational Linguistics: Human Language Technologies (Volume 2: Short Papers)},
    pages     = {926--943},
    year      = {2025},
}

@inproceedings{han2026makes,
    title     = {What Makes Good Instruction-Tuning Data? An In-Context Learning Perspective},
    author    = {Han, Guangzeng  and},
    booktitle = {Proceedings of the 64th Annual Meeting of the {A}ssociation for {C}omputational {L}inguistics (Volume 1: Long Papers)},
    month     = {jul},
    year      = {2026},
    publisher = {Association for Computational Linguistics},
    pages     = {1013--1027},
}

@inproceedings{mirza-etal-2025-stratified,
    title     = {Stratified Selective Sampling for Instruction Tuning with Dedicated Scoring Strategy},
    author    = {Mirza, Paramita and Weber, Lucas and K{\"u}ch, Fabian},
    booktitle = {Findings of the Association for Computational Linguistics: EMNLP 2025},
    pages     = {19949--19974},
    year      = {2025},
}

@misc{zhou2023instruction,
    title         = {Instruction-Following Evaluation for Large Language Models},
    author        = {Jeffrey Zhou and Tianjian Lu and Swaroop Mishra and Siddhartha Brahma and Sujoy Basu and Yi Luan and Denny Zhou and Le Hou},
    year          = {2023},
    eprint        = {2311.07911},
    archiveprefix = {arXiv},
    primaryclass  = {cs.CL},
    url           = {https://arxiv.org/abs/2311.07911},
}

@misc{peng2023instruction,
    title         = {Instruction Tuning with GPT-4},
    author        = {Baolin Peng and Chunyuan Li and Pengcheng He and Michel Galley and Jianfeng Gao},
    year          = {2023},
    eprint        = {2304.03277},
    archiveprefix = {arXiv},
    primaryclass  = {cs.CL},
    url           = {https://arxiv.org/abs/2304.03277},
}

@misc{chen2026algorithmgoalsdualscoring,
      title={One Algorithm, Two Goals: Dual Scoring for Parameter and Data Selection in LLM Fine-Tuning}, 
      author={Xinrui Chen and Liu Yang and Ou Wu},
      year={2026},
      eprint={2605.06166},
      archivePrefix={arXiv},
      primaryClass={cs.LG},
      url={https://arxiv.org/abs/2605.06166}, 
}

@misc{qing2026alphatokendecouplingadaptationstability,
      title={AlphaToken: Decoupling Adaptation and Stability for Path-Aware Response Token Valuation in LLM Post-Training}, 
      author={Liu Qing and Ou Wu and Yi Du},
      year={2026},
      eprint={2606.01635},
      archivePrefix={arXiv},
      primaryClass={cs.CL},
      url={https://arxiv.org/abs/2606.01635}, 
}

@inproceedings{NEURIPS2024_3322a9a7,
 author = {Lin, Zhenghao and Gou, Zhibin and Gong, Yeyun and Liu, Xiao and Shen, Yelong and Xu, Ruochen and Lin, Chen and Yang, Yujiu and Jiao, Jian and Duan, Nan and Chen, Weizhu},
 booktitle = {Advances in Neural Information Processing Systems},
 pages = {29029--29063},
 title = {Not All Tokens Are What You Need for Pretraining},
 volume = {37},
 year = {2024}
}

@inproceedings{NEURIPS2023_dcba6be9,
 author = {Xie, Sang Michael and Pham, Hieu and Dong, Xuanyi and Du, Nan and Liu, Hanxiao and Lu, Yifeng and Liang, Percy S and Le, Quoc V and Ma, Tengyu and Yu, Adams Wei},
 booktitle = {Advances in Neural Information Processing Systems},
 pages = {69798--69818},
 title = {DoReMi: Optimizing Data Mixtures Speeds Up Language Model Pretraining},
 volume = {36},
 year = {2023}
}
}


\newpage
\appendix

\newpage
\section{Comparison with Existing Model-Centric Data Selection Methods}
\label{appendix_method_discussion}

In this section, we clarify the key distinction between the navigator model in NADS and the reference models adopted in prior model-centric data selection methods. Existing methods have shown that model differences can be used to construct informative selection signals. DavIR~\citep{zhou2025davir} scores samples by comparing the loss difference between the pretrained model and a reference model fine tuned on the full training set, and treats this score as an implicit reward. The resulting signal mainly reflects how much a sample contributes to target task adaptation. ToV~\citep{jain2026train} updates the model using a small validation set from the target distribution, and then scores samples according to the loss change before and after this update. Its reference model therefore also serves target domain utility estimation. TOSS~\citep{li2026tokenlevel} trains a safety-degraded model and a utility-oriented model on harmful reference data and high quality reference data, respectively, and identifies unsafe tokens through their loss difference. Forgetting~\citep{taheri2025forgettingnewmechanismbetter} further fine tunes the base model on a sampled reference subset, and uses the resulting token level influence score to partition tokens for subsequent learning or forgetting. These methods use reference model differences for their respective selection or partitioning objectives.

The navigator in NADS is designed for a different role. It is obtained by applying the same target task adaptation recipe used in the final fine-tuning stage, but without retention regularization or constraint samples. The resulting navigator provides a recipe-conditioned estimate of forgetting risk for candidate constraint samples by comparing its predictions with those of the pretrained model. A larger predictive divergence indicates that the pretrained behavior on the sample is more likely to be altered under unconstrained target task adaptation. NADS therefore selects samples with higher estimated forgetting risk as preservation samples for the final fine-tuning stage.

\section{Theoretical Analysis}
\label{Theoretical_Analysis}
\subsection{Proof of Proposition 1}
\label{Proposition_1}

In this section, we prove that the checkpoint selected by validation-based early stopping is near-best on the unconstrained adaptation trajectory.

\textit{\textbf{Proof.}}
Let $\{\bm{\theta}_t\}_{t=0}^{T}$ denote the unconstrained adaptation trajectory induced by the target recipe. Along this trajectory, define $R(t)\triangleq \mathcal{L}_{\text{pop}}(\bm{\theta}_t)$ and $\hat R(t)\triangleq \mathcal{L}_{\text{val}}(\bm{\theta}_t)$, where $R(t)$ is the population new-task risk at checkpoint $\bm{\theta}_t$, and $\hat R(t)$ is the empirical validation loss on the held-out validation set. Let
\begin{equation}
t_{\text{opt}} \in \operatornamewithlimits{arg\,min}_{t\in\{0,\dots,T\}} R(t),
\qquad
t^* \in \operatornamewithlimits{arg\,min}_{t\in\{0,\dots,T\}} \hat R(t),
\label{eq:prop1_opt_def}
\end{equation}
where $t_{\text{opt}}$ denotes the population-risk minimizer along the unconstrained adaptation trajectory, and $t^*$ denotes the checkpoint selected by validation-based early stopping. By the definition of $t^*$, we have $\hat R(t^*)\le \hat R(t)$ for all $t\in\{0,\dots,T\}$. In particular, taking $t=t_{\text{opt}}$ gives
\begin{equation}
\hat R(t^*)\le \hat R(t_{\text{opt}}).
\label{eq:prop1_val_compare}
\end{equation}
By assumption, there exists $\varepsilon>0$ such that $|\hat R(t)-R(t)|\le \varepsilon$ holds for all $t\in\{0,\dots,T\}$. Equivalently,
\begin{equation}
R(t)-\varepsilon \le \hat R(t)\le R(t)+\varepsilon, \qquad \forall t\in\{0,\dots,T\}.
\label{eq:prop1_uniform_gap_eqv}
\end{equation}
Applying the left inequality in Eq.~\eqref{eq:prop1_uniform_gap_eqv} at $t=t^*$ and the right inequality at $t=t_{\text{opt}}$, and combining them with Eq.~\eqref{eq:prop1_val_compare}, we obtain
\begin{equation}
R(t^*)-\varepsilon
\le
\hat R(t^*)
\le
\hat R(t_{\text{opt}})
\le
R(t_{\text{opt}})+\varepsilon.
\label{eq:prop1_chain}
\end{equation}
Rearranging Eq.~\eqref{eq:prop1_chain} yields $R(t^*)-R(t_{\text{opt}})\le 2\varepsilon$. Therefore, the validation-selected checkpoint has population new-task risk within $2\varepsilon$ of the best checkpoint on the unconstrained adaptation trajectory.

\subsection{Proof of Proposition 2}
\label{Proposition_2}

In this section, we prove the local deviation recursion in Proposition~\ref{prop:anchoring_navigator}, showing that bounded preservation updates control the mismatch between the training trajectory of $M_{\text{final}}$ and the unconstrained target-task trajectory over local horizons.

\textit{\textbf{Proof.}}
Let $d_t:=\|\bm{\theta}_t-\bar{\bm{\theta}}_t\|$ denote the deviation between the constrained trajectory $\{\bm{\theta}_t\}_{t=0}^T$ and the unconstrained trajectory $\{\bar{\bm{\theta}}_t\}_{t=0}^T$ at step $t$. By the update rules
$\bm{\theta}_{t+1}=\bm{\theta}_t-\eta\bigl(\nabla L_{\text{task}}(\bm{\theta}_t)+\xi_t\bigr)$
and
$\bar{\bm{\theta}}_{t+1}=\bar{\bm{\theta}}_t-\eta\nabla L_{\text{task}}(\bar{\bm{\theta}}_t)$,
their difference satisfies
\begin{equation}
\bm{\theta}_{t+1}-\bar{\bm{\theta}}_{t+1}
=
\bm{\theta}_t-\bar{\bm{\theta}}_t
-\eta\bigl(\nabla L_{\text{task}}(\bm{\theta}_t)-\nabla L_{\text{task}}(\bar{\bm{\theta}}_t)\bigr)
-\eta\xi_t.
\end{equation}
Taking norms and applying the triangle inequality yields
\begin{equation}
d_{t+1}
\le
d_t+\eta\|\nabla L_{\text{task}}(\bm{\theta}_t)-\nabla L_{\text{task}}(\bar{\bm{\theta}}_t)\|+\eta\|\xi_t\|.
\end{equation}
By the $L$-Lipschitz continuity of $\nabla L_{\text{task}}$ and the bound $\|\xi_t\|\le\kappa$, we obtain $d_{t+1}\le (1+\eta L)d_t+\eta\kappa$.  Then, we fix a reference step $t$ and an integer local horizon $\tau\in\{0,\dots,T-t\}$. Iterating the one-step bound from step $t$ onward gives that, for any $s\in\{0,\dots,\tau\}$,
\begin{equation}
d_{t+s}
\le
d_t+\sum_{j=0}^{s-1}\eta\bigl(Ld_{t+j}+\kappa\bigr).
\end{equation}
Taking the supremum over $0\le s\le \tau$ on the left-hand side yields
\begin{equation}
\sup_{0\le s\le \tau}d_{t+s}
\le
d_t+\sum_{j=0}^{\tau-1}\eta\bigl(Ld_{t+j}+\kappa\bigr).
\end{equation}
Substituting back $d_t=\|\bm{\theta}_t-\bar{\bm{\theta}}_t\|$ proves
\begin{equation}
\sup_{0\le s\le \tau}\|\bm{\theta}_{t+s}-\bar{\bm{\theta}}_{t+s}\|
\le
\|\bm{\theta}_t-\bar{\bm{\theta}}_t\|
+
\sum_{j=0}^{\tau-1}\eta\bigl(L\|\bm{\theta}_{t+j}-\bar{\bm{\theta}}_{t+j}\|+\kappa\bigr).
\end{equation}
Therefore, bounded preservation updates locally control trajectory mismatch.

\subsection{Local perturbation of navigator-side forgetting-risk ordering}
\label{app:corollary_ranking_stability}

\begin{corollary}[Local perturbation of navigator-side forgetting-risk ordering]
Let $\Delta_\theta(x)$ denote the forgetting-risk score of sample $x$. Fix a reference step $t$ and an integer local horizon $\tau \in \{0,\dots,T-t\}$, and let $\mathcal N_{t,\tau}$ denote a neighborhood containing the navigator and final trajectories over this horizon. Assume that, for each sample $x$, $\Delta_\theta(x)$ is Lipschitz on $\mathcal N_{t,\tau}$ with local constant $G_{t,\tau}(x)$. Define the local trajectory mismatch $B_{t,\tau} \;:=\; \sup_{0\le s\le \tau}\|\bm{\theta}_{t+s}-\bar{\bm{\theta}}_{t+s}\|$. Then, for any two samples $x_a$ and $x_b$,
\[
\Big|
[\Delta_{\bm{\theta}_{t+\tau}}(x_a)-\Delta_{\bm{\theta}_{t+\tau}}(x_b)]
-
[\Delta_{\bar{\bm{\theta}}_{t+\tau}}(x_a)-\Delta_{\bar{\bm{\theta}}_{t+\tau}}(x_b)]
\Big|
\le
\bigl(G_{t,\tau}(x_a)+G_{t,\tau}(x_b)\bigr) B_{t,\tau}.
\]
Consequently, if $\big|
\Delta_{\bar{\bm{\theta}}_{t+\tau}}(x_a)-\Delta_{\bar{\bm{\theta}}_{t+\tau}}(x_b)
\big|
>
\bigl(G_{t,\tau}(x_a)+G_{t,\tau}(x_b)\bigr) B_{t,\tau}$, then the pairwise ordering between $x_a$ and $x_b$ is preserved at step $t+\tau$, i.e.,
\[
\operatorname{sign}\!\big(\Delta_{\bm{\theta}_{t+\tau}}(x_a)-\Delta_{\bm{\theta}_{t+\tau}}(x_b)\big)
=
\operatorname{sign}\!\big(\Delta_{\bar{\bm{\theta}}_{t+\tau}}(x_a)-\Delta_{\bar{\bm{\theta}}_{t+\tau}}(x_b)\big).
\]
\end{corollary}
Thus, this corollary is a conditional local perturbation statement. When the relevant score sensitivities are moderate and the local trajectory mismatch remains limited, pairwise orderings with sufficiently large navigator-side margins are preserved under the constrained final trajectory. Appendix~\ref{app:trajectory_ordering} provides an empirical analysis of this ordering fidelity by varying the KD strength and measuring the agreement between navigator-induced candidate rankings and final-trajectory reference rankings.

\subsection{Theoretical connection between the KL-based utility and the KD objective}
\label{app:utility_analysis}

In this section, we clarify a local relation between the KL-based utility in NADS and the subset-level KD discrepancy in final training. We analyze how the utility shares the same token-level discrepancy as the KD objective, with an entropy-based reweighting that yields a confidence-aware local alignment with the capability-preserving regularizer at the navigator checkpoint.

For a sample $x$, define the distillation discrepancy under model parameter $\bm{\theta}$ as
\begin{equation}
f_x(\bm{\theta})
:=
\frac{1}{|x|}\sum_{t=1}^{|x|}
D_{\text{KL}}\!\left(
P_{\text{pretrain}}(\cdot\mid x_{<t})
\,\big\|\,
P_{\bm{\theta}}(\cdot\mid x_{<t})
\right).
\label{eq:app_sample_kd}
\end{equation}
For a subset $S\subseteq \mathcal C$, define the corresponding set-level discrepancy as
\begin{equation}
F_S(\bm{\theta})
:=
\frac{1}{|S|}\sum_{x\in S} f_x(\bm{\theta}).
\label{eq:app_set_kd}
\end{equation}
Evaluating Eq.~\eqref{eq:app_set_kd} at the navigator checkpoint $\bm{\theta}_{\text{nav}}$ yields the average KD discrepancy between the pretrained model and the navigator model over subset $S$. The utility used in NADS is defined as
\begin{equation}
u(x)
=
\frac{1}{|x|}\sum_{t=1}^{|x|}
w_t(x)\,
D_{\text{KL}}\!\left(
P_{\text{pretrain}}(\cdot\mid x_{<t})
\,\big\|\,
P_{\text{nav}}(\cdot\mid x_{<t})
\right),
\label{eq:app_weighted_utility}
\end{equation}
where $w_t(x)=\exp\bigl(-\gamma H(P_{\text{pretrain}}(\cdot\mid x_{<t}))\bigr)$. We define the set-level averaged utility as
\begin{equation}
\bar U(S):=\frac{1}{|S|}\sum_{x\in S}u(x),
\qquad
U(S):=\sum_{x\in S}u(x).
\end{equation}
Let $
\delta_t(x)
:=
D_{\text{KL}}\!\left(
P_{\text{pretrain}}(\cdot\mid x_{<t})
\,\big\|\,
P_{\text{nav}}(\cdot\mid x_{<t})
\right)$ denote the token-level KD discrepancy at the navigator checkpoint. Then both $F_S(\bm{\theta}_{\text{nav}})$ and $\bar U(S)$ are constructed from the same set of token-level discrepancies $\{\delta_t(x)\}$:
\begin{equation}
F_S(\bm{\theta}_{\text{nav}})
=
\frac{1}{|S|}\sum_{x\in S}\frac{1}{|x|}\sum_{t=1}^{|x|}\delta_t(x), \qquad \bar U(S)
=
\frac{1}{|S|}\sum_{x\in S}\frac{1}{|x|}\sum_{t=1}^{|x|} w_t(x)\,\delta_t(x).
\end{equation}
Hence, the utility and the KD discrepancy share the same underlying signal, differing only in whether token-level contributions are reweighted.

Importantly, the entropy gate $w_t(x)=\exp(-\gamma H(\cdot))$ is monotone decreasing in the pretrained entropy. Thus, high-entropy positions are systematically down-weighted, whereas low-entropy positions contribute more strongly to the utility. This yields a confidence-aware reweighting of the KD discrepancy rather than an arbitrary transformation of the signal. Under this view, $\bar U(S)$ can be interpreted as a confidence-aware, robustness-oriented proxy of $F_S(\theta_{\text{nav}})$: it preserves the same discrepancy structure while reducing the influence of positions more likely to reflect intrinsic uncertainty or token-level noise.

This provides a local alignment rationale for the utility design in NADS. The final training objective reduces the KD discrepancy on the selected constraint set, while the selection utility evaluates the same navigator-side discrepancy with confidence-aware reweighting. Hence, the selection signal remains locally aligned with the KD objective while being less sensitive to high-entropy discrepancy patterns. Since $\gamma\ge 0$ and $0\le H(\cdot)\le \log |V|$, the gate satisfies $|V|^{-\gamma}\le w_t(x)\le 1$, implying $|V|^{-\gamma}F_S(\bm{\theta}_{\text{nav}})\le \bar U(S)\le F_S(\bm{\theta}_{\text{nav}})$. This boundedness is mainly a consistency property; the primary role of the entropy gate is its monotone confidence-aware reweighting effect.

Overall, the KL-based utility and the KD regularizer are locally connected through the same navigator-side discrepancy. Around the navigator checkpoint, the utility serves as a confidence-aware proxy of this discrepancy, providing a locally aligned signal for constructing capability-preserving constraints. Appendix~\ref{app:utility_constraint_effectiveness} further examines this utility signal empirically by comparing constraint sets selected from different utility ranges under the same final KD objective.

\section{Derivation Details of the Selection Algorithm}
\label{Derivation_Details}

\subsection{Derivation of the Selection Objective}
\label{app:kdpp_to_ud}

In this section, we derive the utility--diversity selection objective from the standard $k$-DPP formulation. We start from
\begin{equation}
 S^*
=
\operatornamewithlimits{arg\,max}_{S\subseteq \mathcal{Y},\,|S|=K}
\log\det(\mathbf{L}_S),
\label{eq:app_kdpp_start}
\end{equation}
where $\mathbf{L}_S$ denotes the principal submatrix of the $L$-ensemble kernel indexed by the subset $S$. Following the $L$-ensemble decomposition, let $\mathbf{L}=\mathbf{Q}\mathbf{G}\mathbf{Q}$, where $\mathbf{Q}\triangleq \text{diag}(q_1,\ldots,q_N)$ and $\mathbf{G}\succeq 0$. For any subset $S$, the corresponding principal submatrix can be written as $\mathbf{L}_S = \mathbf{Q}_S\,\mathbf{G}_S\,\mathbf{Q}_S$, so taking determinants gives $\det(\mathbf{L}_S) = \det(\mathbf{Q}_S)\,\det(\mathbf{G}_S)\,\det(\mathbf{Q}_S)$. Since $\mathbf{Q}_S$ is diagonal, $\det(\mathbf{Q}_S)=\prod_{i\in S} q_i$, and therefore $\det(\mathbf{L}_S) = \Big(\prod_{i\in S} q_i^2\Big)\det(\mathbf{G}_S)$. Taking logarithms yields
\begin{equation}
\log\det(\mathbf{L}_S)
=
2\sum_{i\in S}\log q_i
+
\log\det(\mathbf{G}_S).
\label{eq:app_logdet_split}
\end{equation}
Substituting Eq.~\eqref{eq:app_logdet_split} into Eq.~\eqref{eq:app_kdpp_start} gives the equivalent objective
\begin{equation}
 S^*
=
\operatornamewithlimits{arg\,max}_{S\subseteq \mathcal{Y},\,|S|=K}
\left(
2\sum_{i\in S}\log q_i
+
\log\det(\mathbf{G}_S)
\right).
\label{eq:app_kdpp_equiv1}
\end{equation}
We next substitute the quality parametrization $q_i \triangleq \exp\!\big(u_i/(2\beta)\big)$ with $\beta>0$. Under this choice, and noting that multiplying the objective by the positive constant $\beta$ does not change the argmax, Eq.~\eqref{eq:app_kdpp_equiv1} becomes
\begin{equation}
 S^*
=
\operatornamewithlimits{arg\,max}_{S\subseteq \mathcal{Y},\,|S|=K}
\left(
\frac{1}{\beta}\sum_{i\in S} u_i
+
\log\det(\mathbf{G}_S)
\right)
=
\operatornamewithlimits{arg\,max}_{S\subseteq \mathcal{Y},\,|S|=K}
\left(
\sum_{i\in S} u_i
+
\beta\,\log\det(\mathbf{G}_S)
\right),
\label{eq:app_kdpp_equiv2}
\end{equation}
Finally, we instantiate $\mathbf{G}$ as a Gram kernel built from the sample embeddings $\mathbf{z}_i\in\mathbb{R}^d$, so that $\mathbf{G}_S = \mathbf{Z}_S^\top \mathbf{Z}_S$, where $\mathbf{Z}_S \triangleq [\mathbf{z}_i]_{i\in S}\in\mathbb{R}^{d\times K}$. To improve numerical stability and avoid rank deficiency, we replace $\log\det(\mathbf{G}_S)$ with the stabilized diversity term $\log\det(\mathbf{I}_K+\mathbf{G}_S)$~\citep{chen2026posttraining} and apply Sylvester's determinant identity~\citep{horn2012matrix}:
\begin{equation}
\det(\mathbf{I}_K+\mathbf{Z}_S^\top \mathbf{Z}_S)
=
\det(\mathbf{I}_d+\mathbf{Z}_S \mathbf{Z}_S^\top)
=
\det\!\Big(\mathbf{I}_d+\sum_{i\in S}\mathbf{z}_i\mathbf{z}_i^\top\Big).
\label{eq:app_sylvester}
\end{equation}
Using this stabilized diversity term, Sylvester's identity gives the objective
\begin{equation}
 S^*
=
\operatornamewithlimits{arg\,max}_{S\subseteq \mathcal{Y},\,|S|=K}
\sum_{i\in S} u_i
+
\beta \log\det\!\left(\mathbf{I}_d+\sum_{i\in S} \mathbf{z}_i \mathbf{z}_i^\top\right),
\label{eq:app_ud_final}
\end{equation}
which is the stabilized utility--diversity objective used in the main text.

\subsection{Objective Discrepancy Between the Discrete and Sparse Objectives}
\label{app:subset_to_sparse_gap}

In this section, we analyze the discrepancy between the discrete subset-selection objective in Eq.~\eqref{eq:final_obj} and the sparse continuous objective in Eq.~\eqref{eq:iht_obj}.

Let $F_{\text{disc}}(S)$ denote the discrete objective in Eq.~\eqref{eq:final_obj}, $F_0(\mathbf{w})$ the unregularized sparse objective, and $F_{\text{spr}}(\mathbf{w})\triangleq F_0(\mathbf{w})-\frac{\nu}{2}\|\mathbf{w}\|_2^2$ the full sparse objective in Eq.~\eqref{eq:iht_obj}. For any subset $S\subseteq[N]$ with $|S|=K$, let $\mathbf 1_S\in\{0,1\}^N$ be its indicator vector. Then, by direct substitution, $F_0(\mathbf 1_S)=F_{\text{disc}}(S)$. This shows that the unregularized sparse objective exactly recovers the discrete objective when the weights are restricted to binary indicator vectors. For each coordinate,
\begin{equation}
\frac{\partial F_0(\mathbf{w})}{\partial w_i}
=
u_i
+
\beta\,\mathbf{z}_i^\top
\Bigl(
\mathbf{I}_d+\sum_{j=1}^N w_j \mathbf{z}_j \mathbf{z}_j^\top
\Bigr)^{-1}
\mathbf{z}_i.
\end{equation}
Since $\mathbf{I}_d+\sum_{j=1}^N w_j \mathbf{z}_j \mathbf{z}_j^\top \succeq \mathbf{I}_d$, we have $\Bigl(
\mathbf{I}_d+\sum_{j=1}^N w_j \mathbf{z}_j \mathbf{z}_j^\top
\Bigr)^{-1}\preceq \mathbf{I}_d$. Hence
\begin{equation}
0
\le
\mathbf{z}_i^\top
\Bigl(
\mathbf{I}_d+\sum_{j=1}^N w_j \mathbf{z}_j \mathbf{z}_j^\top
\Bigr)^{-1}
\mathbf{z}_i
\le
\|\mathbf{z}_i\|_2^2,
\end{equation}
which implies $\left|\frac{\partial F_0(\mathbf{w})}{\partial w_i}\right| \le |u_i|+\beta\|\mathbf{z}_i\|_2^2$. Let $L \triangleq \max_{1\le i\le N}\bigl(|u_i|+\beta\|\mathbf{z}_i\|_2^2\bigr)$. Then $\|\nabla F_0(\mathbf{w})\|_\infty \le L$ uniformly over the feasible region. Now fix any $\mathbf{w}\succeq 0$ with $\|\mathbf{w}\|_0=K$, and let $S=\{i:\,w_i\neq 0\}$ denote its induced support. Because both $\mathbf{w}$ and $\mathbf 1_S$ are nonnegative and supported on $S$, every point on the line segment $\xi(\tau)=\mathbf 1_S+\tau(\mathbf{w}-\mathbf 1_S)$ for $\tau\in[0,1]$ remains nonnegative and supported on $S$, and the matrix inside the log-determinant remains positive definite along this segment. Applying the mean-value theorem, there exists some $\bar\tau\in(0,1)$ such that $F_0(\mathbf{w})-F_0(\mathbf 1_S)=\left\langle \nabla F_0(\xi(\bar\tau)),\, \mathbf{w}-\mathbf 1_S \right\rangle$. Hence,
\begin{equation}
\begin{aligned}
|F_0(\mathbf{w})-F_{\text{disc}}(S)|
&=
|F_0(\mathbf{w})-F_0(\mathbf 1_S)|
\le
\|\nabla F_0(\xi(\bar\tau))\|_\infty\,\|\mathbf{w}-\mathbf 1_S\|_1
\le
L\,\|\mathbf{w}-\mathbf 1_S\|_1.
\end{aligned}
\end{equation}
Using the definition of $F_{\text{spr}}$, we further have
\begin{equation}
F_{\text{spr}}(\mathbf{w})-F_{\text{disc}}(S)
=
F_0(\mathbf{w})-F_{\text{disc}}(S)-\frac{\nu}{2}\|\mathbf{w}\|_2^2.
\end{equation}
Combining this with the previous bound yields
\begin{equation}
-\,L\|\mathbf{w}-\mathbf 1_S\|_1-\frac{\nu}{2}\|\mathbf{w}\|_2^2
\;\le\;
F_{\text{spr}}(\mathbf{w})-F_{\text{disc}}(S)
\;\le\;
L\|\mathbf{w}-\mathbf 1_S\|_1-\frac{\nu}{2}\|\mathbf{w}\|_2^2.
\label{eq:local_sparse_disc_gap_final_tight}
\end{equation}
Eq.~\eqref{eq:local_sparse_disc_gap_final_tight} shows that the discrepancy between the sparse objective and its corresponding discrete subset objective is controlled by the deviation of the relaxed weights from the binary indicator vector and the explicit quadratic regularization term.

Let $\mathbf{w}^*$ be any global maximizer of Eq.~\eqref{eq:iht_obj}, and let $S^*=\{i:\,w_i^*\neq 0\}$ be the selected index set induced by $\mathbf{w}^*$. Applying Eq.~\eqref{eq:local_sparse_disc_gap_final_tight} to $\mathbf{w}^*$ gives
\begin{equation}
-\,L\|\mathbf{w}^*-\mathbf 1_{S^*}\|_1-\frac{\nu}{2}\|\mathbf{w}^*\|_2^2
\;\le\;
F_{\text{spr}}(\mathbf{w}^*)-F_{\text{disc}}(S^*)
\;\le\;
L\|\mathbf{w}^*-\mathbf 1_{S^*}\|_1-\frac{\nu}{2}\|\mathbf{w}^*\|_2^2.
\label{eq:optimal_local_gap_final_tight}
\end{equation}

Finally, let $S^*_{\text{disc}}$ denote any optimizer of Eq.~\eqref{eq:final_obj}. By optimality of the discrete solution, $F_{\text{disc}}(S^*)\le F_{\text{disc}}(S^*_{\text{disc}})$. Combining this with the upper bound in Eq.~\eqref{eq:optimal_local_gap_final_tight}, we obtain
\begin{equation}
F_{\text{spr}}(\mathbf{w}^*)
\le
F_{\text{disc}}(S^*_{\text{disc}})
+
L\|\mathbf{w}^*-\mathbf 1_{S^*}\|_1
-\frac{\nu}{2}\|\mathbf{w}^*\|_2^2.
\label{eq:sparse_disc_upper_final_tight}
\end{equation}

On the other hand, since $\mathbf 1_{S^*_{\text{disc}}}$ is feasible for Eq.~\eqref{eq:iht_obj}, optimality of $\mathbf{w}^*$ implies
\begin{equation}
F_{\text{spr}}(\mathbf{w}^*)
\ge
F_{\text{spr}}(\mathbf 1_{S^*_{\text{disc}}})
=
F_{\text{disc}}(S^*_{\text{disc}})
-\frac{\nu}{2}\|\mathbf 1_{S^*_{\text{disc}}}\|_2^2
=
F_{\text{disc}}(S^*_{\text{disc}})
-\frac{\nu}{2}K.
\label{eq:sparse_disc_lower_final}
\end{equation}

Combining Eq.~\eqref{eq:sparse_disc_upper_final_tight} and Eq.~\eqref{eq:sparse_disc_lower_final}, we arrive at the sandwich bound
\begin{equation}
F_{\text{disc}}(S^*_{\text{disc}})
-\frac{\nu}{2}K
\;\le\;
F_{\text{spr}}(\mathbf{w}^*)
\;\le\;
F_{\text{disc}}(S^*_{\text{disc}})
+
L\|\mathbf{w}^*-\mathbf 1_{S^*}\|_1
-\frac{\nu}{2}\|\mathbf{w}^*\|_2^2.
\label{eq:sandwich_final}
\end{equation}

Therefore, Eq.~\eqref{eq:iht_obj} can be viewed as a subset-structured continuous relaxation of Eq.~\eqref{eq:final_obj}. On binary indicator vectors, the unregularized sparse objective exactly recovers the discrete objective, while the quadratic term acts as an explicit regularizer for the relaxed continuous weights. The optimal sparse objective value is bounded around the discrete optimum, with the gap controlled by the deviation from the binary indicator vector and the regularization term.

\subsection{Error Bound for the Diagonal Log-Determinant Approximation}
\label{app:diag_error}

In this section, we show that the error induced by the diagonal log-determinant approximation is characterized by the spectrum of the normalized off-diagonal coupling matrix of the whitened covariance accumulator.

Let $\mathbf{A}(\mathbf{w}) \triangleq \sum_{i=1}^N w_i \tilde{\mathbf{z}}_i \tilde{\mathbf{z}}_i^\top \in \mathbb{R}^{E\times E}$ with $\mathbf{w} \succeq \mathbf{0}$, and define $\mathbf{B}(\mathbf{w}) \triangleq \mathbf{I}_E + \mathbf{A}(\mathbf{w})$. Since each outer product $\tilde{\mathbf{z}}_i \tilde{\mathbf{z}}_i^\top$ is positive semidefinite and each weight satisfies $w_i \ge 0$, we have $\mathbf{A}(\mathbf{w}) \succeq \mathbf{0}$ and therefore $\mathbf{B}(\mathbf{w}) \succ \mathbf{0}$. Applying Hadamard's inequality to $\mathbf{B}(\mathbf{w})$ gives $\det(\mathbf{B}(\mathbf{w})) \le \prod_{e=1}^{E}[\mathbf{B}(\mathbf{w})]_{ee}$, with equality if and only if $\mathbf{B}(\mathbf{w})$ is diagonal. Taking logarithms yields
\begin{equation}
\log\det\!\big(\mathbf{I}_E+\mathbf{A}(\mathbf{w})\big)
\le
\sum_{e=1}^{E}\log\!\big([\mathbf{B}(\mathbf{w})]_{ee}\big).
\end{equation}
Since $[\mathbf{B}(\mathbf{w})]_{ee}=1+[\mathbf{A}(\mathbf{w})]_{ee}$, we obtain
\begin{equation}
\log\det\!\big(\mathbf{I}_E+\mathbf{A}(\mathbf{w})\big)
\le
\sum_{e=1}^{E}\log\!\big(1+[\mathbf{A}(\mathbf{w})]_{ee}\big)
=
\log\det\!\big(\mathbf{I}_E+\operatorname{diag}(\mathbf{A}(\mathbf{w}))\big).
\end{equation}
Moreover, by the definition of $\mathbf{A}(\mathbf{w})$, we have $[\mathbf{A}(\mathbf{w})]_{ee}=\sum_{i=1}^{N} w_i\,\tilde z_{i,e}^{2}$, and therefore
\begin{equation}
\log\det\!\big(\mathbf{I}_E+\mathbf{A}(\mathbf{w})\big)
\le
\sum_{e=1}^{E}\log\!\left(1+\sum_{i=1}^{N} w_i\,\tilde z_{i,e}^{2}\right).
\end{equation}
Thus, the diagonal log-determinant term provides a valid Hadamard-based upper approximation to the original diversity term. We next characterize the induced approximation gap.

We decompose $\mathbf{A}(\mathbf{w})$ into diagonal and off-diagonal components by defining $\mathbf{D}(\mathbf{w}) \triangleq \mathbf{I}_E + \operatorname{diag}(\mathbf{A}(\mathbf{w}))$ and $\mathbf{R}(\mathbf{w}) \triangleq \mathbf{A}(\mathbf{w})-\operatorname{diag}(\mathbf{A}(\mathbf{w}))$, so that $\mathbf{I}_E+\mathbf{A}(\mathbf{w})=\mathbf{D}(\mathbf{w})+\mathbf{R}(\mathbf{w})$. Since $\mathbf{A}(\mathbf{w})\succeq \mathbf{0}$, the matrix $\mathbf{D}(\mathbf{w})$ is positive definite. We then define the normalized off-diagonal coupling matrix $\mathbf{C}(\mathbf{w}) \triangleq \mathbf{D}(\mathbf{w})^{-\frac{1}{2}}\mathbf{R}(\mathbf{w})\mathbf{D}(\mathbf{w})^{-\frac{1}{2}}$. Define the approximation gap
\begin{equation}
\Delta_{\text{diag}}(\mathbf{w})
\triangleq
\log\det\bigl(\mathbf{I}_E+\operatorname{diag}(\mathbf{A}(\mathbf{w}))\bigr)
-
\log\det\bigl(\mathbf{I}_E+\mathbf{A}(\mathbf{w})\bigr).
\end{equation}
From the definition of $\mathbf{C}(\mathbf{w})$, we have $\mathbf{R}(\mathbf{w})=\mathbf{D}(\mathbf{w})^{\frac{1}{2}}\mathbf{C}(\mathbf{w})\mathbf{D}(\mathbf{w})^{\frac{1}{2}}$. It then follows that
\begin{equation}
\mathbf{I}_E+\mathbf{A}(\mathbf{w})
=
\mathbf{D}(\mathbf{w})+\mathbf{R}(\mathbf{w})
=
\mathbf{D}(\mathbf{w})^{\frac{1}{2}}
\bigl(\mathbf{I}_E+\mathbf{C}(\mathbf{w})\bigr)
\mathbf{D}(\mathbf{w})^{\frac{1}{2}} .
\end{equation}
Taking determinants on both sides gives $\det\bigl(\mathbf{I}_E+\mathbf{A}(\mathbf{w})\bigr)=\det\bigl(\mathbf{D}(\mathbf{w})\bigr)\det\bigl(\mathbf{I}_E+\mathbf{C}(\mathbf{w})\bigr)$. Taking logarithms and using $\mathbf{D}(\mathbf{w})=\mathbf{I}_E+\operatorname{diag}(\mathbf{A}(\mathbf{w}))$, we obtain
\begin{equation}
\Delta_{\text{diag}}(\mathbf{w})
=
-\log\det\bigl(\mathbf{I}_E+\mathbf{C}(\mathbf{w})\bigr).
\end{equation}
The Hadamard inequality above immediately implies that $\Delta_{\text{diag}}(\mathbf{w})\ge 0$. Since
\begin{equation}
\mathbf{I}_E+\mathbf{C}(\mathbf{w})
=
\mathbf{D}(\mathbf{w})^{-\frac{1}{2}}
\bigl(\mathbf{I}_E+\mathbf{A}(\mathbf{w})\bigr)
\mathbf{D}(\mathbf{w})^{-\frac{1}{2}},
\end{equation}
and both $\mathbf{D}(\mathbf{w})$ and $\mathbf{I}_E+\mathbf{A}(\mathbf{w})$ are positive definite, $\mathbf{I}_E+\mathbf{C}(\mathbf{w})$ is also positive definite. Let $\lambda_1,\dots,\lambda_E$ be the eigenvalues of $\mathbf{C}(\mathbf{w})$. Then $\lambda_j>-1$ for all $j$, so $\log(1+\lambda_j)$ is well defined. Since $\mathbf{C}(\mathbf{w})$ has zero diagonal, $\sum_{j=1}^{E}\lambda_j=\operatorname{Tr}\bigl(\mathbf{C}(\mathbf{w})\bigr)=0$. Therefore,
\begin{equation}
\Delta_{\text{diag}}(\mathbf{w})
=
-\log\det\bigl(\mathbf{I}_E+\mathbf{C}(\mathbf{w})\bigr)
=
\sum_{j=1}^{E}\left(\lambda_j-\log(1+\lambda_j)\right),
\end{equation}
where the last equality uses $\sum_j\lambda_j=0$. This identity shows that the diagonal approximation gap is exactly determined by the spectrum of the normalized off-diagonal coupling matrix.

We next derive a split-spectrum upper bound. Let $f(x)=x-\log(1+x)$ on $(-1,\infty)$. For $\lambda_j\ge 0$, Taylor's theorem around zero gives $f(\lambda_j)=\frac{1}{2}f''(\xi_j)\lambda_j^2$ for some $\xi_j\in[0,\lambda_j]$. Since $f''(x)=1/(1+x)^2\le 1$ for $x\ge 0$, we have $f(\lambda_j)\le \lambda_j^2/2$. For $\lambda_j<0$, Taylor's theorem gives $f(\lambda_j)=\frac{1}{2}f''(\xi_j)\lambda_j^2$ for some $\xi_j\in[\lambda_j,0]$. Since $f''(x)=1/(1+x)^2$ is decreasing on $(-1,\infty)$, it is maximized over $[\lambda_j,0]$ at $x=\lambda_j$. Hence $f''(\xi_j)\le 1/(1+\lambda_j)^2$, and
\begin{equation}
\lambda_j-\log(1+\lambda_j)
\le
\begin{cases}
\frac{1}{2}\lambda_j^2, & \lambda_j\ge 0,\\[2mm]
\frac{\lambda_j^2}{2(1+\lambda_j)^2}, & \lambda_j<0.
\end{cases}
\end{equation}
Applying this bound to all eigenvalues yields
\begin{equation}
0
\le
\Delta_{\text{diag}}(\mathbf{w})
\le
\frac{1}{2}\sum_{\lambda_j\ge 0}\lambda_j^2
+
\frac{1}{2}\sum_{\lambda_j<0}\frac{\lambda_j^2}{(1+\lambda_j)^2}.
\end{equation}
The asymmetry between the two branches follows from the monotone decrease of $f''$: positive eigenvalues admit a uniform quadratic bound, while negative eigenvalues approaching $-1$ introduce the amplification factor $1/(1+\lambda_j)^2$. Thus, the split-spectrum bound separates benign positive eigen-directions from negative eigen-directions that can amplify the log-determinant gap.

For comparison, define $\delta(\mathbf{w})\triangleq\lambda_{\min}\bigl(\mathbf{I}_E+\mathbf{C}(\mathbf{w})\bigr)>0$. Since $1+\lambda_j\ge\delta(\mathbf{w})$ for all $j$, the split-spectrum bound further implies the coarser conditioning-aware bound
\begin{equation}
0
\le
\Delta_{\text{diag}}(\mathbf{w})
\le
\frac{\|\mathbf{C}(\mathbf{w})\|_F^2}{2\delta(\mathbf{w})^2}.
\end{equation}
This coarser form highlights the role of conditioning but can be conservative when only a small part of the spectrum lies near $-1$.

For the relative gap used in the empirical diagnostics, let $D_{\text{full}}(\mathbf{w})=\log\det\bigl(\mathbf{I}_E+\mathbf{A}(\mathbf{w})\bigr)$. Following Appendix~\ref{app:selection_surrogate_consistency}, let $\epsilon>0$ be a small numerical stabilizer. Then,
\begin{equation}
\text{RelGap}_{\text{diag}}(\mathbf{w})
=
\frac{\Delta_{\text{diag}}(\mathbf{w})}
{|D_{\text{full}}(\mathbf{w})|+\epsilon}
\le
\frac{
\frac{1}{2}\sum_{\lambda_j\ge 0}\lambda_j^2
+
\frac{1}{2}\sum_{\lambda_j<0}\frac{\lambda_j^2}{(1+\lambda_j)^2}
}
{|D_{\text{full}}(\mathbf{w})|+\epsilon}.
\end{equation}
This normalized form matches the relative approximation gap reported in Appendix~\ref{app:selection_surrogate_consistency}.

Finally, since $\|\mathbf{C}(\mathbf{w})\|_F^2=\sum_{j=1}^{E}\lambda_j^2$, the normalized off-diagonal energy can also be written entry-wise as
\begin{equation}
\|\mathbf{C}(\mathbf{w})\|_F^2
=
\sum_{e\neq f}
\frac{[\mathbf{A}(\mathbf{w})]_{ef}^{2}}
{\bigl(1+[\mathbf{A}(\mathbf{w})]_{ee}\bigr)\bigl(1+[\mathbf{A}(\mathbf{w})]_{ff}\bigr)}.
\end{equation}
Therefore, the approximation gap is small when the normalized off-diagonal energy is small and the negative spectrum of $\mathbf{C}(\mathbf{w})$ stays away from $-1$, equivalently when $\delta(\mathbf{w})=\lambda_{\min}(\mathbf{I}_E+\mathbf{C}(\mathbf{w}))$ is bounded away from zero. Thus, whitening can reduce the empirical diagonal approximation gap by lowering normalized off-diagonal coupling and improving the normalized covariance spectrum. Appendix~\ref{app:selection_surrogate_consistency} evaluates the observed approximation gap and ranking consistency of the diagonal surrogate, while Appendix~\ref{app:diag_bound_validation} analyzes this bound on the final selected support by comparing the observed gap with the split-spectrum certificate.

\subsection{Closed-Form Gradient for IHT}
\label{tF_derivation}

In this section, we derive the closed-form gradient $\nabla \tilde{F}(\mathbf{w})$ used in the IHT update in Eq.~\eqref{eq:iht_update}. The surrogate objective in Eq.~\eqref{eq:Ft_surrogate} is
\begin{equation}
\tilde{F}(\mathbf{w}) = \sum_{i=1}^{N} w_i u_i + \beta \sum_{e=1}^{E}\log\!\left(1+\sum_{i=1}^{N} w_i\,\tilde{z}_{i,e}^{2}\right) - \frac{\nu}{2}\|\mathbf{w}\|_2^2.
\label{eq:Ft_surrogate_app}
\end{equation}
For notational convenience, define the aggregated per-dimension energy as $a_e(\mathbf{w}) \triangleq \sum_{i=1}^{N} w_i\,\tilde{z}_{i,e}^{2}$ for $e=1,\ldots,E$. Then the diversity term can be written as $\sum_{e=1}^{E}\log(1+a_e(\mathbf{w}))$.

We differentiate the three terms in $\tilde{F}(\mathbf{w})$ separately. The linear utility term is immediate: since $\sum_{j=1}^{N} w_j u_j$ is linear in $\mathbf{w}$, its partial derivative with respect to $w_i$ is $\frac{\partial}{\partial w_i}\sum_{j=1}^{N} w_j u_j = u_i$. For the diversity term, applying the chain rule gives
\begin{equation}
\frac{\partial}{\partial w_i}\sum_{e=1}^{E}\log\!\left(1+a_e(\mathbf{w})\right)
=
\sum_{e=1}^{E}\frac{1}{1+a_e(\mathbf{w})}\cdot \frac{\partial a_e(\mathbf{w})}{\partial w_i}.
\label{eq:grad_div_step1}
\end{equation}
By the definition of $a_e(\mathbf{w})$, we have $\frac{\partial a_e(\mathbf{w})}{\partial w_i}=\tilde z_{i,e}^{2}$. Substituting this into Eq.~\eqref{eq:grad_div_step1} yields
\begin{equation}
\frac{\partial}{\partial w_i}\sum_{e=1}^{E}\log\!\left(1+a_e(\mathbf{w})\right)
=
\sum_{e=1}^{E}\frac{\tilde z_{i,e}^{2}}{1+a_e(\mathbf{w})}.
\label{eq:grad_div_app}
\end{equation}
For the quadratic regularization term, we directly obtain $\frac{\partial}{\partial w_i}\left(-\frac{\nu}{2}\|\mathbf{w}\|_2^2\right)=-\nu w_i$. Combining the derivatives of the linear utility term, the diversity term in Eq.~\eqref{eq:grad_div_app}, and the quadratic regularizer, the partial derivative of $\tilde{F}(\mathbf{w})$ with respect to $w_i$ is
\begin{equation}
\frac{\partial \tilde{F}(\mathbf{w})}{\partial w_i}
=
u_i
+
\beta\sum_{e=1}^{E}\frac{\tilde{z}_{i,e}^{2}}{1+\sum_{\ell=1}^{N} w_\ell\,\tilde{z}_{\ell,e}^{2}}
-\nu w_i,
\qquad i=1,\ldots,N.
\label{eq:grad_surrogate_app}
\end{equation}
Accordingly, the full gradient vector is
\begin{equation}
\nabla \tilde{F}(\mathbf{w})
=
\left[
\frac{\partial \tilde{F}(\mathbf{w})}{\partial w_1},
\ldots,
\frac{\partial \tilde{F}(\mathbf{w})}{\partial w_N}
\right]^{\top}.
\label{eq:grad_vector_app}
\end{equation}

\subsection{Closed-Form Projected Gradient Update}
\label{app:iht_update_derivation}

In this section, we derive the closed-form projected gradient update used in IHT. We consider the objective $\tilde{F}(\mathbf{w})$ defined in Eq.~\eqref{eq:Ft_surrogate} and the constrained maximization problem
\begin{equation}
\max_{\mathbf{w}\in\mathcal{C}}\ \tilde{F}(\mathbf{w}),
\qquad
\mathcal{C}\triangleq \big\{\mathbf{w}\in\mathbb{R}^{N}:\ \mathbf{w}\succeq \mathbf{0},\ \|\mathbf{w}\|_{0} = K\big\}.
\label{eq:app_iht_problem}
\end{equation}
We focus on the nondegenerate projection case where the thresholded vector has at least $K$ strictly positive entries after nonnegative truncation, as used in the closed-form derivation below. Starting from the current iterate $\mathbf{w}^{(t-1)}$, we consider the first-order expansion of $\tilde{F}$ around $\mathbf{w}^{(t-1)}$,
\begin{equation}
\tilde F(\mathbf w)
\approx
\tilde F(\mathbf w^{(t-1)})
+
\nabla \tilde F(\mathbf w^{(t-1)})^\top
(\mathbf w-\mathbf w^{(t-1)}).
\label{eq:app_first_order}
\end{equation}
This local approximation shows that the ascent direction is given by the gradient $\nabla \tilde F(\mathbf w^{(t-1)})$. Therefore, taking a first-order ascent step with step size $\eta>0$ yields the intermediate iterate
\begin{equation}
\mathbf{v}^{(t)} \triangleq \mathbf{w}^{(t-1)}+\eta\,\nabla \tilde{F}\!\left(\mathbf{w}^{(t-1)}\right).
\label{eq:app_v_def}
\end{equation}
Because the intermediate iterate may violate the nonnegativity and sparsity constraints, we project it back onto $\mathcal{C}$ in the Euclidean sense and define the next iterate as
\begin{equation}
\mathbf{w}^{(t)} \triangleq \Pi_{\mathcal{C}}\!\big(\mathbf{v}^{(t)}\big),
\qquad
\Pi_{\mathcal{C}}(\mathbf{v}) \triangleq \operatornamewithlimits{arg\,min}_{\mathbf{w}\in\mathcal{C}} \|\mathbf{w}-\mathbf{v}\|_2^2 .
\label{eq:app_proj_def}
\end{equation}
Consider the projection subproblem $\min_{\mathbf{w}\succeq \mathbf{0},\,\|\mathbf{w}\|_0 = K}\ \|\mathbf{w}-\mathbf{v}\|_2^2$. Under this condition, the projection onto the exact-$K$ nonnegative set is obtained by retaining $K$ positive coordinates. We first note that negative coordinates of $\mathbf{v}$ can never receive positive mass in the optimum. Indeed, for any coordinate $i$ with $v_i<0$ and any feasible $w_i\ge 0$, we have $(w_i-v_i)^2 \ge (0-v_i)^2$. Hence replacing $w_i$ by zero cannot increase the objective, so an optimal solution must satisfy $w_i=0$ whenever $v_i<0$. This reduces the problem to the nonnegative truncation $\mathbf{v}_{+}\triangleq (\mathbf{v})_{+}$, where $[\mathbf{v}_{+}]_i \triangleq \max(v_i,0)$.

For any fixed index set $S\subseteq\{1,\ldots,N\}$ with $|S|=K$, consider minimizing $\|\mathbf{w}-\mathbf{v}_{+}\|_2^2$ over all nonnegative vectors $\mathbf{w}$ satisfying $w_i=0$ for all $i\notin S$. Once the support is fixed to $S$, the objective is minimized by matching $\mathbf{v}_{+}$ on the active coordinates, that is, by setting $w_i=v_{+,i}$ for $i\in S$ and $w_i=0$ otherwise. Substituting this minimizer gives $\|\mathbf{w}-\mathbf{v}_{+}\|_2^2=\sum_{i\notin S} v_{+,i}^2$. Therefore, minimizing the projection objective is equivalent to choosing a set $S$ of size $K$ that maximizes the retained energy $\sum_{i\in S} v_{+,i}^2$. The optimal set is thus given by the indices of the $K$ largest entries of $\mathbf{v}_{+}$. Let $\Omega_K(\mathbf{v}_{+})$ denote an index set corresponding to the $K$ largest entries of $\mathbf{v}_{+}$. Then the hard-thresholding operator applied after nonnegative truncation can be written componentwise as
\begin{equation}
[\mathcal{P}_K(\mathbf{v}_{+})]_i=
\begin{cases}
v_{+,i}, & i\in \Omega_K(\mathbf{v}_{+}),\\
0, & \text{otherwise}.
\end{cases}
\label{eq:hard_threshold_def}
\end{equation}
Therefore, under the same condition, the Euclidean projection onto $\mathcal{C}$ admits the closed form $\Pi_{\mathcal{C}}(\mathbf{v}) = \mathcal{P}_{K}\!\big(\mathbf{v}_{+}\big)$. Substituting this into Eq.~\eqref{eq:app_proj_def}, we obtain the IHT update
\begin{equation}
\mathbf{w}^{(t)}
=
\mathcal{P}_{K}\!\Big(\big(\mathbf{w}^{(t-1)}+\eta\,\nabla \tilde{F}(\mathbf{w}^{(t-1)})\big)_{+}\Big).
\label{eq:iht_update_app}
\end{equation}

\subsection{Convergence Analysis of IHT after Selection Stabilization}
\label{app:iht_convergence}

In this section, we establish the convergence behavior of the IHT iteration after the selected subset stabilizes. We consider the update
\begin{equation}
\mathbf{w}^{(t)}
=
\mathcal{P}_{K}\!\Big(\big(\mathbf{w}^{(t-1)}+\eta\,\nabla \tilde{F}(\mathbf{w}^{(t-1)})\big)_{+}\Big),
\label{eq:app_iht_update_main}
\end{equation}
From Appendix~\ref{tF_derivation}, the Hessian of $\tilde F$ is
\begin{equation}
\nabla^2\tilde F(\mathbf w)
=
-\beta\sum\nolimits_{e=1}^E
\frac{\mathbf s_e \mathbf s_e^\top}{(1+a_e(\mathbf w))^2}
-\nu I,
\qquad
\mathbf s_e=\bigl(\tilde z_{1,e}^2,\ldots,\tilde z_{N,e}^2\bigr)^\top,
\label{eq:app_hessian_conv}
\end{equation}
where $a_e(\mathbf w)\triangleq \sum_{i=1}^N w_i \tilde z_{i,e}^2$. Assuming $\nu>0$, since $\beta>0$ and each $\mathbf s_e\mathbf s_e^\top$ is positive semidefinite, we have $\nabla^2\tilde F(\mathbf w)\preceq -\nu I$, so $\tilde F$ is $\nu$-strongly concave. Moreover, since $\mathbf w\succeq\mathbf 0$ implies $a_e(\mathbf w)\ge0$, we have $(1+a_e(\mathbf w))^{-2}\le1$. Because the projected-whitened features are fixed, there exists
\begin{equation}
L \triangleq \nu+\beta\sum\nolimits_{e=1}^E \|\mathbf s_e\|_2^2 < \infty
\label{eq:app_L_conv}
\end{equation}
such that $\|\nabla^2 \tilde F(\mathbf w)\|_2\le L$, hence $\nabla \tilde F$ is $L$-Lipschitz. Let $\mathbf w^*$ be an accumulation point of the IHT sequence $\{\mathbf w^{(t)}\}$, and define the truncated forward vector $\bar{\mathbf v}^* \triangleq \bigl(\mathbf w^*+\eta\nabla \tilde F(\mathbf w^*)\bigr)_+$. Suppose that $\bar{\mathbf v}^*$ satisfies the strict top-$K$ margin condition $[\bar{\mathbf v}^*]_{(K)} > [\bar{\mathbf v}^*]_{(K+1)}$, where $[\bar{\mathbf v}^*]_{(k)}$ is its $k$-th largest entry. Since $\mathbf w^*$ is an accumulation point, there exists a subsequence $\{\mathbf w^{(t_k)}\}$ with $\mathbf w^{(t_k)}\to \mathbf w^*$.

Define $\bar{\mathbf v}^{(t)}\triangleq \bigl(\mathbf w^{(t)}+\eta\nabla \tilde F(\mathbf w^{(t)})\bigr)_+$. Since $\nabla \tilde F$ is continuous, any convergent subsequence $\mathbf w^{(t_k)}\to \mathbf w^*$ induces the corresponding convergence $\bar{\mathbf v}^{(t_k)}\to \bar{\mathbf v}^*$. Let
\begin{equation}
S^* \triangleq \Omega_K(\bar{\mathbf v}^*),
\qquad
\delta \triangleq [\bar{\mathbf v}^*]_{(K)}-[\bar{\mathbf v}^*]_{(K+1)} > 0.
\label{eq:app_support_delta}
\end{equation}
Then there exists an index $k_0$ such that $\|\bar{\mathbf v}^{(t_k)}-\bar{\mathbf v}^*\|_\infty < \frac{\delta}{3}$ for all $k\ge k_0$. For any $i\in S^*$ and any $j\notin S^*$, the definition of $\delta$ gives $\bar v_i^*-\bar v_j^* \ge \delta$. Combining this with the above bound, we have $\bar v_i^{(t_k)} \ge \bar v_i^*-\frac{\delta}{3}$ and $\bar v_j^{(t_k)} \le \bar v_j^*+\frac{\delta}{3}$ for all $k\ge k_0$, and therefore
\begin{equation}
\bar v_i^{(t_k)}-\bar v_j^{(t_k)}
\ge
(\bar v_i^*-\bar v_j^*)-\frac{2\delta}{3}
\ge
\frac{\delta}{3}
>0.
\end{equation}
Therefore, $\Omega_K(\bar{\mathbf v}^{(t_k)}) = S^*$ for all sufficiently large $k$. This shows that the top-$K$ selection map is locally stable around $\mathbf w^*$ under the strict margin condition. If the selected subset is fixed to $S^*$ from some iteration onward, for instance because the subsequent iterates remain in this stability neighborhood, the IHT dynamics reduce to projected gradient ascent over $\mathcal C_{S^*}\triangleq \{\mathbf w\succeq \mathbf 0:\ w_i=0 \text{ for } i\notin S^*\}$, and the update takes the form
\begin{equation}
\mathbf w^{(t)}
=
\Pi_{\mathcal C_{S^*}}
\Bigl(
\mathbf w^{(t-1)}+\eta \nabla \tilde F(\mathbf w^{(t-1)})
\Bigr).
\label{eq:app_pg_fixed_support}
\end{equation}
Since $\mathcal C_{S^*}$ is convex and $\tilde F$ is $L$-smooth and $\nu$-strongly concave on it, standard projected-gradient ascent results imply that, for any $0<\eta\le 1/L$, the fixed-subset iteration converges to the unique maximizer $\hat{\mathbf w}_{S^*}=\operatornamewithlimits{arg\,max}_{\mathbf w\in \mathcal C_{S^*}} \tilde F(\mathbf w)$. Thus, the strict margin condition ensures local support stability, and once the support is fixed, the dynamics converge to the unique subset-restricted optimum of the regularized diagonal surrogate.

\subsection{Theoretical Analysis of the Selection Objective}
\label{app:end_to_end_gap}

This section provides a unified analysis of the gap between the original discrete objective in Eq.~\eqref{eq:final_obj} and the final surrogate objective optimized in Eq.~\eqref{eq:Ft_surrogate}. We first decompose the end-to-end discrepancy into three explicit sources of error. We then further show that this discrepancy analysis implies near-optimality of the surrogate-induced subset under the original discrete objective.

\subsubsection{Analysis of the Objective Value Under Eq.~\ref{eq:final_obj} and Eq.~\ref{eq:Ft_surrogate}}
\label{Objective_value}

Let $F_{\text{disc}}(S)$, $F_{\text{spr}}(w)$, and $\tilde F(w)$ denote the objectives in Eq.~\eqref{eq:final_obj}, Eq.~\eqref{eq:iht_obj}, and Eq.~\eqref{eq:Ft_surrogate}, respectively. We further define $F_{\text{pw}}(w)$ as the counterpart of Eq.~\eqref{eq:iht_obj} obtained by replacing the original features with the projected-and-whitened features $\tilde{\mathbf z}_i$. For a feasible sparse weight vector $w \succeq 0$ with $\|w\|_0 = K$, let $S(w) := \{i : w_i \neq 0\}$ denote the subset induced by $w$. Then, by the triangle inequality, the discrepancy between Eq.~\eqref{eq:final_obj} and Eq.~\eqref{eq:Ft_surrogate} is bounded as
\begin{equation}
\bigl|F_{\text{disc}}(S(w))-\tilde F(w)\bigr|
\le
\bigl|F_{\text{disc}}(S(w))-F_{\text{spr}}(w)\bigr|
+
\bigl|F_{\text{spr}}(w)-F_{\text{pw}}(w)\bigr|
+
\bigl|F_{\text{pw}}(w)-\tilde F(w)\bigr|.
\label{eq:triangle_end2end_final}
\end{equation}

The first term captures the discrepancy introduced by relaxing subset selection into continuous sparse optimization, together with the explicit $\ell_2$ regularization term. By Appendix~\ref{app:subset_to_sparse_gap}, this term satisfies
\begin{equation}
\bigl|F_{\text{disc}}(S(w))-F_{\text{spr}}(w)\bigr|
\le
L\|w-\mathbf{1}_{S(w)}\|_1+\frac{\nu}{2}\|w\|_2^2,
\label{eq:first_gap_end2end_final}
\end{equation}
where $L := \max_{1\le i\le N}\bigl(|u_i|+\beta\|\mathbf z_i\|_2^2\bigr)$. The second term quantifies the distortion introduced by random projection and whitening. Since the utility and regularization terms cancel, we have
\begin{equation}
\bigl|F_{\text{spr}}(w)-F_{\text{pw}}(w)\bigr|
=
\beta\Bigl|
\log\det\Bigl(I_d+\sum_{i\in S(w)} w_i \mathbf z_i \mathbf z_i^\top\Bigr)
-
\log\det\Bigl(I_E+\sum_{i\in S(w)} w_i \tilde{\mathbf z}_i\tilde{\mathbf z}_i^\top\Bigr)
\Bigr|.
\label{eq:second_gap_start_final}
\end{equation}
Let $Z_{S(w)}\in\mathbb{R}^{d\times K}$ and $\tilde Z_{S(w)}\in\mathbb{R}^{E\times K}$ denote the feature matrices of the selected subset, let $D_{S(w)}(w) := \text{diag}(w_i)_{i\in S(w)}$, and let $G_{S(w)} := Z_{S(w)}^\top Z_{S(w)}$ and $\tilde G_{S(w)} := \tilde Z_{S(w)}^\top \tilde Z_{S(w)}$ denote the corresponding unweighted Gram matrices. By Sylvester's determinant identity, the above difference can be rewritten as the discrepancy between $\log\det\bigl(I_K + D_{S(w)}(w)^{\frac{1}{2}} G_{S(w)} D_{S(w)}(w)^{\frac{1}{2}}\bigr)$ and $\log\det\bigl(I_K + D_{S(w)}(w)^{\frac{1}{2}} \tilde G_{S(w)} D_{S(w)}(w)^{\frac{1}{2}}\bigr)$.

Assume that projection and whitening preserve the local weighted Gram geometry of the selected subset in the sense that
\begin{equation}
\bigl\|D_{S(w)}(w)^{\frac{1}{2}}\tilde G_{S(w)} D_{S(w)}(w)^{\frac{1}{2}}
-
D_{S(w)}(w)^{\frac{1}{2}}G_{S(w)} D_{S(w)}(w)^{\frac{1}{2}}\bigr\|_2
\le
\varepsilon_{\text{pw}}(S(w),w).
\label{eq:pw_local_geometry_final}
\end{equation}
Then, by the matrix mean value theorem, there exists a positive semidefinite matrix $\mathbf H_{S(w),w}$ on the segment between the two weighted Gram matrices such that
\begin{equation}
\begin{aligned}
    &\bigl|F_{\text{spr}}(w)-F_{\text{pw}}(w)\bigr| \\
&\qquad\le
\beta\|(I_K+\mathbf H_{S(w),w})^{-1}\|_2 \\
&\qquad\qquad \cdot
\bigl\|D_{S(w)}(w)^{\frac{1}{2}}\tilde G_{S(w)} D_{S(w)}(w)^{\frac{1}{2}}
-
D_{S(w)}(w)^{\frac{1}{2}}G_{S(w)} D_{S(w)}(w)^{\frac{1}{2}}\bigr\|_*
\end{aligned}
\end{equation}
Since $\mathbf H_{S(w),w}\succeq 0$, we have $\|(I_K+\mathbf H_{S(w),w})^{-1}\|_2\le 1$. Using $\|M\|_*\le K\|M\|_2$ for any $K\times K$ matrix $M$, we obtain
\begin{equation}
\bigl|F_{\text{spr}}(w)-F_{\text{pw}}(w)\bigr|
\le
\beta K \varepsilon_{\text{pw}}(S(w),w).
\label{eq:second_gap_end2end_final}
\end{equation}

The final term corresponds to the discrepancy introduced by replacing the full log-determinant of the projected covariance matrix with its diagonal surrogate. By the definition of $\Delta_{\text{diag}}(w)$ and the analysis in Appendix~\ref{app:diag_error}, this term satisfies
\begin{equation}
\bigl|F_{\text{pw}}(w)-\tilde F(w)\bigr|
=
\beta\,\Delta_{\text{diag}}(w),
\label{eq:third_gap_end2end_final}
\end{equation}
where $\Delta_{\text{diag}}(w)\ge 0$ denotes the diagonal approximation gap. Combining Eqs.~\eqref{eq:first_gap_end2end_final}, \eqref{eq:second_gap_end2end_final}, and \eqref{eq:third_gap_end2end_final} with Eq.~\eqref{eq:triangle_end2end_final} yields
\begin{equation}
\bigl|F_{\text{disc}}(S(w))-\tilde F(w)\bigr|
\le
\underbrace{L\|w-\mathbf{1}_{S(w)}\|_1+\frac{\nu}{2}\|w\|_2^2}_{\text{Relaxation Gap}}
+
\underbrace{\beta\,K\,\varepsilon_{\text{pw}}(S(w),w)}_{\text{Projection Gap}}
+
\underbrace{\beta\,\Delta_{\text{diag}}(w)}_{\text{Diagonal Approx. Gap}}.
\label{eq:end2end_main_final_summary}
\end{equation}
Accordingly, under the local weighted-Gram preservation assumption, the gap between Eq.~\eqref{eq:final_obj} and Eq.~\eqref{eq:Ft_surrogate} is governed by the relaxation gap, projection-and-whitening distortion, and diagonal log-determinant approximation gap. When these terms are controlled, the surrogate in Eq.~\eqref{eq:Ft_surrogate} remains close to the original utility--diversity objective in Eq.~\eqref{eq:final_obj}.

\subsubsection{Analysis of Near-Optimality Under the Original Discrete Objective}
\label{Near_Optimality}

The discrepancy bound above can be pushed one step further. Beyond objective-value closeness between Eq.~\eqref{eq:Ft_surrogate} and Eq.~\eqref{eq:final_obj}, it shows that maximizing the surrogate yields a subset whose value under the original discrete objective is near-optimal up to an explicit error term.

To connect the pointwise discrepancy bound to optimizer behavior, define the feasible sparse set $\mathcal W_K := \{w \succeq 0 : \|w\|_0 = K\}$ and, for any $w \in \mathcal W_K$, the aggregated discrepancy term
\begin{equation}
\Gamma(w)
:=
L\|w-\mathbf{1}_{S(w)}\|_1+\frac{\nu}{2}\|w\|_2^2
+\beta K \varepsilon_{\text{pw}}(S(w),w)
+\beta \Delta_{\text{diag}}(w).
\label{eq:def_gamma_end2end}
\end{equation}
Here, $\Gamma(w)$ aggregates the relaxation, regularization, projection-whitening, and diagonal-approximation errors between the surrogate and original discrete objectives. Then Eq.~\eqref{eq:end2end_main_final_summary} becomes
\begin{equation}
\bigl|F_{\text{disc}}(S(w))-\tilde F(w)\bigr|
\le
\Gamma(w),
\qquad \forall\, w\in\mathcal W_K.
\label{eq:end2end_gamma_pointwise}
\end{equation}
Now let $\hat w \in \arg\max_{w\in\mathcal W_K}\tilde F(w)$ and $\hat S := S(\hat w)$, and let $S^* \in \arg\max_{S\subseteq \mathcal Y,\ |S|=K} F_{\text{disc}}(S)$ denote an optimal subset of the original discrete objective. Define $w^* := \mathbf{1}_{S^*}\in\mathcal W_K$. Applying Eq.~\eqref{eq:end2end_gamma_pointwise} to $\hat w$ gives $F_{\text{disc}}(\hat S)\ge \tilde F(\hat w)-\Gamma(\hat w)$. Since $\hat w$ maximizes $\tilde F$ over $\mathcal W_K$, we have $\tilde F(\hat w)\ge \tilde F(w^*)$. Applying Eq.~\eqref{eq:end2end_gamma_pointwise} to $w^*$ gives $\tilde F(w^*)\ge F_{\text{disc}}(S^*)-\Gamma(w^*)$. Combining these inequalities yields $F_{\text{disc}}(\hat S)\ge F_{\text{disc}}(S^*)-\Gamma(\hat w)-\Gamma(w^*)$. When both endpoint discrepancies satisfy $\Gamma(\hat w)\le \bar\Gamma$ and $\Gamma(w^*)\le \bar\Gamma$, we obtain $F_{\text{disc}}(\hat S)\ge F_{\text{disc}}(S^*)-2\bar\Gamma$.

Therefore, Eq.~\eqref{eq:end2end_gamma_pointwise} further implies subset-level consistency. When the discrepancy terms are controlled at both the surrogate maximizer $\hat w$ and the discrete optimum indicator $w^*$, the subset induced by the surrogate maximizer remains within an explicit gap of $2\bar\Gamma$ from the optimum of the original discrete objective. This extends the previous discrepancy analysis from objective-value approximation to optimizer-level consistency, indicating that Eq.~\eqref{eq:Ft_surrogate} remains connected to the original utility--diversity objective beyond pointwise objective values. Appendix~\ref{app:support_alignment} further examines this subset-level consistency empirically by comparing the subsets selected by NADS, Fast Greedy, and Lazy-Fast Greedy under the original discrete objective $F_{\text{disc}}$.

\section{Validation Experiments for Theoretical Analysis}
\label{Validation_Experiments}

\subsection{Trajectory Mismatch and Ordering Fidelity}
\label{app:trajectory_ordering}

Appendix~\ref{app:corollary_ranking_stability} gives a local perturbation view of navigator-side forgetting-risk ordering. The navigator follows the unconstrained adaptation trajectory, whereas the final model is trained with KD regularization on the selected constraint set. Thus, the two trajectories need not coincide. This does not conflict with the role of the navigator in NADS, since the navigator is used to rank candidate samples by forgetting-aware utility rather than to reproduce the exact final training trajectory. We therefore examine whether the navigator preserves the ordering of forgetting-sensitive candidates when the final constrained trajectory deviates from the unconstrained one.

We conduct this analysis on Magicoder with LLaMA3.2-3B. 
We fix the navigator and the synthetic candidate pool, and vary the KD strength over $\lambda\in\{0.1,1,5,10,50\}$ when training the final model. 
For each $\lambda$, we quantify trajectory mismatch by measuring the symmetric divergence between the predictive behaviors of the navigator and the final model on a fixed diagnostic subset $\mathcal{C}_{\text{diag}}$ as
\begin{equation}
\text{Mismatch}
=
\frac{1}{|\mathcal{C}_{\text{diag}}|}
\sum_{x_i\in\mathcal{C}_{\text{diag}}}
\frac{1}{|x_i|}
\sum_{t=1}^{|x_i|}
D_{\text{JS}}
\left(
P_{\text{nav}}(\cdot\mid x_{i,<t})
\|
P_{\text{final}}(\cdot\mid x_{i,<t})
\right).
\end{equation}
This metric captures how far the constrained final model moves away from the navigator's unconstrained adaptation behavior at the prediction level. 
We then evaluate whether this deviation changes the utility-induced ordering of candidate samples by comparing the navigator-based scores with final-trajectory reference scores using Spearman correlation and Overlap@$K$, where $K$ follows the same constraint-set budget as in the main experiments.

\begin{figure}[t]
\centering
\includegraphics[width=0.95\linewidth]{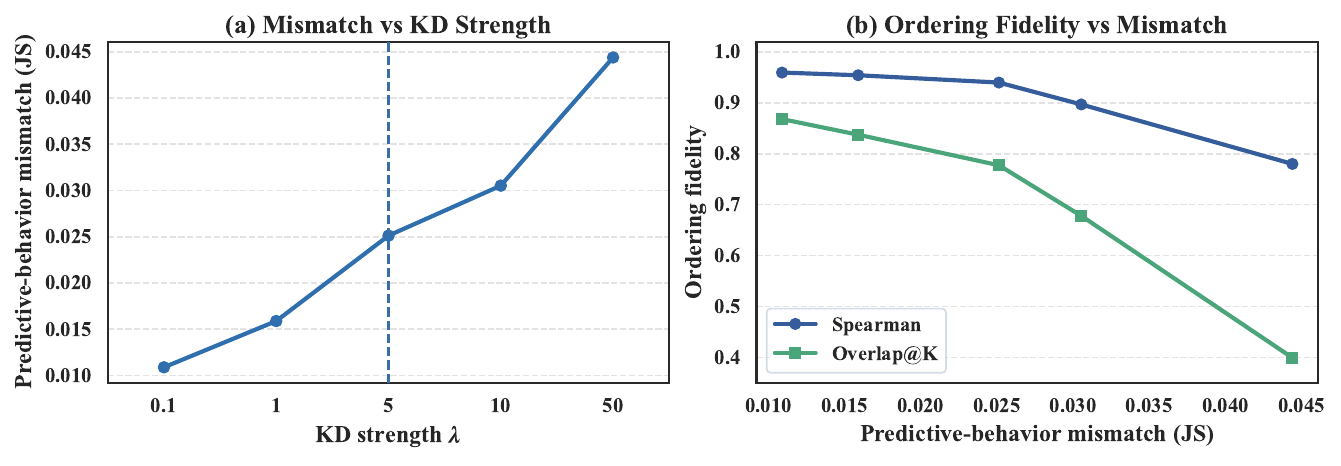}\vspace{-0.1in}
\caption{Trajectory mismatch and ordering fidelity under different KD strengths on Magicoder with LLaMA3.2-3B. The dashed line marks the default setting $\lambda=5$ used in the main experiments. Increasing $\lambda$ enlarges the predictive-behavior mismatch between the unconstrained navigator and the constrained final model, while the navigator preserves high ordering fidelity in the practical regime.}
\label{fig:trajectory_ordering}\vspace{-0.1in}
\end{figure}

As shown in Fig.~\ref{fig:trajectory_ordering}, larger $\lambda$ increases the predictive-behavior mismatch between the navigator and the final model. This is consistent with the KD objective, where stronger distillation pulls the final model closer to the pretrained distribution on selected constraints. Importantly, ordering fidelity remains high near the default NADS setting. At $\lambda=5$, Spearman is $0.94$ and Overlap@$K$ is $0.78$, showing that the navigator still provides a reliable ranking signal for forgetting-aware constraint selection. Under the stress setting $\lambda=50$, the mismatch grows substantially and both ordering metrics decline, indicating that overly strong preservation constraints can disrupt ranking agreement. These results support the intended use of the navigator as a ranking proxy: exact trajectory matching is unnecessary as long as the ordering of forgetting-sensitive candidates is preserved within the practical mismatch range induced by the KD objective.

\subsection{Utility Ranking Stability Across Navigator Checkpoints}
\label{app:rank_stability}

The ordering analysis in Appendix~\ref{app:corollary_ranking_stability} provides a local and conditional view of when navigator-induced utility rankings can be preserved under small changes of the navigator state. We further verify this local view experimentally by examining whether the NADS utility ranking remains stable across nearby navigator checkpoints. To this end, we recompute the utility scores on the same candidate pool using navigator checkpoints from different stages of the unconstrained adaptation trajectory.

We use the three navigator checkpoints \(M_{\text{nav}}-1\), \(M_{\text{nav}}-2\), and \(M_{\text{nav}}-3\) considered in Fig.~\ref{fig3}(c). For each checkpoint \(s \in \{1,2,3\}\), we compute the NADS utility score \(u_i^{(s)}\) for every candidate sample \(x_i\), and obtain the corresponding score-induced ranking. We then measure the ranking agreement between two checkpoints using Spearman correlation and top-\(K\) overlap, defined as
\[
\text{Overlap}(a,b)=
\frac{
\left|\text{TopK}\!\left(u^{(a)}\right)\cap \text{TopK}\!\left(u^{(b)}\right)\right|
}{K}.
\]
Spearman correlation evaluates global ranking consistency over the candidate pool, while top-\(K\) overlap focuses on whether the high-utility candidates used for constraint construction remain stable.

\begin{table}[ht]
\centering
\caption{Utility ranking stability across navigator checkpoints.}
\label{tab:rank_stability}
\renewcommand{\arraystretch}{1.15}
\setlength{\tabcolsep}{5mm}
\small
\begin{tabular}{lcc}
\toprule
\textbf{Checkpoint pair} & \textbf{Spearman} & \textbf{Top-$K$ overlap} \\
\midrule
\(M_{\text{nav}}-1\) vs. \(M_{\text{nav}}-2\) & 0.67 & 0.56 \\
\(M_{\text{nav}}-2\) vs. \(M_{\text{nav}}-3\) & 0.72 & 0.61 \\
\(M_{\text{nav}}-1\) vs. \(M_{\text{nav}}-3\) & 0.47 & 0.39 \\
\bottomrule
\end{tabular}
\end{table}

As shown in Table~\ref{tab:rank_stability}, adjacent navigator checkpoints show stronger ranking agreement than the distant pair, suggesting that the utility ranking is locally consistent along the unconstrained adaptation trajectory but changes across substantially different navigator stages. This agrees with the checkpoint sensitivity in Fig.~\ref{fig3}(c): an early navigator may stay too close to the pretrained model and provide a weak drift signal, whereas an overly late navigator may reflect excessive task-induced drift. Overall, this diagnostic supports the local ordering view in Appendix~\ref{app:corollary_ranking_stability} and explains the navigator-stage sensitivity in Fig.~\ref{fig3}(c) at the ranking level.

\begin{table}[h]
\centering
\caption{Utility-range analysis under the same final KD objective on Magicoder with LLaMA3.2-3B.}
\label{tab:utility_range_intervention}
\renewcommand{\arraystretch}{1.15}
\setlength{\tabcolsep}{5mm}
\small
\begin{tabular}{lccc}
\toprule
\textbf{Constraint source} & \textbf{General Avg.} & \textbf{Target} & \textbf{Overall} \\
\midrule
Low-utility    & 37.82 & 41.78 & 39.80 \\
Middle-utility & 39.38 & 42.52 & 40.95 \\
High-utility   & 41.78 & 43.81 & 42.80 \\
Full NADS           & 42.52 & 44.39 & 43.46 \\
\bottomrule
\end{tabular}
\end{table}

\subsection{Utility Ranking and Constraint Effectiveness}
\label{app:utility_constraint_effectiveness}

Appendix~\ref{app:utility_analysis} establishes a local connection between the KL-based utility and the KD discrepancy in final training. Motivated by the utility--degradation correlation in Fig.~\ref{fig2}, we examine whether the utility ranking translates into effective preservation constraints. We split the candidate pool into high-, middle-, and low-utility ranges by the NADS score. From each range, we select a diverse constraint set with the same budget \(K\), then train the final model with the same \(D_{\text{new}}\), KD objective, distillation weight, and training recipe. Thus, variants differ only in the utility range used for \(D_{\text{constraint}}\).

Specifically, the high-, middle-, and low-utility ranges contain the top, middle, and bottom \(20\%\) of candidates ranked by NADS utility, respectively. Within each range, we apply the same diversity-aware subset construction procedure to select \(K\) constraint samples. We also include NADS, which performs utility--diversity selection over the full candidate pool, as the reference.

As shown in Table~\ref{tab:utility_range_intervention}, high-utility constraint sets yield stronger general-capability retention than middle- and low-utility sets, while target performance remains stable. This shows that the utility score not only correlates with navigator-side degradation but also provides a useful ranking signal for preservation-constraint construction. Full NADS performs best, suggesting that hard truncation to one utility range is less effective than joint utility--diversity selection over the full candidate pool. These results further support navigator utility as a preservation-oriented signal for constraint construction.

\subsection{Consistency of the Scalable Selection Surrogate}
\label{app:selection_surrogate_consistency}

Appendix~\ref{app:diag_error} analyzes the approximation gap introduced by replacing the full log-determinant diversity term with its diagonal surrogate. We further examine this approximation experimentally in the selection stage. The diversity term in NADS is based on a log-determinant objective, but repeatedly evaluating the full log-det term over high-dimensional features is computationally expensive. To make selection scalable, we use projected and whitened features and replace the full log-det diversity term with a diagonal surrogate. This surrogate is efficient, but it should preserve the behavior of the full diversity objective used for subset selection. We therefore examine the consistency between the diagonal surrogate and the full log-det diversity term.

Let $\tilde{\bm z}_i\in\mathbb{R}^{E}$ be the projected feature of sample $x_i$. For a candidate subset $S$, define $\bm A(S)=\sum_{i\in S}\tilde{\bm z}_i\tilde{\bm z}_i^{\top}$. The full and diagonal log-det diversity terms are $D_{\text{full}}(S)=\log\det\left(\bm I+\bm A(S)\right)$ and $D_{\text{diag}}(S)=\sum_{e=1}^{E}\log\left(1+A_{ee}(S)\right)$, respectively. Since the diagonal surrogate ignores the off-diagonal structure of $\bm A(S)$, we measure the relative strength of the ignored component. Let $\bm R(S)=\bm A(S)-\text{diag}(\bm A(S))$. We report the off-diagonal ratio
\begin{equation}
\text{OffDiagRatio}(S)
=
\frac{\|\bm R(S)\|_F}{\|\bm A(S)\|_F+\epsilon},
\end{equation}
and the relative diagonal approximation gap
\begin{equation}
\text{RelGap}_{\text{diag}}(S)
=
\frac{
\left|D_{\text{full}}(S)-D_{\text{diag}}(S)\right|
}{
\left|D_{\text{full}}(S)\right|+\epsilon
}.
\end{equation}
Both metrics are lower when the diagonal surrogate is a tighter approximation. 
We further evaluate whether the surrogate preserves the ranking behavior of the full diversity objective by computing the Spearman correlation between $D_{\text{full}}$ and $D_{\text{diag}}$ over candidate subsets of the same size.

\begin{figure}[t]
\centering
\includegraphics[width=0.95\linewidth]{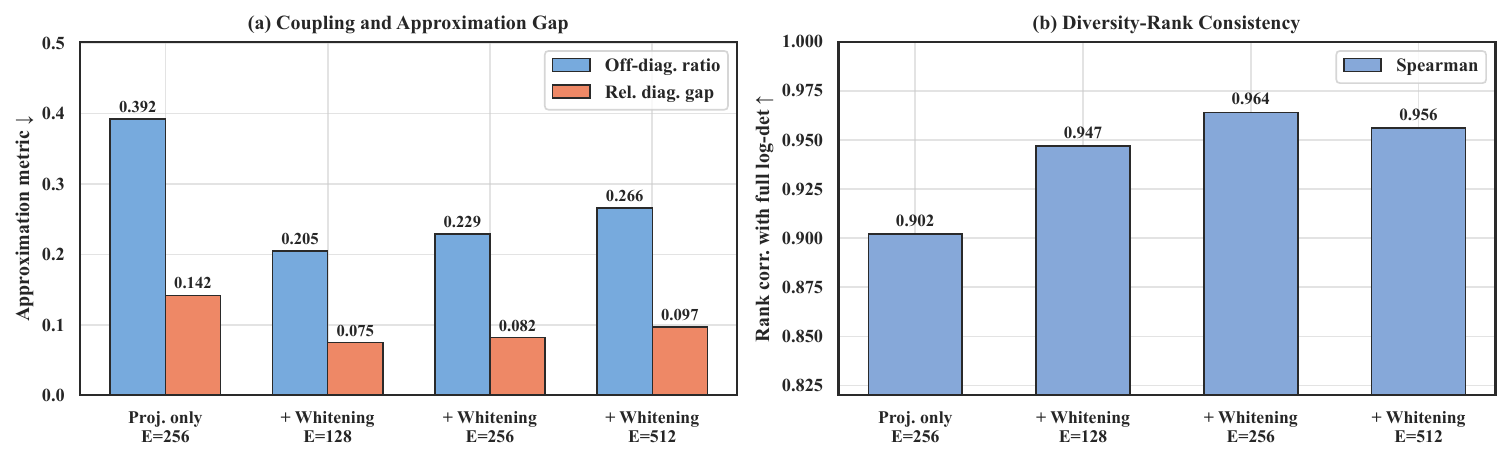}\vspace{-0.1in}
\caption{Scalable selection surrogate consistency on Magicoder with LLaMA3.2. (a) reports the off-diagonal ratio and relative diagonal approximation gap. (b) reports Spearman rank correlation between the full log-det diversity term and its diagonal surrogate over same-size candidate subsets.}
\label{fig:diag_surrogate_fidelity}\vspace{-0.1in}
\end{figure}

As shown in Fig.~\ref{fig:diag_surrogate_fidelity}, projection-only features exhibit the largest off-diagonal ratio and relative diagonal gap, indicating that the diagonal surrogate is less consistent with the full log-det diversity term when feature correlations are not reduced. 
After whitening, both metrics decrease substantially. 
For the default setting $E=256$, the off-diagonal ratio decreases from $0.392$ to $0.229$, and the relative diagonal gap decreases from $0.142$ to $0.082$. 
This supports the role of whitening in reducing the off-diagonal structure ignored by the diagonal approximation.

The rank-consistency results further show that the diagonal surrogate preserves the subset-ranking behavior of the full diversity term. Projection-only features achieve a Spearman correlation of $0.902$, while whitening improves the correlation across all projected dimensions. Although $E=128$ yields the smallest approximation gap, its rank consistency is lower than that of $E=256$, suggesting stronger projection-induced information loss. The default setting $E=256$ achieves the highest rank correlation of $0.964$ while keeping the approximation gap low. Increasing the dimension to $E=512$ slightly increases the off-diagonal ratio and approximation gap, with no further gain in rank consistency. These results provide experimental support for the diagonal approximation analysis in Appendix~\ref{app:diag_error}. They show that whitening reduces the off-diagonal structure ignored by the diagonal surrogate, and that the default setting $E=256$ largely preserves the ranking behavior of the full log-determinant diversity term while keeping the approximation gap low.

\subsection{Validation of the Diagonal Log-Determinant Bound}
\label{app:diag_bound_validation}

\begin{figure}[t]
\centering
\includegraphics[width=0.9\linewidth]{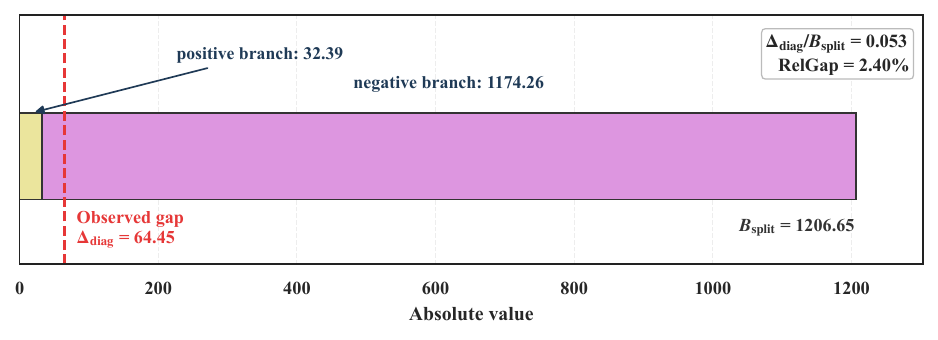}\vspace{-0.1in}
\caption{Split-spectrum certificate for the diagonal log-determinant bound on the final selected constraint set. The stacked bar gives $B_{\text{split}}=B_{+}+B_{-}=1206.65$. The dashed red line marks the observed gap $\Delta_{\text{diag}}=64.45$. The certificate ratio is $\Delta_{\text{diag}}/B_{\text{split}}=0.053$, and the observed relative approximation gap is $2.40\%$.}
\label{fig:diag_bound_certificate}
\end{figure}

Appendix~\ref{app:diag_error} derives a split-spectrum upper bound for the diagonal log-determinant approximation gap. We evaluate both sides of this bound on the final constraint set selected by NADS under the default setting. Let $\Delta_{\text{diag}}=D_{\text{diag}}-D_{\text{full}}$ denote the observed Hadamard gap. Following Appendix~\ref{app:diag_error}, the corresponding upper bound is
\begin{equation}
B_{\text{split}}
=
\underbrace{
\frac{1}{2}\sum_{\lambda_j\ge0}\lambda_j^2
}_{B_{+}}
+
\underbrace{
\frac{1}{2}\sum_{\lambda_j<0}
\frac{\lambda_j^2}{(1+\lambda_j)^2}
}_{B_{-}},
\end{equation}
where $\lambda_j$ are the eigenvalues of the normalized off-diagonal coupling matrix $\mathbf{C}$. Fig.~\ref{fig:diag_bound_certificate} evaluates the \ref{app:diag_error} bound in the actual selection setting. The observed gap $\Delta_{\text{diag}}=64.45$ lies below the split-spectrum certificate $B_{\text{split}}=1206.65$, with $\Delta_{\text{diag}}/B_{\text{split}}=0.053$. This shows that the diagonal approximation gap on the final selected constraint set satisfies the theoretical upper bound. Meanwhile, the relative gap between the diagonal surrogate and the full log-determinant term is only $2.40\%$, indicating that the scalable surrogate remains close to the full diversity term on the selected support.

The decomposition into $B_{+}$ and $B_{-}$ also follows the structure of Appendix~\ref{app:diag_error}. The bound separates standard positive-spectrum contributions from the negative-spectrum branch that controls the worst-case log-determinant behavior. Thus, the certificate supports the split-spectrum bound on the actual selected support while keeping the interpretation appropriately conservative.

\subsection{Consistency Analysis of the Optimization Objective}
\label{app:support_alignment}

To further support the scalable selection strategy introduced in Sec.~\ref{Selection_Algorithm}, we examine whether the subsets selected by NADS remain consistent with the original utility--diversity objective. We compare NADS with two greedy baselines, Fast Greedy~\citep{chen2018fast} and Lazy-Fast Greedy~\citep{hemmi2022lazy}, by evaluating all selected subsets under the original discrete objective $F_{\text{disc}}$ in Eq.~\eqref{eq:final_obj}.

\begin{figure}[t]
    \centering
    \includegraphics[width=0.95\linewidth]{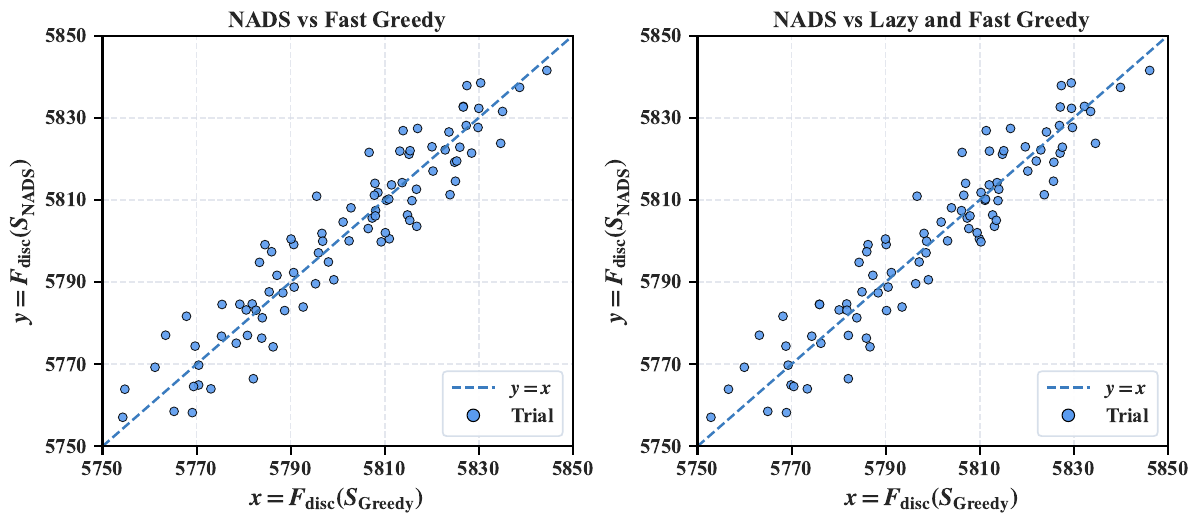}\vspace{-0.1in}
    \caption{Approximation consistency under the original discrete objective. Each point corresponds to one random trial under one greedy comparison. The x-axis denotes the objective value of the subset selected by Fast Greedy or Lazy-Fast Greedy, and the y-axis denotes that of NADS. Points near the diagonal indicate that the scalable NADS procedure remains close to greedy optimization when all subsets are evaluated under the original utility--diversity objective.}
    \label{fig:support_alignment}
\end{figure}

We sample a candidate pool of size $30{,}000$ from the synthetic dataset and select $1000$ samples in each trial. We repeat this procedure over $100$ independent random trials. For each comparison, let $S_{\text{NADS}}^{(t)}$ and $S_{\text{Greedy}}^{(t)}$ denote the subsets returned by NADS and the corresponding greedy baseline in the $t$-th trial, respectively. Here, $S_{\text{Greedy}}^{(t)}$ refers to the subset selected by either Fast Greedy or Lazy-Fast Greedy in the corresponding comparison. We evaluate both subsets using the original discrete objective $F_{\text{disc}}(\cdot)$ in Eq.~\eqref{eq:final_obj} and plot the paired values $x_t = F_{\text{disc}}(S_{\text{Greedy}}^{(t)})$ and $y_t = F_{\text{disc}}(S_{\text{NADS}}^{(t)})$, so that each point $(x_t,y_t)$ in Fig.~\ref{fig:support_alignment} corresponds to one random trial under one greedy comparison. The dashed line denotes $y=x$. Points above this line indicate that NADS achieves a higher objective value than the corresponding greedy baseline under the original discrete objective.

To quantify the signed deviation from the greedy solution, we define
\begin{equation}
\Delta_{\text{rel}}^{(t)}
=
\frac{
F_{\text{disc}}(S_{\text{NADS}}^{(t)})
-
F_{\text{disc}}(S_{\text{Greedy}}^{(t)})
}{
F_{\text{disc}}(S_{\text{Greedy}}^{(t)})
}.
\end{equation}
Points on the diagonal satisfy $x_t=y_t$ and therefore correspond to zero objective gap. As shown in Fig.~\ref{fig:support_alignment}, most points remain close to the diagonal across random trials and greedy comparisons. This indicates that the subsets selected by the scalable NADS procedure achieve objective values close to those of the greedy baselines when evaluated under the original discrete objective. The result suggests that the proposed approximation does not substantially deviate from the behavior of the original utility--diversity objective while enabling scalable selection.

\section{Experimental Details}
\label{app:implementation_details}
In this section, we provide detailed descriptions of the experimental setup for reproducibility. Section~\ref{baseline_details} introduces the implementation and configurations of the baselines. Sections~\ref{training_data} and~\ref{datasetsandevaluation} describe the benchmarks and evaluation details used in our experiments. Section~\ref{implementation} further presents the implementation details, including hyperparameter settings and training configurations. Section~\ref{app:navigator_training} describes the navigator training and validation setting for checkpoint selection. 

\subsection{Baseline Details}
\label{baseline_details}
We compare NADS against SOTA methods spanning PEFT, replay-based constraint selection methods, and target-data selection methods.
\begin{itemize}[leftmargin=*]
    \item \textbf{Standard Fine-Tuning~(Standard FT):} This baseline updates all model parameters using the standard cross-entropy loss. It typically suffers from severe catastrophic forgetting and thus serves as a lower-bound reference for general capability preservation.
    \item \textbf{LoRA~\citep{hu2022lora}:} This baseline injects trainable low-rank decomposition matrices into the Transformer layers while freezing the backbone parameters. Restricting the number of trainable parameters can itself effectively mitigate forgetting~\citep{bidermanlora}. 
    \item \textbf{InsCL~\citep{wang2024inscl}:} InsCL is a replay-based method that selects replay data using instruction similarity measured by Wasserstein distance, and uses the instruction information metric InsInfo to characterize instruction complexity and diversity. By jointly training on target-task data and selected replay samples, InsCL mitigates catastrophic forgetting while preserving target-task adaptation.
    \item \textbf{LER~\citep{m2024leitner}:} LER is a replay-based method that uses Leitner queuing, a human-inspired spaced repetition strategy, to select which historical samples should be replayed at each stage of learning. By prioritizing informative samples and adaptively scheduling their replay according to learning status, LER mitigates catastrophic forgetting while preserving adaptation to new data.
    \item \textbf{Mix-cd~\citep{bai2025efficient}:} Mix-cd is a replay-based method that selects and replays collateral damage samples, which are correctly predicted by the previous model but forgotten after incremental fine-tuning. By incorporating these selected samples into new-task training, Mix-cd mitigates catastrophic forgetting while improving the retention of previously acquired capabilities.
    \item \textbf{LESS~\citep{xia2024less}:} LESS is a selection method. It estimates the influence of training samples in an optimizer-aware manner and selects data by low-rank gradient similarity search with respect to a small set of target examples. By training on the selected subset, LESS improves adaptation to the target capability while substantially reducing the amount of training data required.
    \item \textbf{FisherSFT~\citep{deb2025fishersft}:} FisherSFT is a selection-based method that selects training examples by maximizing information gain under the SFT objective, quantified via the Fisher information matrix. To make this objective practical for large language models, it adopts a last-layer linearization approximation, which enables efficient subset selection. Fine-tuning on the resulting subset improves adaptation efficiency while reducing the amount of training data required.
    \item \textbf{SPICE~\citep{chang2026spice}:} SPICE is a selection-based method using a conflict-aware selector to balance sample informativeness and gradient alignment, while improving efficiency with early stopping and proxy models. Fine-tuning on the selected subset improves target performance at lower selection cost.
\end{itemize}

\begin{table}[t]
\centering
\small
\setlength{\tabcolsep}{7mm}
\renewcommand{\arraystretch}{1.2}
\caption{Summary of training datasets used in our experiments.}
\label{tab:training_data_summary}
\begin{tabular}{l l c}
\toprule
\textbf{Datasets} & \textbf{Aspect} & \textbf{Nums} \\
\midrule
Magicoder~\citep{pmlr-v235-wei24h} & Code Generation & 111K \\
MetaMathQA~\citep{yu2024metamath} & Math Reasoning & 395K \\
Magpie-Qwen2.5-Pro-300K-Filtered~\citep{xu2025magpie} & \textbackslash & 300K \\
\bottomrule
\end{tabular}\vspace{-0.1in}
\end{table}

\subsection{Training Data Details}
\label{training_data}

In our experiments, we consider two target training datasets: code generation and mathematical reasoning. For code generation, we use Magicoder~\citep{pmlr-v235-wei24h}, a decontaminated code dataset. For mathematical reasoning, we use MetaMathQA~\citep{yu2024metamath}, a large-scale mathematical question-answering dataset constructed by augmenting questions from the training sets of GSM8K and MATH. In addition, we adopt Magpie-Qwen2.5-Pro-300K-Filtered~\citep{xu2025magpie} as a large-scale synthetic dataset, generated with the Magpie pipeline using Qwen2.5-72B-Instruct. Detailed dataset statistics are summarized in Table~\ref{tab:training_data_summary}.

\subsection{Evaluation Details}
\label{datasetsandevaluation}
We evaluate model performance from two perspectives: general capability retention and target-task performance. Following recent studies~\citep{wang2025on,lin2026sft,sanyalupweighting,liu2026learn,yuan2026differential}, we decouple them to assess task adaptation and preservation of pre-existing abilities, with benchmark roles adjusted across fine-tuning scenarios. For models fine-tuned on Magicoder~\citep{pmlr-v235-wei24h}, HumanEval~\citep{chen2021evaluatinglargelanguagemodels} is the target-task benchmark for code generation, while HellaSwag~\citep{zellers2019hellaswag}, OpenBookQA~\citep{mihaylov2018can}, MMLU~\citep{hendrycks2021measuring}, and GSM8K~\citep{cobbe2021training} evaluate general capability retention. For models fine-tuned on MetaMathQA~\citep{yu2024metamath}, GSM8K is the target-task benchmark for mathematical reasoning, while HellaSwag, OpenBookQA, MMLU, and HumanEval evaluate general capability retention. Here, HumanEval serves as an out-of-domain benchmark for retained code generation ability after math fine-tuning. Table~\ref{tab:eval_settings} summarizes the settings. All benchmarks are evaluated with \texttt{lm-evaluation-harness}~\citep{eval-harness} using standard prompting setups. General Capability Avg. is the arithmetic mean over the four general-capability benchmarks, and Overall Avg. is the arithmetic mean of General Capability Avg. and target-task performance.

\begin{table}[t]
\centering
\small
\setlength{\tabcolsep}{8pt}
\renewcommand{\arraystretch}{1.2}
\caption{Evaluation benchmarks and detailed settings.}
\begin{tabular}{l l l c c}
\toprule
\textbf{Datasets} & \textbf{Aspects} & \textbf{Metrics} & \textbf{Nums} & \textbf{Evaluate} \\
\midrule
HellaSwag~\citep{zellers2019hellaswag} & Commonsense Reasoning & Accuracy & 10,042 & 0-shot \\
OpenBookQA~\citep{mihaylov2018can} & Commonsense Reasoning & Accuracy & 500 & 0-shot \\
MMLU~\citep{hendrycks2021measuring} & Multi Task Understanding & Accuracy & 14,042 & 5-shot \\
GSM8K~\citep{cobbe2021training} & Math Reasoning & Exact Match & 1,319 & 5-shot \\
HumanEval~\citep{chen2021evaluatinglargelanguagemodels} & Code Generation & PASS@1 & 164 & 0-shot \\
\bottomrule
\end{tabular}
\label{tab:eval_settings}\vspace{-0.1in}
\end{table}

\begin{itemize}[leftmargin=*]
    \item \textbf{HellaSwag:} HellaSwag~\citep{zellers2019hellaswag} is a challenge dataset for commonsense natural language inference. Given an event description or partial context, the model is required to choose the most plausible continuation from several candidate endings. The dataset is constructed via Adversarial Filtering, where a series of discriminators iteratively select adversarial machine-generated negative options, making the benchmark easy for humans but substantially more difficult for SOTA models. Models are typically evaluated using the \textbf{Accuracy} metrics.
    \item \textbf{OpenBookQA:} Open Book Question Answering~\citep{mihaylov2018can} is a question-answering dataset inspired by open-book exams for evaluating a model’s understanding of a subject. It consists of elementary-school-level multiple-choice science questions and assesses both the understanding of a core set of science facts and the ability to apply this knowledge to new situations. Models are typically evaluated using the \textbf{Accuracy} metric.
    \item \textbf{MMLU:} Massive Multitask Language Understanding~\citep{hendrycks2021measuring} is a benchmark for evaluating the broad knowledge and reasoning abilities of models across multiple domains. It covers 57 subjects spanning STEM, humanities, social sciences, and professional knowledge. The dataset consists of multiple-choice questions, each with four or five answer options. For each subject, the model is required to select the correct answer based on its knowledge and reasoning. Models are typically evaluated using the \textbf{Accuracy} metric.
    \item \textbf{GSM8K:} Grade School Math 8K~\citep{cobbe2021training} is a benchmark dataset for evaluating the mathematical reasoning ability of models. It consists of grade-school-level math word problems, each accompanied by a step-by-step solution. Given a natural language problem, the model is required to generate a solution containing the correct numerical answer. The questions range from simple arithmetic to more complex multi-step reasoning. Models are typically evaluated using the \textbf{EM} metric.
    \item \textbf{HumanEval:} HumanEval~\citep{chen2021evaluatinglargelanguagemodels} is a benchmark dataset proposed to evaluate the functional correctness of code generation models. It consists of 164 Python programming problems and is commonly evaluated using the \textbf{PASS@1} metric. Given a problem description and a function signature, the model is required to generate Python code that implements the specified function. A generated solution is considered correct if it passes all provided unit tests.
\end{itemize}

We report three evaluation metrics in our experiments, which are defined as follows.
\begin{itemize}[leftmargin=*]
    \item \textbf{Accuracy (ACC):} Accuracy measures the proportion of correct predictions among all evaluation examples. Compared with Exact Match, it is less strict, as it allows multiple acceptable answers for a single example. Formally, it is defined as
\begin{equation}
\text{ACC}=\frac{1}{N}\sum_{i=1}^{N}\mathbb{I}(\hat{y}_i \in Y_i),
\end{equation}
where $N$ denotes the total number of evaluation examples, $\hat{y}_i$ is the model prediction for the $i$-th example, and $Y_i$ denotes the set of acceptable ground-truth answers for that example.

    \item \textbf{Exact Match (EM):} Exact Match measures whether the predicted answer exactly matches the reference answer after normalization. Compared with Accuracy, it is stricter because it requires an exact match between the predicted output and the gold answer string. Formally, it is defined as
\begin{equation}
\text{EM}=\frac{1}{N}\sum_{i=1}^{N}\mathbb{I}(\hat{y}_i=y_i),
\end{equation}
where $\hat{y}_i$ and $y_i$ denote the predicted and reference answers for the $i$-th example, respectively.

    \item \textbf{PASS@k:} PASS@k is a standard metric for code generation. For each problem, the model generates $k$ candidate solutions, and the metric estimates the probability that at least one solution passes all associated unit tests. Formally, it is defined as
    \begin{equation}
    \text{PASS}@k=
    \begin{cases}
    1-\dfrac{\binom{n-c}{k}}{\binom{n}{k}}, & \text{if } n-c \ge k, \\
    1, & \text{if } n-c < k,
    \end{cases}
    \end{equation}
    where $n$ denotes the total number of generated solutions, $c$ denotes the number of correct solutions among them, and $k$ is the number of sampled solutions used for evaluation.
\end{itemize}

\subsection{Implementation Details}
\label{implementation}

All methods are implemented under a unified fine-tuning setup and optimized with AdamW using a cosine learning-rate scheduler. For Magicoder, we use a global batch size of 64, maximum sequence length 4096, and 3 training epochs; for MetaMathQA, we use a global batch size of 128, maximum sequence length 1024, and 2 epochs. The warmup ratio is 0.03 in both settings. For each target dataset, we hold out $1{,}000$ examples for validation and use the remaining target-training split for all methods. NADS uses the entire target-training split for target-task learning and selects $\mathcal{D}_{\text{constraint}}$ from the synthetic candidate pool $\mathcal{C}$, with size fixed at $5\%$ of $\mathcal{C}$. Replay-based constraint selection methods, including Random selection, InsCL~\citep{wang2024inscl}, LER~\citep{m2024leitner}, and Mix-cd~\citep{bai2025efficient}, use the same target-training split and replay size. Target-data selection methods, including LESS~\citep{xia2024less}, FisherSFT~\citep{deb2025fishersft}, and SPICE~\citep{chang2026spice}, select $10\%$ of the target-training split without using $\mathcal{D}_{\text{constraint}}$. We use official repositories for all baselines except LER and Mix-cd, whose code is unavailable. For NADS, we set the confidence gate $\gamma=0.7$, regularization $\nu=3\times10^{-3}$, diversity $\beta=1.0$, distillation $\lambda=5$, IHT iterations $T=50$, and projection dimension $E=256$. We tune the learning rate for all methods in $[1\times10^{-5},\,5\times10^{-4}]$ and report the best results. For NADS, the selected learning rates on Magicoder are $1\times10^{-4}$, $4\times10^{-5}$, and $7\times10^{-5}$ for LLaMA3.2-3B, Gemma3-4B, and Qwen3.5-9B, respectively; on MetaMathQA, they are $5\times10^{-5}$ and $2\times10^{-5}$ for LLaMA3.2-3B and Gemma3-4B, respectively. All experiments are conducted on 4 NVIDIA A100 GPUs.

\begin{table}[t]
\centering
\scriptsize
\renewcommand{\arraystretch}{1.1}
\setlength{\tabcolsep}{2.3mm}
\caption{Performance comparison with methods selecting $\mathcal{D}_{\text{constraint}}$ for replay on MetaMathQA (\%). The \textcolor{red}{best results} are highlighted \textcolor{red}{in red}, and the \textcolor{blue}{second-best} results are highlighted \textcolor{blue}{in blue}.}
\label{tab:MetaMathQA_main}


\begin{tabular}{clccccc|c|c}
\toprule
\multirow{2}{*}{\textbf{Model}} 
& \multirow{2}{*}{\textbf{Method}} 
& \multicolumn{5}{c|}{\textbf{General Capability} $\uparrow$} 
& \multicolumn{1}{c|}{\textbf{Target} $\uparrow$} 
& \multicolumn{1}{c}{\textbf{Overall} $\uparrow$} \\
\cmidrule(lr){3-7} \cmidrule(lr){8-8} \cmidrule(lr){9-9}
& & \textbf{HASW} & \textbf{OBQA} & \textbf{MMLU} & \textbf{HE} & \textbf{Avg.} & \textbf{GSM8K} & \textbf{Avg.} \\
\midrule


& Pre-trained & $\text{55.83}_{\pm \text{0.00}}$ & $\text{31.20}_{\pm \text{0.00}}$ & $\text{56.92}_{\pm \text{0.00}}$ & $\text{28.16}_{\pm \text{0.00}}$ & $\text{43.03}_{\pm \text{0.00}}$ & $\text{27.90}_{\pm \text{0.00}}$ & $\text{35.47}_{\pm \text{0.00}}$ \\
& Standard FT & $\text{50.02}_{\pm \text{0.50}}$ & $\text{28.40}_{\pm \text{0.53}}$ & $\text{48.49}_{\pm \text{0.41}}$ & $\text{15.27}_{\pm \text{0.49}}$ & $\text{35.55}_{\pm \text{0.24}}$ & $\text{65.93}_{\pm \text{0.81}}$ & $\text{50.74}_{\pm \text{0.42}}$ \\
& LoRA        & $\text{52.06}_{\pm \text{0.50}}$ & $\text{28.60}_{\pm \text{0.53}}$ & $\text{52.34}_{\pm \text{0.40}}$ & $\text{25.31}_{\pm \text{0.52}}$ & $\text{39.58}_{\pm \text{0.25}}$ & $\text{61.03}_{\pm \text{0.83}}$ & $\text{50.31}_{\pm \text{0.43}}$ \\
& Random        & $\text{50.38}_{\pm \text{0.50}}$ & $\text{28.47}_{\pm \text{0.58}}$ & $\text{49.54}_{\pm \text{0.40}}$ & $\text{17.35}_{\pm \text{0.46}}$ & $\text{36.44}_{\pm \text{0.24}}$ & $\text{62.70}_{\pm \text{0.79}}$ & $\text{49.57}_{\pm \text{0.41}}$ \\

& InsCL  & $\text{52.85}_{\pm \text{0.50}}$ & $\text{28.93}_{\pm \text{0.58}}$ & $\text{55.17}_{\pm \text{0.40}}$ & $\text{26.34}_{\pm \text{0.21}}$ & $\text{40.82}_{\pm \text{0.22}}$ & $\text{62.77}_{\pm \text{0.87}}$ & $\text{51.80}_{\pm \text{0.45}}$ \\
& LER    & $\text{52.47}_{\pm \text{0.50}}$ & $\text{30.13}_{\pm \text{0.58}}$ & $\text{54.75}_{\pm \text{0.41}}$ & $\textcolor{blue}{\text{27.78}_{\pm \text{0.33}}}$ & $\text{41.28}_{\pm \text{0.23}}$ & $\textcolor{blue}{\text{63.03}_{\pm \text{0.82}}}$ & $\textcolor{blue}{\text{52.16}_{\pm \text{0.43}}}$ \\
& Mix-cd & $\textcolor{blue}{\text{53.35}_{\pm \text{0.49}}}$ & $\textcolor{blue}{\text{30.20}_{\pm \text{0.53}}}$ & $\textcolor{blue}{\text{55.49}_{\pm \text{0.40}}}$ & $\text{26.23}_{\pm \text{0.21}}$ & $\textcolor{blue}{\text{41.32}_{\pm \text{0.21}}}$ & $\text{62.14}_{\pm \text{0.86}}$ & $\text{51.73}_{\pm \text{0.44}}$ \\

\rowcolor{monaPurple}
\cellcolor{white}\multirow{-8}{*}{\rotatebox{90}{\textbf{LLaMA3.2-3B}}}
& NADS (Ours) & $\textcolor{red}{\text{54.86}_{\pm \text{0.50}}}$ & $\textcolor{red}{\text{30.80}_{\pm \text{0.60}}}$ & $\textcolor{red}{\text{56.23}_{\pm \text{0.41}}}$ & $\textcolor{red}{\text{28.21}_{\pm \text{0.42}}}$ & $\textcolor{red}{\text{42.53}_{\pm \text{0.24}}}$ & $\textcolor{red}{\text{65.18}_{\pm \text{0.83}}}$ & $\textcolor{red}{\text{53.86}_{\pm \text{0.43}}}$ \\

\midrule
& Pre-trained & $\text{57.41}_{\pm \text{0.00}}$ & $\text{34.60}_{\pm \text{0.00}}$ & $\text{60.24}_{\pm \text{0.00}}$ & $\text{35.77}_{\pm \text{0.00}}$ & $\text{47.01}_{\pm \text{0.00}}$ & $\text{36.69}_{\pm \text{0.00}}$ & $\text{41.85}_{\pm \text{0.00}}$ \\
& Standard FT & $\text{51.42}_{\pm \text{0.50}}$ & $\text{30.20}_{\pm \text{0.53}}$ & $\text{53.01}_{\pm \text{0.40}}$ & $\text{29.62}_{\pm \text{0.43}}$ & $\text{41.06}_{\pm \text{0.24}}$ & $\text{67.83}_{\pm \text{0.79}}$ & $\text{54.45}_{\pm \text{0.41}}$ \\
& LoRA        & $\text{54.34}_{\pm \text{0.49}}$ & $\text{32.13}_{\pm \text{0.64}}$ & $\text{55.84}_{\pm \text{0.39}}$ & $\text{30.36}_{\pm \text{0.74}}$ & $\text{43.17}_{\pm \text{0.29}}$ & $\text{63.03}_{\pm \text{0.78}}$ & $\text{53.10}_{\pm \text{0.42}}$ \\
& Random        & $\text{51.12}_{\pm \text{0.49}}$ & $\text{30.33}_{\pm \text{0.64}}$ & $\text{53.56}_{\pm \text{0.39}}$ & $\text{29.58}_{\pm \text{0.86}}$ & $\text{41.15}_{\pm \text{0.31}}$ & $\text{63.91}_{\pm \text{0.72}}$ & $\text{52.53}_{\pm \text{0.39}}$ \\

& InsCL  & $\text{54.79}_{\pm \text{0.50}}$ & $\text{32.20}_{\pm \text{0.60}}$ & $\textcolor{blue}{\text{58.47}_{\pm \text{0.40}}}$ & $\text{31.63}_{\pm \text{0.90}}$ & $\text{44.27}_{\pm \text{0.32}}$ & $\textcolor{blue}{\text{65.50}_{\pm \text{0.86}}}$ & $\textcolor{blue}{\text{54.89}_{\pm \text{0.46}}}$ \\
& LER    & $\textcolor{blue}{\text{55.18}_{\pm \text{0.50}}}$ & $\text{32.87}_{\pm \text{0.64}}$ & $\text{58.23}_{\pm \text{0.40}}$ & $\textcolor{blue}{\text{32.54}_{\pm \text{0.75}}}$ & $\textcolor{blue}{\text{44.71}_{\pm \text{0.30}}}$ & $\text{65.05}_{\pm \text{0.84}}$ & $\text{54.88}_{\pm \text{0.45}}$ \\
& Mix-cd & $\text{55.14}_{\pm \text{0.50}}$ & $\textcolor{blue}{\text{33.13}_{\pm \text{0.64}}}$ & $\text{57.98}_{\pm \text{0.39}}$ & $\text{32.14}_{\pm \text{0.89}}$ & $\text{44.60}_{\pm \text{0.32}}$ & $\text{64.72}_{\pm \text{0.86}}$ & $\text{54.66}_{\pm \text{0.46}}$ \\

\rowcolor{monaPurple}
\cellcolor{white}\multirow{-8}{*}[-0.3ex]{\rotatebox{90}{\textbf{Gemma3-4B}}}
& NADS (Ours) & $\textcolor{red}{\text{55.46}_{\pm \text{0.50}}}$ & $\textcolor{red}{\text{33.47}_{\pm \text{0.61}}}$ & $\textcolor{red}{\text{59.31}_{\pm \text{0.40}}}$ & $\textcolor{red}{\text{34.75}_{\pm \text{0.89}}}$ & $\textcolor{red}{\text{45.75}_{\pm \text{0.31}}}$ & $\textcolor{red}{\text{66.52}_{\pm \text{0.77}}}$ & $\textcolor{red}{\text{56.14}_{\pm \text{0.42}}}$ \\

\bottomrule
\end{tabular}
\end{table}

\subsection{Navigator Training and Validation Setting}
\label{app:navigator_training}

For each target dataset, we construct the navigator by fine-tuning $M_{\text{pretrain}}$ on the target-task data without preservation regularization or constraint samples. The navigator follows the same target fine-tuning recipe as the final adaptation stage, including the optimizer, scheduler, batch size, maximum sequence length, warmup ratio, and training epochs. For navigator early stopping, we hold out $1{,}000$ examples from each of the two target datasets as the validation split and use the remaining examples for navigator training. During navigator training, we evaluate checkpoints every 500 training steps and select the checkpoint with the lowest target-task validation negative log-likelihood on the corresponding validation split as $M_{\text{nav}}$.

\section{Experimental Results and Analysis}
\label{Experimental_Results}

\subsection{Results on MetaMathQA Benchmark}
\label{MetaMathQA_result}
In this subsection, we present the experimental results on the MetaMathQA benchmark. We report results on LLaMA3.2-3B and Gemma3-4B. Qwen3.5-9B is not included in this setting because its mathematical performance is already strong before MetaMathQA fine-tuning, which leaves limited room for observing target-side adaptation and makes the trade-off analysis less informative.

Table~\ref{tab:MetaMathQA_main} reports the results under each method's native protocol. On LLaMA3.2-3B, NADS achieves 42.53\% in general capability average and 53.86\% in overall average, compared with 41.32\% and 52.16\% for the strongest baselines. On Gemma3-4B, NADS again obtains the best general capability average and overall average, reaching 45.75\% and 56.14\%, respectively. Although Standard FT achieves strong target performance, it suffers from a much larger degradation in general capability. In contrast, NADS maintains competitive target task performance while substantially improving general capability preservation, leading to the best overall trade off across both model architectures. 

Table~\ref{tab:MetaMathQA_main2} further reports cross-paradigm reference results with methods that select samples from $\mathcal{D}_{\text{new}}$. Since these methods optimize the target-task subset rather than construct synthetic constraint data, this comparison serves as an additional reference rather than a direct same-protocol comparison. Under this reference setting, NADS achieves the best overall average on both LLaMA3.2-3B and Gemma3-4B, while maintaining strong target-task performance. These reference results indicate that NADS achieves a favorable adaptation--preservation balance under the MetaMathQA setting, maintaining strong target-task performance while better preserving general capabilities.

\begin{table}[t]
\centering
\scriptsize
\renewcommand{\arraystretch}{1.2}
\setlength{\tabcolsep}{2.3mm}
\caption{Cross-paradigm comparison with methods selecting $\mathcal{D}_{\text{new}}$ on MetaMathQA (\%).}
\label{tab:MetaMathQA_main2}


\begin{tabular}{clccccc|c|c}
\toprule
\multirow{2}{*}{\textbf{Model}} 
& \multirow{2}{*}{\textbf{Method}} 
& \multicolumn{5}{c|}{\textbf{General Capability} $\uparrow$} 
& \multicolumn{1}{c|}{\textbf{Target} $\uparrow$} 
& \multicolumn{1}{c}{\textbf{Overall} $\uparrow$} \\
\cmidrule(lr){3-7} \cmidrule(lr){8-8} \cmidrule(lr){9-9}
& & \textbf{HASW} & \textbf{OBQA} & \textbf{MMLU} & \textbf{HE} & \textbf{Avg.} & \textbf{GSM8K} & \textbf{Avg.} \\
\midrule

& LESS      & $\textcolor{blue}{\text{53.09}_{\pm \text{0.49}}}$ & $\textcolor{blue}{\text{30.47}_{\pm \text{0.58}}}$ & $\text{53.62}_{\pm \text{0.40}}$ & $\text{26.27}_{\pm \text{0.52}}$ & $\text{40.86}_{\pm \text{0.25}}$ & $\textcolor{blue}{\text{63.63}_{\pm \text{0.87}}}$ & $\textcolor{blue}{\text{52.25}_{\pm \text{0.45}}}$ \\
& FisherSFT & $\text{53.01}_{\pm \text{0.49}}$ & $\text{29.60}_{\pm \text{0.60}}$ & $\text{55.07}_{\pm \text{0.40}}$ & $\text{26.94}_{\pm \text{0.61}}$ & $\text{41.16}_{\pm \text{0.26}}$ & $\text{61.46}_{\pm \text{0.87}}$ & $\text{51.31}_{\pm \text{0.45}}$ \\
& SPICE     & $\text{52.50}_{\pm \text{0.50}}$ & $\text{29.13}_{\pm \text{0.58}}$ & $\textcolor{blue}{\text{55.57}_{\pm \text{0.40}}}$ & $\textcolor{blue}{\text{27.48}_{\pm \text{0.59}}}$ & $\textcolor{blue}{\text{41.17}_{\pm \text{0.26}}}$ & $\text{62.55}_{\pm \text{0.85}}$ & $\text{51.86}_{\pm \text{0.44}}$ \\

\rowcolor{monaPurple}
\cellcolor{white}\multirow{-4}{*}{\rotatebox{90}{\tiny \textbf{LLaMA3.2}}}
& NADS (Ours) & $\textcolor{red}{\text{54.86}_{\pm \text{0.50}}}$ & $\textcolor{red}{\text{30.80}_{\pm \text{0.60}}}$ & $\textcolor{red}{\text{56.23}_{\pm \text{0.41}}}$ & $\textcolor{red}{\text{28.21}_{\pm \text{0.42}}}$ & $\textcolor{red}{\text{42.53}_{\pm \text{0.24}}}$ & $\textcolor{red}{\text{65.18}_{\pm \text{0.83}}}$ & $\textcolor{red}{\text{53.86}_{\pm \text{0.43}}}$ \\

\midrule

& LESS      & $\text{54.93}_{\pm \text{0.49}}$ & $\text{32.00}_{\pm \text{0.60}}$ & $\text{56.24}_{\pm \text{0.40}}$ & $\text{30.97}_{\pm \text{0.89}}$ & $\text{43.54}_{\pm \text{0.32}}$ & $\textcolor{blue}{\text{65.23}_{\pm \text{0.77}}}$ & $\text{54.39}_{\pm \text{0.42}}$ \\
& FisherSFT & $\text{55.26}_{\pm \text{0.49}}$ & $\textcolor{blue}{\text{33.07}_{\pm \text{0.61}}}$ & $\text{56.63}_{\pm \text{0.40}}$ & $\text{32.48}_{\pm \text{0.75}}$ & $\text{44.36}_{\pm \text{0.29}}$ & $\text{64.09}_{\pm \text{0.83}}$ & $\text{54.23}_{\pm \text{0.44}}$ \\
& SPICE     & $\textcolor{red}{\text{55.52}_{\pm \text{0.50}}}$ & $\text{32.33}_{\pm \text{0.64}}$ & $\textcolor{blue}{\text{57.49}_{\pm \text{0.39}}}$ & $\textcolor{blue}{\text{32.87}_{\pm \text{0.76}}}$ & $\textcolor{blue}{\text{44.55}_{\pm \text{0.29}}}$ & $\text{64.85}_{\pm \text{0.82}}$ & $\textcolor{blue}{\text{54.70}_{\pm \text{0.44}}}$ \\

\rowcolor{monaPurple}
\cellcolor{white}\multirow{-4}{*}[-0.3ex]{\rotatebox{90}{\tiny \textbf{Gemma3}}}
& NADS (Ours) & $\textcolor{blue}{\text{55.46}_{\pm \text{0.50}}}$ & $\textcolor{red}{\text{33.47}_{\pm \text{0.61}}}$ & $\textcolor{red}{\text{59.31}_{\pm \text{0.40}}}$ & $\textcolor{red}{\text{34.75}_{\pm \text{0.89}}}$ & $\textcolor{red}{\text{45.75}_{\pm \text{0.31}}}$ & $\textcolor{red}{\text{66.52}_{\pm \text{0.77}}}$ & $\textcolor{red}{\text{56.14}_{\pm \text{0.42}}}$ \\

\bottomrule
\end{tabular}
\end{table}

\begin{table}[t]
\centering
\renewcommand{\arraystretch}{1.3}
\caption{Effect of constraint-set quality under the same KD objective. Random + KD uses randomly selected constraint data, while Navigator + KD uses navigator-selected constraint data.}
\label{tab:kd_fairness}
\scriptsize
\begin{tabular}{lccccc|c|c}
\toprule
\multirow{2}{*}{\textbf{Method}} 
& \multicolumn{5}{c|}{\textbf{General Capability} $\uparrow$} 
& \multicolumn{1}{c|}{\textbf{Target} $\uparrow$} 
& \multicolumn{1}{c}{\textbf{Overall} $\uparrow$} \\
\cmidrule(lr){2-6} \cmidrule(lr){7-7} \cmidrule(lr){8-8}
& \textbf{HASW} & \textbf{OBQA} & \textbf{MMLU} & \textbf{GSM8K} & \textbf{Avg.} & \textbf{HE} & \textbf{Avg.} \\
\midrule
LESS & $\text{54.12}_{\pm \text{0.49}}$ & $\text{30.67}_{\pm \text{0.61}}$ & $\text{54.38}_{\pm \text{0.40}}$ & $\text{23.73}_{\pm \text{0.62}}$ & $\text{40.73}_{\pm \text{0.27}}$ & $\textbf{43.01}_{\pm \textbf{0.84}}$ & $\text{41.87}_{\pm \text{0.44}}$ \\

LESS + Random + KD & $\text{53.21}_{\pm \text{0.49}}$ & $\text{29.73}_{\pm \text{0.61}}$ & $\text{53.52}_{\pm \text{0.39}}$ & $\text{22.31}_{\pm \text{0.64}}$ & $\text{39.69}_{\pm \text{0.27}}$ & $\text{42.15}_{\pm \text{0.81}}$ & $\text{40.92}_{\pm \text{0.43}}$ \\

LESS + Navigator + KD & $\textbf{55.39}_{\pm \textbf{0.49}}$ & $\textbf{31.13}_{\pm \textbf{0.61}}$ & $\textbf{55.74}_{\pm \textbf{0.41}}$ & $\textbf{25.35}_{\pm \textbf{0.64}}$ & $\textbf{41.90}_{
\pm \textbf{0.27}}$ & $\text{42.76}_{\pm \text{0.86}}$ & $\textbf{42.33}_{\pm \textbf{0.45}}$ \\

\midrule

FisherSFT & $\text{54.03}_{\pm \text{0.49}}$ & $\text{29.80}_{\pm \text{0.60}}$ & $\text{52.89}_{\pm \text{0.41}}$ & $\text{23.15}_{\pm \text{0.62}}$ & $\text{39.97}_{\pm \text{0.27}}$ & $\text{41.85}_{\pm \text{0.74}}$ & $\text{40.91}_{\pm \text{0.39}}$ \\

FisherSFT + Random + KD & $\text{52.44}_{\pm \text{0.49}}$ & $\text{29.33}_{\pm \text{0.61}}$ & $\text{51.71}_{\pm \text{0.40}}$ & $\text{21.86}_{\pm \text{0.64}}$ & $\text{38.84}_{\pm \text{0.27}}$ & $\text{41.26}_{\pm \text{0.82}}$ & $\text{40.05}_{\pm \text{0.43}}$ \\

FisherSFT + Navigator + KD & $\textbf{55.61}_{\pm \textbf{0.50}}$ & $\textbf{30.47}_{\pm \textbf{0.61}}$ & $\textbf{54.33}_{\pm \textbf{0.40}}$ & $\textbf{25.27}_{\pm \textbf{0.70}}$ & $\textbf{41.42}_{\pm \textbf{0.28}}$ & $\textbf{42.06}_{\pm \textbf{0.79}}$ & $\textbf{41.74}_{\pm \textbf{0.42}}$ \\

\midrule

SPICE & $\text{53.58}_{\pm \text{0.50}}$ & $\text{29.60}_{\pm \text{0.60}}$ & $\text{53.39}_{\pm \text{0.40}}$ & $\text{24.92}_{\pm \text{0.73}}$ & $\text{40.37}_{\pm \text{0.29}}$ & $\textbf{42.61}_{\pm \textbf{0.66}}$ & $\text{41.49}_{\pm \text{0.36}}$ \\

SPICE + Random + KD & $\text{53.42}_{\pm \text{0.50}}$ & $\text{29.13}_{\pm \text{0.61}}$ & $\text{52.58}_{\pm \text{0.41}}$ & $\text{23.63}_{\pm \text{0.75}}$ & $\text{39.69}_{\pm \text{0.29}}$ & $\text{42.23}_{\pm \text{0.72}}$ & $\text{40.96}_{\pm \text{0.39}}$ \\

SPICE + Navigator + KD & $\textbf{55.21}_{\pm \textbf{0.49}}$ & $\textbf{30.60}_{\pm \textbf{0.60}}$ & $\textbf{55.78}_{\pm \textbf{0.39}}$ & $\textbf{26.54}_{\pm \textbf{0.79}}$ & $\textbf{42.03}_{\pm \textbf{0.30}}$ & $\text{42.49}_{\pm \text{0.80}}$ & $\textbf{42.26}_{\pm \textbf{0.43}}$ \\

\midrule

NADS & $\text{55.17}_{\pm \text{0.49}}$ & $\text{31.07}_{\pm \text{0.58}}$ & $\text{56.45}_{\pm \text{0.41}}$ & $\text{27.37}_{\pm \text{0.67}}$ & $\text{42.52}_{\pm \text{0.27}}$ & $\text{44.39}_{\pm \text{0.40}}$ & $\text{43.46}_{\pm \text{0.24}}$ \\
\bottomrule
\end{tabular}\vspace{-0.1in}
\end{table}

\subsection{Controlled comparison under the same KD objective}
\label{app:training_objective_fairness}

In the main experiments, selection-based baselines fine-tune on selected subsets, creating a training-scale mismatch with NADS. Since they use less new-task supervision, they naturally reduce update-induced drift and operate in a preservation-favorable setting. To test whether this mismatch explains the performance gap, we conduct a controlled comparison on Magicoder with LLaMA3.2 by equipping these baselines with the same final KD objective as NADS. For each baseline, we train a separate navigator on its selected subset and construct a navigator-guided constraint set from the same synthetic pool. We also include a random-constraint variant under the same KD objective. As shown in Table~\ref{tab:kd_fairness}, random constraints do not reliably improve retention and can even reduce general capability, whereas navigator-guided constraints consistently improve general capability across selection-based baselines. This indicates that NADS gains are not merely from adding KD, but from selecting more effective retention constraints from the noisy and redundant synthetic pool. The consistent gains across LESS, FisherSFT, and SPICE further suggest that navigator-guided constraint construction is transferable rather than method-specific.

\subsection{Checkpoint Selection across Models and Datasets}
\label{sec:checkpoint_generalization}

We evaluate Navigator checkpoint selection across five model--dataset combinations covering three models and two target datasets. All settings use the same validation-based early-stopping rule. We compare three checkpoints in each setting and report target-task, general-capability, and overall scores using a single seed. As shown in Table~\ref{tab:checkpoint_generalization}, Nav-2 selected by validation-based early stopping achieves the highest overall score in all five model--dataset combinations. This consistent result demonstrates the robustness of the checkpoint selection rule across different models and target datasets.

\begin{table}[t]
\centering
\footnotesize
\setlength{\tabcolsep}{15pt}
\renewcommand{\arraystretch}{1.3}
\caption{Comparison of three Navigator checkpoints across five additional model--dataset combinations. Results (\%) are obtained using a single seed.}

\begin{tabular}{lllccc}
\toprule
\textbf{Model}
& \textbf{Target Dataset}
& \textbf{Navigator}
& \textbf{Target}
& \textbf{General}
& \textbf{Overall} \\
\midrule

\multirow{3}{*}{LLaMA3.2-3B}
& \multirow{3}{*}{MetaMathQA}
& Nav-1
& $\text{64.75}$ & $\text{41.21}$ & $\text{52.98}$ \\
& & Nav-2
& $\text{65.35}$ & $\text{42.71}$ & $\text{\textbf{54.03}}$ \\
& & Nav-3
& $\text{65.05}$ & $\text{41.73}$ & $\text{53.39}$ \\
\midrule

\multirow{6}{*}{Gemma3-4B}
& \multirow{3}{*}{Magicoder}
& Nav-1
& $\text{57.32}$ & $\text{45.05}$ & $\text{51.19}$ \\
& & Nav-2
& $\text{58.54}$ & $\text{46.53}$ & $\text{\textbf{52.54}}$ \\
& & Nav-3
& $\text{57.93}$ & $\text{45.51}$ & $\text{51.72}$ \\
\cmidrule(lr){2-6}
& \multirow{3}{*}{MetaMathQA}
& Nav-1
& $\text{66.11}$ & $\text{44.67}$ & $\text{55.39}$ \\
& & Nav-2
& $\text{66.72}$ & $\text{45.98}$ & $\text{\textbf{56.35}}$ \\
& & Nav-3
& $\text{66.41}$ & $\text{45.09}$ & $\text{55.75}$ \\
\midrule

\multirow{6}{*}{Qwen3.5-9B}
& \multirow{3}{*}{Magicoder}
& Nav-1
& $\text{73.78}$ & $\text{61.05}$ & $\text{67.42}$ \\
& & Nav-2
& $\text{74.39}$ & $\text{62.17}$ & $\text{\textbf{68.28}}$ \\
& & Nav-3
& $\text{74.39}$ & $\text{61.36}$ & $\text{67.88}$ \\
\cmidrule(lr){2-6}
& \multirow{3}{*}{MetaMathQA}
& Nav-1
& $\text{90.90}$ & $\text{54.74}$ & $\text{72.82}$ \\
& & Nav-2
& $\text{91.43}$ & $\text{55.93}$ & $\text{\textbf{73.68}}$ \\
& & Nav-3
& $\text{91.13}$ & $\text{55.07}$ & $\text{73.10}$ \\
\bottomrule
\end{tabular}
\label{tab:checkpoint_generalization}
\end{table}

\subsection{Sensitivity to Validation-Set Perturbations}
\label{sec:validation_sensitivity}

We assess the sensitivity of Navigator checkpoint selection to validation-set quality and distribution on Magicoder with LLaMA3.2, comparing the original validation set with three variants: 20\% label noise, 40\% label noise, and template shift. Label noise replaces the corresponding proportion of target answers with other samples' answers; template shift uses a template different from the test set to simulate validation--evaluation mismatch. Results are averaged over three seeds and include sample-utility Spearman correlation and Top-$K$ selection overlap relative to the original setting, alongside target-task, general-capability, and overall scores.

\begin{figure}[t]
    \centering
    \includegraphics[width=\linewidth, trim=2mm 1mm 1mm 2mm, clip]{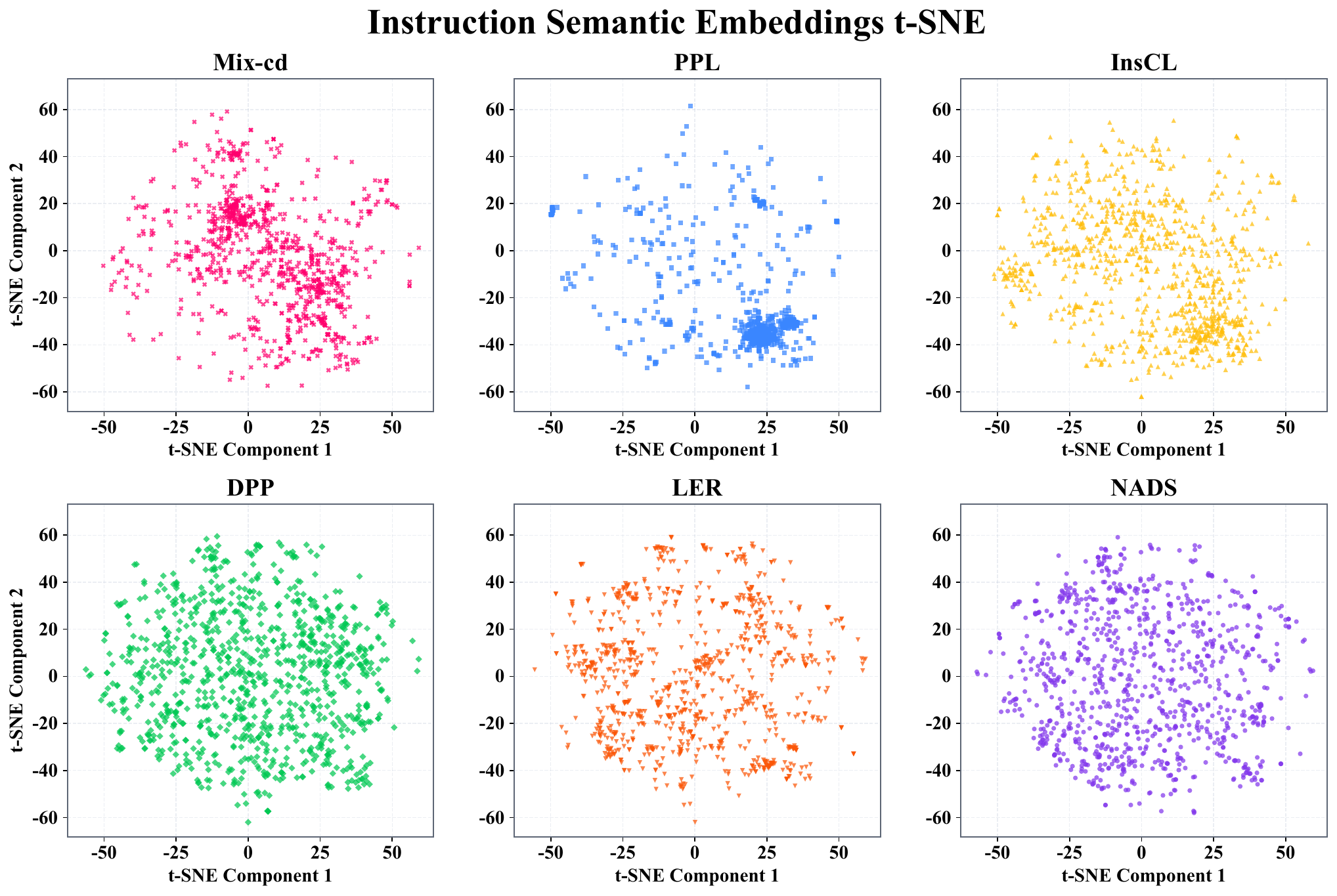}\vspace{-0.15in}
    \caption{t-SNE visualization of the selected data under different selection methods.}
    \label{all_tnse}
\end{figure}

As shown in Table~\ref{tab:validation_sensitivity}, utility rank correlations remain between 0.69 and 0.83, while Top-$K$ overlaps range from 0.58 to 0.72 across the three perturbed validation settings. Although the perturbations change utility rankings and selected samples, target-task, general-capability, and overall scores remain close to those obtained with the original validation set. These results support the robustness of Navigator-based data selection to the tested label noise and template shift.

\begin{table}[t]
\centering
\footnotesize
\caption{Navigator selection sensitivity to validation perturbations on Magicoder with LLaMA3.2.}
\setlength{\tabcolsep}{9.5pt}
\renewcommand{\arraystretch}{1.3}
\begin{tabular}{lccccc}
\toprule
\textbf{Validation}
& \textbf{Utility Spearman}
& \textbf{Top-$K$ Overlap}
& \textbf{Target}
& \textbf{General}
& \textbf{Overall} \\
\midrule
Original
& $1.00$ & $1.00$
& $\text{44.39}_{\pm \text{0.40}}$
& $\text{42.52}_{\pm \text{0.27}}$
& $\text{43.46}_{\pm \text{0.24}}$ \\

20\% label noise
& $0.83$ & $0.72$
& $\text{44.11}_{\pm \text{0.35}}$
& $\text{41.72}_{\pm \text{0.31}}$
& $\text{42.92}_{\pm \text{0.26}}$ \\

40\% label noise
& $0.69$ & $0.58$
& $\text{43.29}_{\pm \text{0.61}}$
& $\text{40.94}_{\pm \text{0.35}}$
& $\text{42.12}_{\pm \text{0.30}}$ \\

Template shifted
& $0.76$ & $0.64$
& $\text{43.70}_{\pm \text{0.35}}$
& $\text{41.20}_{\pm \text{0.33}}$
& $\text{42.45}_{\pm \text{0.28}}$ \\
\bottomrule
\end{tabular}
\label{tab:validation_sensitivity}
\end{table}

\subsection{Diversity Analysis of the Selected Data}
\label{diversity}

The main text presents representative selected-data visualizations. Here, we provide additional t-SNE~\citep{van2008visualizing} visualizations and quantitative diversity diagnostics for more selection methods. As shown in Fig.~\ref{all_tnse}, the DPP-based selector serves as a diversity-oriented reference and broadly covers the semantic space. The PPL-based selector ranks by perplexity and concentrates on dense semantic regions, indicating stronger redundancy. Mix-cd selects samples correctly predicted by the previous model but forgotten by the subsequent model, yielding broad coarse-cluster coverage but less consistent geometric spread. InsCL estimates task similarity via Wasserstein Distance and biases sampling toward high-quality data with InsInfo, giving broader coverage than PPL and Mix-cd but still showing local clustering. LER dynamically updates memory and selects samples by Leitner-based learnability, yet its samples remain less dispersed in the continuous representation space.

\begin{figure}[t]
    \centering
    \includegraphics[width=\linewidth, trim=2mm 1mm 1mm 2mm, clip]{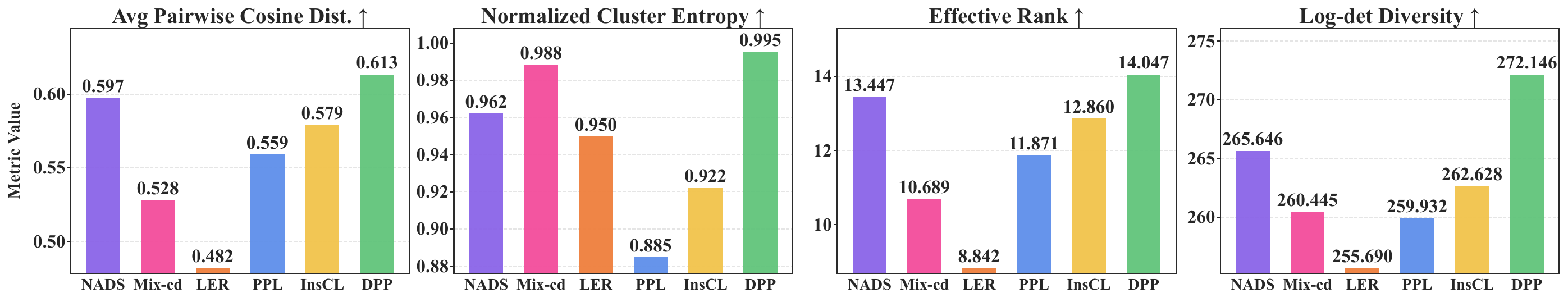}\vspace{-0.15in}
    \caption{Quantitative diversity diagnostics of the selected data under different selection methods. Higher values indicate stronger diversity under the corresponding metric.}\vspace{-0.1in}
    \label{fig:diversity_diagnostics}
\end{figure}

To complement the visualization, Fig.~\ref{fig:diversity_diagnostics} reports four diversity diagnostics: average pairwise cosine distance, normalized cluster entropy, effective rank, and log-det diversity. They measure sample-level separation, coarse cluster coverage, active representation directions, and selected-set geometric volume, respectively. DPP achieves the highest values, as expected from its diversity-oriented objective. NADS remains close to DPP on high-dimensional geometric metrics, including pairwise distance, effective rank, and log-det diversity, while outperforming other non-DPP baselines. Its normalized cluster entropy is lower than DPP and Mix-cd but remains high, indicating broad cluster coverage without simply maximizing uniform coarse-cluster spread. These results show that NADS preserves strong selected-set diversity while incorporating preservation-oriented utility.

\subsection{Computational Cost Analysis} 
\label{Computational}
In this section, we analyze the computational cost and performance–cost trade-off to show that the additional cost introduced by our method is acceptable and justified.

\textbf{Selection Cost.}
Our setting aims to mitigate forgetting during fine-tuning while retaining gains on the new task, which requires estimating forgetting risk before the final training stage. NADS therefore uses a navigator to evaluate candidate samples. In the full-parameter setting, navigator training adds selection-stage cost. On Magicoder with LLaMA3.2-3B, it \textbf{takes about 4 hours on 4 A100 GPUs, or 0.67 GPU-days}. This one-time cost is incurred before final fine-tuning and remains practical at the considered scale. Appendix~\ref{app:lora_navigator} further examines a LoRA-based navigator as an efficient approximation, showing comparable performance with only a small drop. In the same setting, this variant \textbf{uses 11.14 GB of GPU memory and takes about 3 hours on 4 A100 GPUs, or 0.5 GPU-days}. Thus, although some baselines have lower selection overhead, the relevant comparison is the overall cost--performance trade-off, especially when severe forgetting must be controlled.


\label{time_complexity} 
\begin{wrapfigure}{r}{0.52\textwidth}
    \centering\vspace{-0.05in}
    \includegraphics[width=\linewidth, trim=2mm 1mm 1mm 16mm, clip]{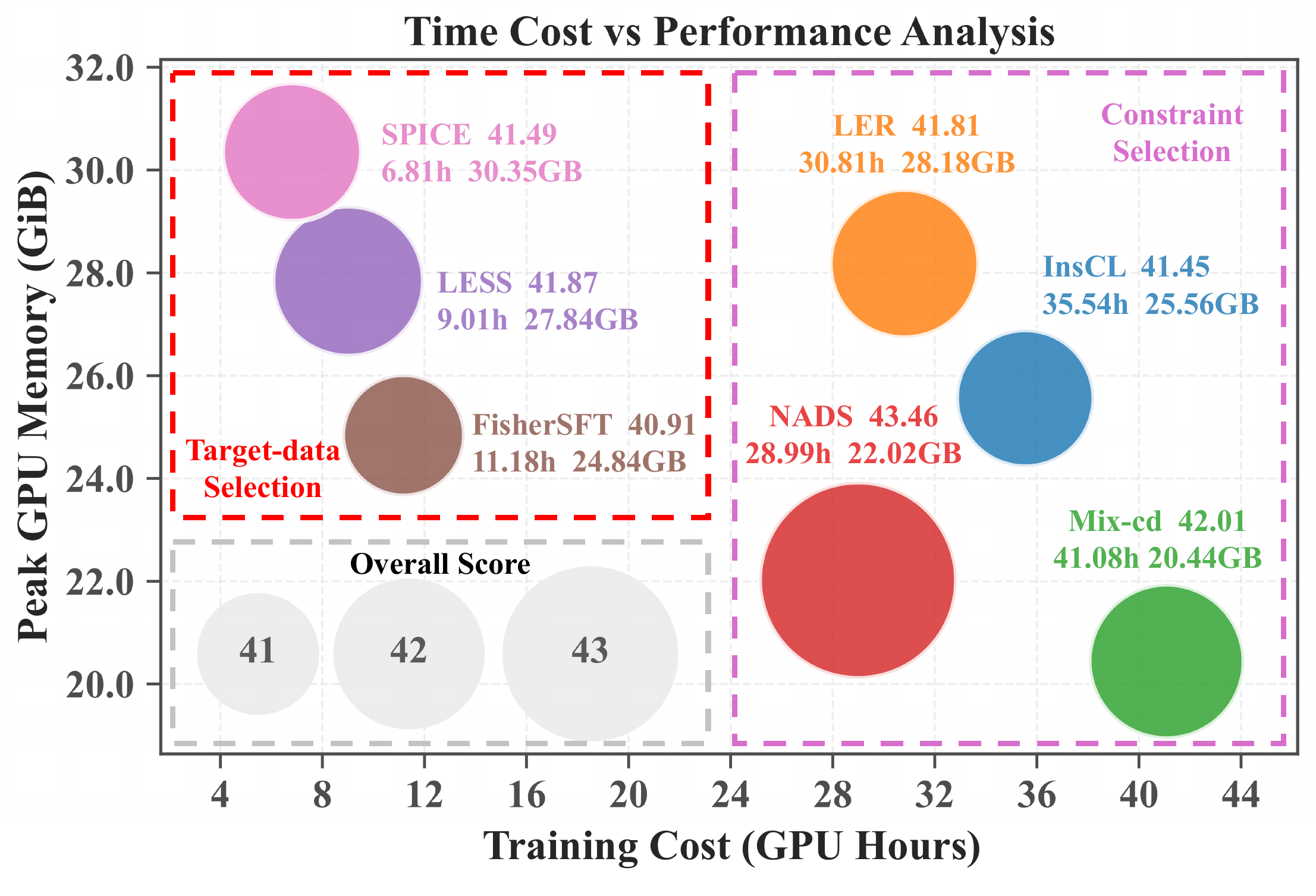}\vspace{-0.2in}
    \caption{Performance and cost visualization of different selection paradigms on Magicoder with LLaMA3.2-3B. Target-data selection methods select from $\mathcal{D}_{\text{new}}$, whereas replay-based methods and NADS select constraints from the synthetic pool $\mathcal{C}$.}\vspace{-0.1in}
    \label{fig:performance_cost_tradeoff}
\end{wrapfigure}
\textbf{Performance--cost Trade-off.}
We further examine the performance--cost trade-off. Fig.~\ref{fig:performance_cost_tradeoff} visualizes methods on Magicoder with LLaMA3.2-3B by training cost, peak GPU memory, and overall score. All methods are run with a per-device micro-batch size of 2 and an effective global batch size of 64 through gradient accumulation. Target-data selection methods select 10\% of the target-training data from $\mathcal{D}_{\text{new}}$, whereas replay-based methods and NADS select 5\% of the synthetic candidates from $\mathcal{C}$ as constraints, matching the main experimental setting. We therefore separate the two groups rather than treating them as a direct ranking. Within the constraint-selection group, NADS achieves the highest Overall score with moderate training cost and peak memory. This suggests that navigator-guided utility and diversity-aware subset construction provide a favorable effectiveness and efficiency balance for constructing preservation constraints.

\subsection{Efficient Navigator Approximation with LoRA}
\label{app:lora_navigator}

The navigator model is designed to trace the unconstrained adaptation path under the target recipe and expose the predictive drift induced by new-task fine-tuning. In principle, its training recipe should match that of the final model, except that preservation constraints are removed, thereby providing a recipe-aligned probe of adaptation-induced forgetting. However, NADS uses the navigator only to induce sample-level drift scores over candidate constraint samples, rather than to reproduce the exact parameter trajectory of the final model. Thus, a LoRA-based navigator can serve as an efficient approximation if it preserves the relative ranking of candidate samples under the drift signal.

Specifically, we train a LoRA-based navigator by adapting $M_{\text{pretrain}}$ on $\mathcal{D}_{\text{new}}$ without anti-forgetting regularization, using $\alpha=16$ and rank $r=32$. The resulting $M_{\text{nav}}^{\text{LoRA}}$ approximates unconstrained adaptation within a low-rank update subspace. To test whether it preserves the NADS selection signal, we compare it with a full-parameter navigator $M_{\text{nav}}^{\text{Full}}$ on Magicoder with LLaMA3.2-3B. Let $s_i^{\text{LoRA}}$ and $s_i^{\text{Full}}$ denote the forgetting-aware scores assigned to candidate $x_i$ by the two navigators. We measure their agreement using \(\text{Spearman}\!\left(\{s_i^{\text{LoRA}}\}_{i=1}^{N},\{s_i^{\text{Full}}\}_{i=1}^{N}\right)\)
and top-$K$ overlap:
\begin{equation}
    \text{Overlap@}K =
\frac{
\left|
\text{TopK}\left(\{s_i^{\text{LoRA}}\}_{i=1}^{N}\right)
\cap
\text{TopK}\left(\{s_i^{\text{Full}}\}_{i=1}^{N}\right)
\right|
}{K}.
\end{equation}
As shown in Table~\ref{tab:lora_navigator}, the LoRA navigator achieves a Spearman correlation of $0.89$ and an Overlap@$K$ of $0.82$ with the full-parameter navigator, while reducing relative training time to $0.74\times$ and relative memory usage to $0.16\times$. This indicates that LoRA preserves most of the sample-ranking signal needed for constraint selection while reducing navigator construction cost.

We further compare the final performance obtained with the two navigator instantiations under the same synthetic candidate pool, constraint selection procedure, and KD objective. 
As shown in Table~\ref{tab:lora_and_full}, NADS with the LoRA navigator achieves performance close to that of the full-parameter navigator, with only a small decrease in general capability average, target performance, and overall score. Overall, these results show that the LoRA navigator provides an efficient approximation to the full-parameter navigator. Thus, LoRA is a practical option when navigator construction cost is a concern, while the full-parameter navigator remains the most recipe-matched instantiation.

\begin{table}[t]
\centering
\small
\caption{Comparison between full-parameter and LoRA-based navigators.}
\label{tab:lora_navigator}
\setlength{\tabcolsep}{16pt}
\begin{tabular}{lcccc}
\toprule
\textbf{Navigator} & \textbf{Spearman} & \textbf{Overlap@K} & \textbf{Rel. Time} & \textbf{Rel. Mem.} \\
\midrule
LoRA navigator & 0.89 & 0.82 & 0.74$\times$ & 0.16$\times$ \\
Full navigator & 1.00 & 1.00 & 1.00$\times$ & 1.00$\times$ \\
\bottomrule
\end{tabular}
\end{table}

\begin{table}[t]
\centering
\footnotesize
\renewcommand{\arraystretch}{1.15}
\setlength{\tabcolsep}{7pt}
\caption{Comparison between LoRA-based and full-parameter navigators under the same final training objective. Both use the same synthetic candidate pool and KD objective.}
\label{tab:lora_and_full}
\begin{tabular}{lccccc|c|c}
\toprule
\multirow{2}{*}{\textbf{Method}} 
& \multicolumn{5}{c|}{\textbf{General Capability} $\uparrow$} 
& \multicolumn{1}{c|}{\textbf{Target} $\uparrow$} 
& \multicolumn{1}{c}{\textbf{Overall} $\uparrow$} \\
\cmidrule(lr){2-6} \cmidrule(lr){7-7} \cmidrule(lr){8-8}
& \textbf{HASW} & \textbf{OBQA} & \textbf{MMLU} & \textbf{GSM8K} & \textbf{Avg.} & \textbf{HE} & \textbf{Avg.} \\
\midrule

NADS + LoRA navigator & $\text{55.03}$ & $\text{30.93}$ & $\text{56.24}$ & $\text{27.22}$ & $\text{42.36}$ & $\text{44.28}$ & $\text{43.32}$ \\

NADS + Full navigator & $\text{55.17}$ & $\text{31.07}$ & $\text{56.45}$ & $\text{27.37}$ & $\text{42.52}$ & $\text{44.39}$ & $\text{43.46}$ \\
\bottomrule
\end{tabular}
\end{table}

\begin{table}[t]
\centering
\footnotesize
\caption{Stage-wise computational costs (GPU-hours) and overall
performance (\%) on Magicoder.
Parenthesized values correspond to LoRA-based Navigator training.}
\label{tab:cost_scaling}
\setlength{\tabcolsep}{3pt}
\renewcommand{\arraystretch}{1.3}
\resizebox{\textwidth}{!}{
\begin{tabular}{llcccccc}
\toprule
\textbf{Model}
& \textbf{Method}
& \shortstack{\textbf{Navigator Training}\\\textbf{(LoRA)}}
& \shortstack{\textbf{Utility}\\\textbf{Computation}}
& \shortstack{\textbf{Subset}\\\textbf{Optimization}}
& \shortstack{\textbf{Final}\\\textbf{Training}}
& \shortstack{\textbf{Total Cost}\\\textbf{(LoRA)}}
& \shortstack{\textbf{Overall}\\\textbf{(LoRA)}} \\
\midrule

LLaMA3.2-3B
& InsCL
& -- & --
& $\text{12.69}$
& $\text{22.85}$
& $\text{35.54}$
& $\text{41.45}$ \\

& LER
& -- & --
& $\text{7.96}$
& $\text{22.85}$
& $\text{30.81}$
& $\text{41.81}$ \\

& Mix-cd
& -- & --
& $\text{18.23}$
& $\text{22.85}$
& $\text{41.08}$
& $\text{42.01}$ \\

& NADS
& $\text{15.83 (11.72)}$
& $\text{5.19}$
& $\text{0.02}$
& $\text{23.78}$
& $\text{44.82 (40.71)}$
& $\text{\textbf{43.46 (43.32)}}$ \\
\midrule

Gemma3-4B
& InsCL
& -- & --
& $\text{12.69}$
& $\text{27.61}$
& $\text{40.30}$
& $\text{50.34}$ \\

& LER
& -- & --
& $\text{9.67}$
& $\text{27.61}$
& $\text{37.28}$
& $\text{50.76}$ \\

& Mix-cd
& -- & --
& $\text{20.99}$
& $\text{27.61}$
& $\text{48.60}$
& $\text{50.63}$ \\

& NADS
& $\text{19.74 (15.39)}$
& $\text{6.02}$
& $\text{0.02}$
& $\text{28.85}$
& $\text{54.63 (50.28)}$
& $\text{\textbf{52.14 (51.96)}}$ \\
\midrule

Qwen3.5-9B
& InsCL
& -- & --
& $\text{12.69}$
& $\text{56.56}$
& $\text{69.25}$
& $\text{65.82}$ \\

& LER
& -- & --
& $\text{18.11}$
& $\text{56.56}$
& $\text{74.67}$
& $\text{66.65}$ \\

& Mix-cd
& -- & --
& $\text{38.08}$
& $\text{56.56}$
& $\text{94.64}$
& $\text{66.06}$ \\

& NADS
& $\text{42.73 (26.58)}$
& $\text{11.27}$
& $\text{0.02}$
& $\text{59.41}$
& $\text{113.43 (97.28)}$
& $\text{\textbf{68.34 (68.13)}}$ \\
\bottomrule
\end{tabular}
}
\end{table}

\subsection{Computational Cost across Model Sizes}
\label{sec:cost_scaling}

We follow the evaluation convention adopted by closely related data-selection methods involving auxiliary training or scoring, including reference-model construction in DavIR~\citep{zhou2025davir} and RHO-1~\citep{NEURIPS2024_3322a9a7}, surrogate or warmup training in ToV~\citep{jain2026train} and LESS~\citep{xia2024less}, and reference--proxy training in DoReMi~\citep{NEURIPS2023_dcba6be9}. Their main performance comparisons generally keep the final target-model training protocol fixed rather than reallocating auxiliary computational costs to longer baseline training. While comparisons under matched end-to-end computational budgets would provide a stricter assessment of computational fairness, we additionally report stage-wise and total costs to make this overhead explicit.

We compare the stage-wise and total computational costs of NADS on LLaMA3.2-3B, Gemma3-4B, and Qwen3.5-9B. All experiments use Magicoder and the same Magpie-Qwen2.5-Pro-300K-Filtered candidate pool, with costs measured in GPU-hours. As shown in Table~\ref{tab:cost_scaling}, NADS achieves the highest overall scores among the compared methods on all three backbones, while subset optimization requires only 0.02 GPU-hours. The total cost increases from 44.82 to 113.43 GPU-hours as model size increases from 3B to 9B parameters. Using LoRA for navigator training reduces the 9B total cost to 97.28 GPU-hours while maintaining a comparable overall score of 68.13, versus 68.34 with full-parameter navigator training.

\subsection{Effect of Constraint-Set Size}
\label{constraint_set_size}
In this section, we study the effect of constraint-set size on NADS. Specifically, we vary the size of the selected constraint set and evaluate how it influences NADS in terms of general capability, target-task performance, and overall performance.

\begin{figure}[t]
    \centering
    \includegraphics[width=0.95\linewidth]{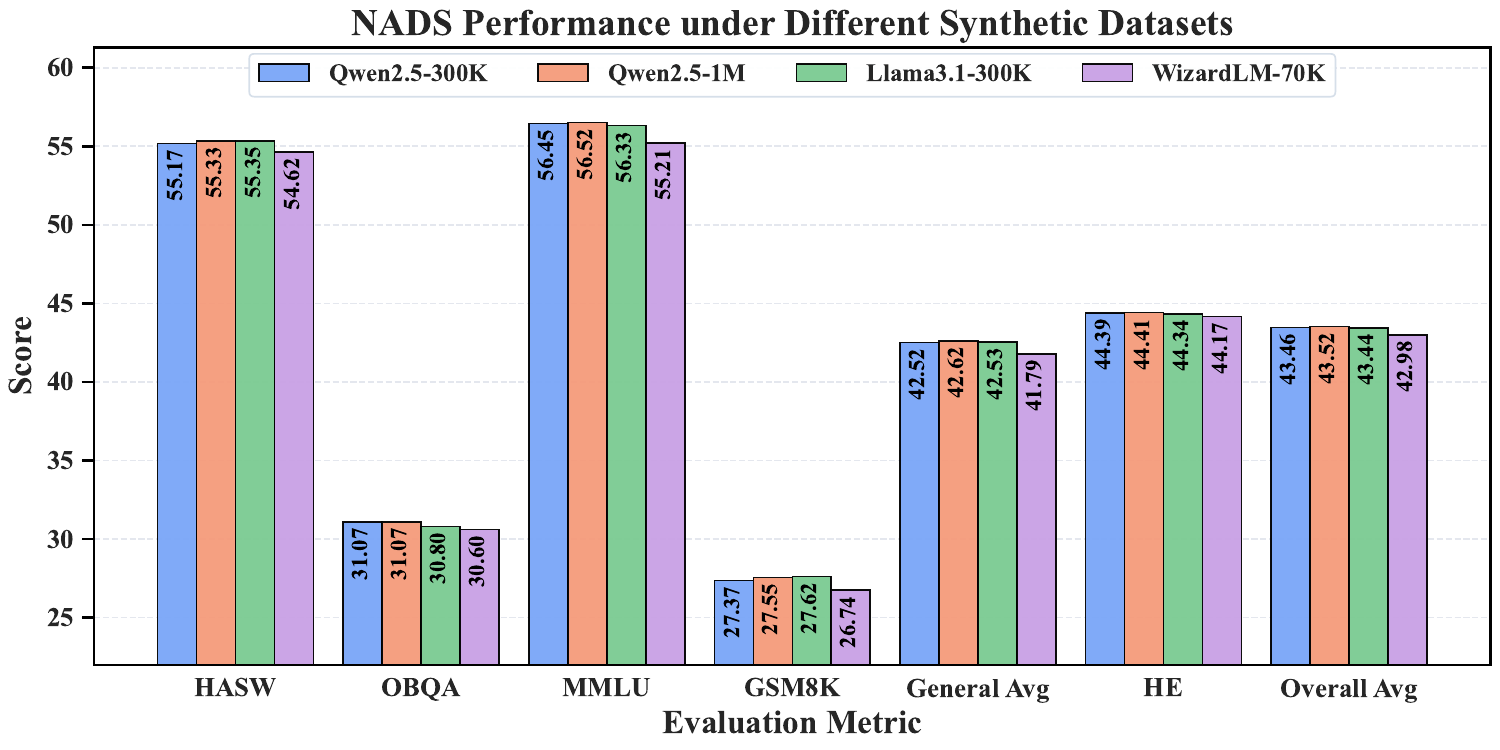}\vspace{-0.05in}
    \caption{NADS performance under different synthetic datasets across seven evaluation metrics on the Magicoder dataset with LLaMA3.2-3B. Each group of bars compares the results obtained with different synthetic data pools under the same setting.}
    \label{fig:different_synthetic}
\end{figure}

\begin{table}[t]
\centering
\footnotesize
\caption{Target performance and general capability of NADS under different constraint-set sizes $K$.}
\label{tab:K_analysis}
\setlength{\tabcolsep}{4mm}
\renewcommand{\arraystretch}{1.15}
\begin{tabular}{c|ccccc|c|c}
\toprule
\multirow{2}{*}{$K$} & \multicolumn{5}{c|}{\textbf{General Capability}} 
& \multicolumn{1}{c|}{\textbf{Target}} 
& \multicolumn{1}{c}{\textbf{Overall}} \\
\cmidrule(lr){2-6} \cmidrule(lr){7-7} \cmidrule(lr){8-8}
& \textbf{HASW} & \textbf{OBQA} & \textbf{MMLU} & \textbf{GSM8K} & \textbf{Avg.} & \textbf{HE} & \textbf{Avg.} \\
\midrule
1\%   & $\text{52.84}$ & $\text{28.73}$ & $\text{52.54}$ & $\text{22.87}$ & $\text{39.25}$ & $\text{44.93}$ & $\text{42.09}$ \\
3\%  & $\text{53.97}$ & $\text{30.13}$ & $\text{54.62}$ & $\text{25.40}$ & $\text{41.03}$ & $\text{44.50}$ & $\text{42.77}$ \\
5\%  & $\text{55.17}$ & $\text{31.07}$ & $\text{56.45}$ & $\text{27.37}$ & $\text{42.52}$ & $\text{44.39}$ & $\text{43.46}$ \\
10\%  & $\text{55.34}$ & $\text{31.13}$ & $\text{56.73}$ & $\text{27.42}$ & $\text{42.66}$ & $\text{44.38}$ & $\text{43.52}$ \\
15\%  & $\text{55.15}$ & $\text{30.87}$ & $\text{56.24}$ & $\text{27.14}$ & $\text{42.35}$ & $\text{43.86}$ & $\text{43.11}$ \\
\bottomrule
\end{tabular}
\end{table}

Table~\ref{tab:K_analysis} reports results under different constraint-set sizes. When $K=1\%$, the subset is too small to provide sufficient preservation support. Increasing $K$ to $3\%$ and $5\%$ markedly improves performance, showing that larger constraint sets provide stronger preservation signals. At $10\%$, performance continues to improve but with limited gain, suggesting saturation beyond a certain size. Thus, we use $5\%$ as the default budget to match the main replay setting while emphasizing efficiency. When $K$ further increases to $15\%$, performance declines, indicating that overly large constraint sets may introduce redundant or less informative samples and weaken selective construction. Overall, increasing $K$ helps when the constraint set is small, but the benefit gradually saturates and can reverse when excessive samples are included.

\subsection{Impact of Different Synthetic Datasets}
\label{synthetic}
In this section, we analyze the impact of synthetic datasets on NADS. NADS does not assume that the synthetic dataset used for constraint construction covers the model's general capabilities. Instead, it treats the synthetic dataset as a candidate source for constructing the constraint set.

We compare \textit{Magpie-Qwen2.5-Pro-300K-Filtered}~\citep{xu2025magpie}, \textit{Magpie-Qwen2.5-Pro-1M-v0.1}~\citep{xu2025magpie}, \textit{Magpie-Llama-3.1-Pro-300K-Filtered}~\citep{xu2025magpie}, and \textit{WizardLM-evol-instruct-70k}~\citep{xu2024wizardlm}, covering both synthetic datasets of different scales and datasets of the same scale generated by different models. For fair comparison, we fix the constraint-set budget to $15{,}000$ samples for all datasets, corresponding to $5\%$ of \textit{Magpie-Qwen2.5-Pro-300K-Filtered}. We then construct the constraint set from each synthetic dataset and evaluate its impact on NADS.

Fig.~\ref{fig:different_synthetic} reports results on seven evaluation metrics. Compared with \textit{Magpie-Qwen2.5-Pro-300K-Filtered}, \textit{Magpie-Qwen2.5-Pro-1M-v0.1} generally yields slightly better or more stable results, suggesting that a larger synthetic pool can better support constraint-set construction. \textit{Magpie-Llama-3.1-Pro-300K-Filtered} shows noticeable differences on some metrics, indicating that the generator model also affects constraint quality at a fixed data scale. In contrast, \textit{WizardLM-evol-instruct-70k} performs worse on several metrics, likely due to its smaller scale and limited candidate coverage. Overall, synthetic dataset choice affects NADS mainly through candidate-pool quality and coverage, while the overall variation remains moderate.

\begin{table}[t]
\centering
\footnotesize
\caption{Performance comparison on general capability and IFEval
benchmarks (\%). Results are obtained using a single random seed. Pre-trained and Standard FT are excluded from
the ranking. The \textcolor{red}{best results} are highlighted
\textcolor{red}{in red}, and the \textcolor{blue}{second-best results}
are highlighted \textcolor{blue}{in blue}.}
\setlength{\tabcolsep}{4mm}
\renewcommand{\arraystretch}{1.2}
\begin{tabular}{lcccccc}
\toprule
\textbf{Tasks}
& \textbf{Pre-trained}
& \textbf{Standard FT}
& \textbf{InsCL}
& \textbf{LER}
& \textbf{Mix-cd}
& \textbf{NADS} \\
\midrule
\multicolumn{7}{c}{\textbf{General Capability} $\uparrow$} \\
\midrule
ARC-Challenge
& $42.92$
& $44.37$
& $44.71$
& $45.05$
& \textcolor{blue}{$45.65$}
& \textcolor{red}{$46.67$} \\

ARC-Easy
& $74.58$
& $73.48$
& \textcolor{blue}{$75.34$}
& $75.13$
& $75.25$
& \textcolor{red}{$75.93$} \\

PIQA
& $76.28$
& $75.41$
& $76.50$
& $76.71$
& \textcolor{red}{$77.58$}
& \textcolor{blue}{$77.37$} \\

HellaSwag
& $55.83$
& $57.42$
& $57.05$
& $57.55$
& \textcolor{blue}{$58.10$}
& \textcolor{red}{$59.00$} \\

OpenBookQA
& $31.20$
& $33.20$
& $33.00$
& \textcolor{blue}{$33.80$}
& $33.60$
& \textcolor{red}{$34.60$} \\

TruthfulQA
& $39.13$
& $50.01$
& $48.39$
& $48.89$
& \textcolor{blue}{$49.50$}
& \textcolor{red}{$50.70$} \\

Winogrande
& $69.22$
& $66.30$
& $66.46$
& $66.85$
& \textcolor{red}{$68.03$}
& \textcolor{blue}{$67.88$} \\

MMLU
& $56.92$
& $53.42$
& $54.20$
& $54.60$
& \textcolor{blue}{$55.15$}
& \textcolor{red}{$55.90$} \\

GSM8K
& $27.90$
& $24.03$
& $24.11$
& $24.72$
& \textcolor{blue}{$25.47$}
& \textcolor{red}{$27.22$} \\

HumanEval
& $28.16$
& $26.22$
& $25.61$
& \textcolor{blue}{$27.44$}
& $26.83$
& \textcolor{red}{$29.27$} \\

Avg.
& $50.21$
& $50.39$
& $50.54$
& $51.07$
& \textcolor{blue}{$51.52$}
& \textcolor{red}{$52.45$} \\
\midrule
\multicolumn{7}{c}{\textbf{IFEval (Target)} $\uparrow$} \\
\midrule
Prompt (S)
& $21.81$
& $30.13$
& $32.35$
& $32.90$
& \textcolor{blue}{$33.64$}
& \textcolor{red}{$34.75$} \\

Inst (S)
& $34.77$
& $40.29$
& $41.85$
& $42.21$
& \textcolor{blue}{$42.93$}
& \textcolor{red}{$43.88$} \\

Prompt (L)
& $24.40$
& $31.79$
& $33.46$
& $34.01$
& \textcolor{blue}{$34.57$}
& \textcolor{red}{$35.49$} \\

Inst (L)
& $38.13$
& $42.81$
& $43.53$
& \textcolor{blue}{$44.48$}
& $44.12$
& \textcolor{red}{$45.44$} \\

Avg.
& $29.78$
& $36.26$
& $37.80$
& $38.40$
& \textcolor{blue}{$38.82$}
& \textcolor{red}{$39.89$} \\
\midrule
\multicolumn{7}{c}{\textbf{Overall} $\uparrow$} \\
\midrule
Avg.
& $40.00$
& $43.33$
& $44.17$
& $44.74$
& \textcolor{blue}{$45.17$}
& \textcolor{red}{$46.17$} \\
\bottomrule
\end{tabular}
\label{tab:nads_performance}\vspace{-0.1in}
\end{table}

\subsection{Results on Instruction Tuning}
\label{app:alpaca_gpt4_results}

We further evaluate all methods by fine-tuning LLaMA3.2-3B on Alpaca-GPT4~\cite{peng2023instruction} to assess broad instruction-tuning performance. Alpaca-GPT4 contains diverse instruction-following tasks, covering question answering, reasoning, writing, and formatting. For general capability, we report results on ARC-Challenge and ARC-Easy~\cite{clark2018think}, PIQA~\cite{bisk2020piqa}, Winogrande~\cite{sakaguchi2020winogrande}, HellaSwag~\cite{zellers2019hellaswag}, OpenBookQA~\cite{mihaylov2018can}, TruthfulQA~\cite{lin2022truthfulqa}, MMLU~\cite{hendrycks2021measuring}, GSM8K~\cite{cobbe2021training}, and HumanEval~\cite{chen2021evaluatinglargelanguagemodels}, and compute General Avg. as their arithmetic mean. Following recent instruction-tuning studies~\cite{li2025ruler,han2026makes,mirza-etal-2025-stratified}, we use IFEval~\cite{zhou2023instruction} and report prompt-level strict accuracy (PS), instruction-level strict accuracy (IS), prompt-level loose accuracy (PL), instruction-level loose accuracy (IL), and their arithmetic mean as IFEval Avg. Overall performance is computed as the average of General Avg. and IFEval Avg. Following the main experimental setup, we fine-tune on Alpaca-GPT4 for 3 epochs with a global batch size of 128 and a maximum sequence length of 4096. For NADS, the learning rate selected on Alpaca-GPT4 is $3 \times 10^{-5}$.

Table~\ref{tab:nads_performance} reports the Alpaca-GPT4 results. Standard FT improves IFEval Avg. from 29.78 to 36.26, while reducing performance on MMLU, GSM8K, and HumanEval relative to the pre-trained model. InsCL, LER, and Mix-cd further improve average instruction-following performance, with Mix-cd achieving the strongest baseline Overall score of 45.17. NADS achieves the highest General Avg., IFEval Avg., and Overall scores, reaching 52.45, 39.89, and 46.17, respectively. Among fine-tuned methods, NADS ranks first on all four IFEval metrics and eight of the ten general-capability benchmarks. These results demonstrate that NADS remains effective in broad instruction tuning, improving instruction following while preserving strong general capability.

\section{Comparison with Catastrophic Forgetting Mitigation Methods}
\label{cfmm}

In this subsection, we compare NADS with representative catastrophic-forgetting mitigation methods, including STM~\citep{wu2025mitigating}, NANOADAM~\citep{zhou2025pay}, FLOW~\citep{sanyalupweighting}, DavIR~\citep{zhou2025davir}, and TALR~\citep{lin2026sft}. Since this is a supplementary comparison beyond the main benchmark, we evaluate two datasets with two representative model backbones.

The full results on Magicoder and MetaMathQA are reported in Tables~\ref{tab:mitigation_magicoder} and \ref{tab:mitigation_MetaMathQA}. Across both datasets and backbones, NADS achieves the best Overall Avg., indicating a more favorable balance between target-task adaptation and general capability preservation. On Magicoder, standard fine-tuning substantially improves the target task but clearly degrades general-capability benchmarks, reflecting the typical adaptation--preservation conflict. Existing mitigation methods reduce this degradation to varying degrees, but their preservation signals are parameter-centric, regularization-based, or not explicitly tied to forgetting induced by the current target-task update. Thus, they often improve one side of the trade-off while leaving noticeable loss on the other. In contrast, NADS constructs preservation constraints from navigator-induced predictive drift, aligning the replay signal more directly with capabilities affected by the current adaptation trajectory. This yields the best Overall Avg. on both LLaMA3.2-3B and Gemma3-4B, with consistently strong General Capability Avg. and competitive target-task performance.

On MetaMathQA, the same trend appears under a different target capability. Standard fine-tuning attains strong GSM8K performance but noticeably weakens general capabilities, while existing mitigation baselines only partially recover this balance. NADS achieves the best Overall Avg. and General Capability Avg. on both backbones, showing that the selected synthetic constraints preserve broad capabilities even when the target task shifts to mathematical reasoning. Meanwhile, NADS maintains highly competitive GSM8K performance, suggesting that the preservation constraints do not simply over-regularize the model toward the pretrained state. Overall, NADS improves the adaptation--preservation trade-off across both code and math fine-tuning settings, rather than merely increasing target performance or retention in isolation.

\newtcbox{\venuebox}{on line,
  boxrule=0pt,
  colback=red!15,
  colframe=gray!15,
  arc=1.5pt,
  left=2pt,right=2pt,top=1pt,bottom=1pt,
  boxsep=0pt}
  
\newcommand{\venuetag}[1]{{\scriptsize\venuebox{\textcolor{black!70}{#1}}}}

\newcommand{\methodvenue}[2]{%
  \makebox[1.45cm][l]{#1}%
  \hspace{0.1cm}%
  \makebox[1.25cm][c]{\venuetag{#2}}%
}

\begin{table}[H]
\centering
\scriptsize
\renewcommand{\arraystretch}{1.2}
\setlength{\tabcolsep}{1.8mm}
\caption{Comparison with catastrophic forgetting mitigation methods on Magicoder dataset (\%). The \textcolor{red}{best results} are highlighted \textcolor{red}{in red}, and the \textcolor{blue}{second-best} results are highlighted \textcolor{blue}{in blue}.}
\label{tab:mitigation_magicoder}


\begin{tabular}{lccccc|c|c}
\toprule
\multirow{2}{*}{\textbf{Method}} 
& \multicolumn{5}{c|}{\textbf{General Capability} $\uparrow$} 
& \multicolumn{1}{c|}{\textbf{Target} $\uparrow$} 
& \multicolumn{1}{c}{\textbf{Overall} $\uparrow$} \\
\cmidrule(lr){2-6} \cmidrule(lr){7-7} \cmidrule(lr){8-8}
& \textbf{HASW} & \textbf{OBQA} & \textbf{MMLU} & \textbf{GSM8K} & \textbf{Avg.} & \textbf{HE} & \textbf{Avg.} \\

\modelrow{LLaMA3.2-3B}

Pre-trained & $\text{55.83}_{\pm \text{0.00}}$ & $\text{31.20}_{\pm \text{0.00}}$ & $\text{56.92}_{\pm \text{0.00}}$ & $\text{27.90}_{\pm \text{0.00}}$ & $\text{42.96}_{\pm \text{0.00}}$ & $\text{28.16}_{\pm \text{0.00}}$ & $\text{35.56}_{\pm \text{0.00}}$ \\
Standard FT & $\text{52.12}_{\pm \text{0.52}}$ & $\text{28.40}_{\pm \text{0.69}}$ & $\text{48.61}_{\pm \text{0.40}}$ & $\text{20.75}_{\pm \text{0.74}}$ & $\text{37.47}_{\pm \text{0.30}}$ & $\text{45.13}_{\pm \text{0.32}}$ & $\text{41.30}_{\pm \text{0.22}}$ \\
\methodvenue{LoRA}{ICLR 22} & $\text{53.38}_{\pm \text{0.55}}$ & $\text{29.47}_{\pm \text{0.64}}$ & $\text{51.98}_{\pm \text{0.42}}$ & $\text{21.18}_{\pm \text{0.79}}$ & $\text{39.00}_{\pm \text{0.31}}$ & $\text{41.92}_{\pm \text{0.30}}$ & $\text{40.46}_{\pm \text{0.21}}$ \\

\methodvenue{STM}{NIPS 25}            & $\text{53.02}_{\pm \text{0.49}}$ & $\text{29.67}_{\pm \text{0.61}}$ & $\text{53.24}_{\pm \text{0.41}}$ & $\text{23.81}_{\pm \text{0.75}}$ & $\text{39.94}_{\pm \text{0.29}}$ & $\text{42.04}_{\pm \text{0.82}}$ & $\text{40.99}_{\pm \text{0.44}}$ \\
\methodvenue{NANOADAM}{NIPS 25}        & $\text{53.89}_{\pm \text{0.49}}$ & $\textcolor{blue}{\text{30.20}_{\pm \text{0.60}}}$ & $\text{52.83}_{\pm \text{0.40}}$ & $\text{24.44}_{\pm \text{0.62}}$ & $\text{40.34}_{\pm \text{0.26}}$ & $\text{43.31}_{\pm \text{0.74}}$ & $\text{41.83}_{\pm \text{0.39}}$ \\
\methodvenue{FLOW}{ICML 25}            & $\text{54.63}_{\pm \text{0.50}}$ & $\text{30.07}_{\pm \text{0.61}}$ & $\textcolor{blue}{\text{55.83}_{\pm \text{0.42}}}$ & $\text{25.12}_{\pm \text{0.63}}$ & $\text{41.41}_{\pm \text{0.27}}$ & $\text{42.34}_{\pm \text{0.74}}$ & $\text{41.88}_{\pm \text{0.39}}$ \\
\methodvenue{DavIR}{ACL 25}            & $\text{53.27}_{\pm \text{0.49}}$ & $\text{29.20}_{\pm \text{0.60}}$ & $\text{53.17}_{\pm \text{0.40}}$ & $\text{25.70}_{\pm \text{0.76}}$ & $\text{40.34}_{\pm \text{0.29}}$ & $\text{43.52}_{\pm \text{0.77}}$ & $\text{41.93}_{\pm \text{0.41}}$ \\
\methodvenue{TALR}{ICLR 26}            & $\textcolor{blue}{\text{54.84}_{\pm \text{0.50}}}$ & $\text{30.13}_{\pm \text{0.64}}$ & $\text{54.01}_{\pm \text{0.41}}$ & $\textcolor{blue}{\text{26.79}_{\pm \text{0.74}}}$ & $\textcolor{blue}{\text{41.44}_{\pm \text{0.29}}}$ & $\textcolor{blue}{\text{43.54}_{\pm \text{0.75}}}$ & $\textcolor{blue}{\text{42.49}_{\pm \text{0.40}}}$ \\

\rowcolor{monaPurple}
NADS (Ours) & $\textcolor{red}{\text{55.17}_{\pm \text{0.49}}}$ & $\textcolor{red}{\text{31.07}_{\pm \text{0.58}}}$ & $\textcolor{red}{\text{56.45}_{\pm \text{0.41}}}$ & $\textcolor{red}{\text{27.37}_{\pm \text{0.67}}}$ & $\textcolor{red}{\text{42.52}_{\pm \text{0.27}}}$ & $\textcolor{red}{\text{44.39}_{\pm \text{0.40}}}$ & $\textcolor{red}{\text{43.46}_{\pm \text{0.24}}}$ \\

\modelrow{Gemma3-4B}

Pre-trained & $\text{57.41}_{\pm \text{0.00}}$ & $\text{34.60}_{\pm \text{0.00}}$ & $\text{60.24}_{\pm \text{0.00}}$ & $\text{36.69}_{\pm \text{0.00}}$ & $\text{47.24}_{\pm \text{0.00}}$ & $\text{35.77}_{\pm \text{0.00}}$ & $\text{41.51}_{\pm \text{0.00}}$ \\
Standard FT & $\text{53.06}_{\pm \text{0.49}}$ & $\text{31.27}_{\pm \text{0.70}}$ & $\text{53.21}_{\pm \text{0.40}}$ & $\text{23.78}_{\pm \text{0.68}}$ & $\text{40.33}_{\pm \text{0.29}}$ & $\text{58.17}_{\pm \text{0.89}}$ & $\text{49.25}_{\pm \text{0.47}}$ \\
\methodvenue{LoRA}{ICLR 22} & $\text{55.81}_{\pm \text{0.50}}$ & $\text{31.80}_{\pm \text{0.60}}$ & $\text{56.74}_{\pm \text{0.39}}$ & $\text{31.61}_{\pm \text{0.82}}$ & $\text{43.99}_{\pm \text{0.30}}$ & $\text{54.69}_{\pm \text{0.91}}$ & $\text{49.34}_{\pm \text{0.48}}$ \\

\methodvenue{STM}{NIPS 25}            & $\text{56.22}_{\pm \text{0.49}}$ & $\text{32.00}_{\pm \text{0.60}}$ & $\text{58.69}_{\pm \text{0.39}}$ & $\text{33.13}_{\pm \text{0.85}}$ & $\text{45.01}_{\pm \text{0.30}}$ & $\text{55.42}_{\pm \text{0.81}}$ & $\text{50.22}_{\pm \text{0.43}}$ \\
\methodvenue{NANOADAM}{NIPS 25}        & $\text{56.48}_{\pm \text{0.49}}$ & $\text{32.07}_{\pm \text{0.64}}$ & $\text{58.91}_{\pm \text{0.39}}$ & $\text{33.06}_{\pm \text{0.84}}$ & $\text{45.13}_{\pm \text{0.31}}$ & $\text{56.27}_{\pm \text{0.90}}$ & $\text{50.70}_{\pm \text{0.48}}$ \\
\methodvenue{FLOW}{ICML 25}            & $\text{55.23}_{\pm \text{0.50}}$ & $\textcolor{blue}{\text{32.40}_{\pm \text{0.60}}}$ & $\text{58.17}_{\pm \text{0.40}}$ & $\text{34.17}_{\pm \text{0.82}}$ & $\text{44.99}_{\pm \text{0.30}}$ & $\text{55.86}_{\pm \text{0.84}}$ & $\text{50.43}_{\pm \text{0.45}}$ \\
\methodvenue{DavIR}{ACL 25}            & $\text{54.55}_{\pm \text{0.49}}$ & $\text{31.60}_{\pm \text{0.69}}$ & $\text{57.12}_{\pm \text{0.39}}$ & $\text{30.60}_{\pm \text{0.86}}$ & $\text{43.47}_{\pm \text{0.31}}$ & $\textcolor{blue}{\text{57.46}_{\pm \text{0.72}}}$ & $\text{50.47}_{\pm \text{0.39}}$ \\
\methodvenue{TALR}{ICLR 26}            & $\textcolor{blue}{\text{56.62}_{\pm \text{0.49}}}$ & $\text{32.27}_{\pm \text{0.64}}$ & $\textcolor{red}{\text{59.79}_{\pm \text{0.39}}}$ & $\textcolor{blue}{\text{35.08}_{\pm \text{0.84}}}$ & $\textcolor{blue}{\text{45.94}_{\pm \text{0.31}}}$ & $\text{56.08}_{\pm \text{0.91}}$ & $\textcolor{blue}{\text{51.01}_{\pm \text{0.48}}}$ \\

\rowcolor{monaPurple}
NADS (Ours) & $\textcolor{red}{\text{56.83}_{\pm \text{0.50}}}$ & $\textcolor{red}{\text{33.13}_{\pm \text{0.61}}}$ & $\textcolor{blue}{\text{59.62}_{\pm \text{0.40}}}$ & $\textcolor{red}{\text{35.99}_{\pm \text{0.74}}}$ & $\textcolor{red}{\text{46.39}_{\pm \text{0.29}}}$ & $\textcolor{red}{\text{57.88}_{\pm \text{0.90}}}$ & $\textcolor{red}{\text{52.14}_{\pm \text{0.47}}}$ \\

\bottomrule
\end{tabular}
\par\vspace{16pt}
\centering
\scriptsize
\renewcommand{\arraystretch}{1.2}
\setlength{\tabcolsep}{1.8mm}
\caption{Comparison with catastrophic forgetting mitigation methods on MetaMathQA dataset (\%). The \textcolor{red}{best results} are highlighted \textcolor{red}{in red}, and the \textcolor{blue}{second-best} results are highlighted \textcolor{blue}{in blue}.}
\label{tab:mitigation_MetaMathQA}


\begin{tabular}{lccccc|c|c}
\toprule
\multirow{2}{*}{\textbf{Method}} 
& \multicolumn{5}{c|}{\textbf{General Capability} $\uparrow$} 
& \multicolumn{1}{c|}{\textbf{Target} $\uparrow$} 
& \multicolumn{1}{c}{\textbf{Overall} $\uparrow$} \\
\cmidrule(lr){2-6} \cmidrule(lr){7-7} \cmidrule(lr){8-8}
& \textbf{HASW} & \textbf{OBQA} & \textbf{MMLU} & \textbf{HE} & \textbf{Avg.} & \textbf{GSM8K} & \textbf{Avg.} \\

\modelrow{LLaMA3.2-3B}

Pre-trained & $\text{55.83}_{\pm \text{0.00}}$ & $\text{31.20}_{\pm \text{0.00}}$ & $\text{56.92}_{\pm \text{0.00}}$ & $\text{28.16}_{\pm \text{0.00}}$ & $\text{43.03}_{\pm \text{0.00}}$ & $\text{27.90}_{\pm \text{0.00}}$ & $\text{35.47}_{\pm \text{0.00}}$ \\
Standard FT & $\text{50.02}_{\pm \text{0.50}}$ & $\text{28.40}_{\pm \text{0.53}}$ & $\text{48.49}_{\pm \text{0.41}}$ & $\text{15.27}_{\pm \text{0.49}}$ & $\text{35.55}_{\pm \text{0.24}}$ & $\text{65.93}_{\pm \text{0.81}}$ & $\text{50.74}_{\pm \text{0.42}}$ \\
\methodvenue{LoRA}{ICLR 22}        & $\text{52.06}_{\pm \text{0.50}}$ & $\text{28.60}_{\pm \text{0.53}}$ & $\text{52.34}_{\pm \text{0.40}}$ & $\text{25.31}_{\pm \text{0.52}}$ & $\text{39.58}_{\pm \text{0.25}}$ & $\text{61.03}_{\pm \text{0.83}}$ & $\text{50.31}_{\pm \text{0.43}}$ \\

\methodvenue{STM}{NIPS 25}            & $\text{53.07}_{\pm \text{0.49}}$ & $\text{29.60}_{\pm \text{0.60}}$ & $\text{53.65}_{\pm \text{0.41}}$ & $\text{26.83}_{\pm \text{0.69}}$ & $\text{40.79}_{\pm \text{0.28}}$ & $\text{63.51}_{\pm \text{0.84}}$ & $\text{52.15}_{\pm \text{0.44}}$ \\
\methodvenue{NANOADAM}{NIPS 25}        & $\text{53.26}_{\pm \text{0.50}}$ & $\text{30.13}_{\pm \text{0.64}}$ & $\text{54.92}_{\pm \text{0.39}}$ & $\text{26.43}_{\pm \text{0.71}}$ & $\text{41.19}_{\pm \text{0.29}}$ & $\text{64.42}_{\pm \text{0.78}}$ & $\textcolor{blue}{\text{52.81}_{\pm \text{0.42}}}$ \\
\methodvenue{FLOW}{ICML 25}            & $\text{53.41}_{\pm \text{0.49}}$ & $\textcolor{blue}{\text{30.33}_{\pm \text{0.61}}}$ & $\text{55.86}_{\pm \text{0.42}}$ & $\text{26.97}_{\pm \text{0.75}}$ & $\text{41.64}_{\pm \text{0.29}}$ & $\text{63.74}_{\pm \text{0.74}}$ & $\text{52.69}_{\pm \text{0.40}}$ \\
\methodvenue{DavIR}{ACL 25}            & $\text{52.37}_{\pm \text{0.49}}$ & $\text{29.20}_{\pm \text{0.60}}$ & $\text{53.72}_{\pm \text{0.39}}$ & $\text{25.83}_{\pm \text{0.82}}$ & $\text{40.28}_{\pm \text{0.30}}$ & $\textcolor{red}{\text{65.25}_{\pm \text{0.81}}}$ & $\text{52.77}_{\pm \text{0.43}}$ \\
\methodvenue{TALR}{ICLR 26}            & $\textcolor{blue}{\text{53.52}_{\pm \text{0.49}}}$ & $\text{30.07}_{\pm \text{0.64}}$ & $\textcolor{blue}{\text{55.94}_{\pm \text{0.39}}}$ & $\textcolor{blue}{\text{27.32}_{\pm \text{0.74}}}$ & $\textcolor{blue}{\text{41.71}_{\pm \text{0.29}}}$ & $\text{63.68}_{\pm \text{0.76}}$ & $\text{52.70}_{\pm \text{0.41}}$ \\

\rowcolor{monaPurple}
NADS (Ours) & $\textcolor{red}{\text{54.86}_{\pm \text{0.50}}}$ & $\textcolor{red}{\text{30.80}_{\pm \text{0.60}}}$ & $\textcolor{red}{\text{56.23}_{\pm \text{0.41}}}$ & $\textcolor{red}{\text{28.21}_{\pm \text{0.42}}}$ & $\textcolor{red}{\text{42.53}_{\pm \text{0.24}}}$ & $\textcolor{blue}{\text{65.18}_{\pm \text{0.83}}}$ & $\textcolor{red}{\text{53.86}_{\pm \text{0.43}}}$ \\

\modelrow{Gemma3-4B}

Pre-trained & $\text{57.41}_{\pm \text{0.00}}$ & $\text{34.60}_{\pm \text{0.00}}$ & $\text{60.24}_{\pm \text{0.00}}$ & $\text{35.77}_{\pm \text{0.00}}$ & $\text{47.01}_{\pm \text{0.00}}$ & $\text{36.69}_{\pm \text{0.00}}$ & $\text{41.85}_{\pm \text{0.00}}$ \\
Standard FT & $\text{51.42}_{\pm \text{0.50}}$ & $\text{30.20}_{\pm \text{0.53}}$ & $\text{53.01}_{\pm \text{0.40}}$ & $\text{29.62}_{\pm \text{0.43}}$ & $\text{41.06}_{\pm \text{0.24}}$ & $\text{67.83}_{\pm \text{0.79}}$ & $\text{54.45}_{\pm \text{0.41}}$ \\
\methodvenue{LoRA}{ICLR 22}        & $\text{54.34}_{\pm \text{0.49}}$ & $\text{32.13}_{\pm \text{0.64}}$ & $\text{55.84}_{\pm \text{0.39}}$ & $\text{30.36}_{\pm \text{0.74}}$ & $\text{43.17}_{\pm \text{0.29}}$ & $\text{63.03}_{\pm \text{0.78}}$ & $\text{53.10}_{\pm \text{0.42}}$ \\

\methodvenue{STM}{NIPS 25}            & $\text{54.83}_{\pm \text{0.51}}$ & $\text{32.53}_{\pm \text{0.64}}$ & $\text{58.15}_{\pm \text{0.40}}$ & $\text{33.06}_{\pm \text{0.82}}$ & $\text{44.64}_{\pm \text{0.31}}$ & $\text{64.85}_{\pm \text{0.81}}$ & $\text{54.75}_{\pm \text{0.43}}$ \\
\methodvenue{NANOADAM}{NIPS 25}        & $\textcolor{blue}{\text{55.19}_{\pm \text{0.50}}}$ & $\textcolor{blue}{\text{33.13}_{\pm \text{0.64}}}$ & $\text{57.94}_{\pm \text{0.40}}$ & $\text{32.93}_{\pm \text{0.69}}$ & $\text{44.80}_{\pm \text{0.28}}$ & $\text{65.73}_{\pm \text{0.79}}$ & $\textcolor{blue}{\text{55.27}_{\pm \text{0.42}}}$ \\
\methodvenue{FLOW}{ICML 25}            & $\text{54.92}_{\pm \text{0.49}}$ & $\text{32.67}_{\pm \text{0.61}}$ & $\textcolor{blue}{\text{58.27}_{\pm \text{0.41}}}$ & $\text{33.24}_{\pm \text{0.85}}$ & $\text{44.78}_{\pm \text{0.31}}$ & $\text{64.77}_{\pm \text{0.85}}$ & $\text{54.78}_{\pm \text{0.45}}$ \\
\methodvenue{DavIR}{ACL 25}            & $\text{54.83}_{\pm \text{0.49}}$ & $\text{31.87}_{\pm \text{0.61}}$ & $\text{56.67}_{\pm \text{0.41}}$ & $\text{31.41}_{\pm \text{0.84}}$ & $\text{43.70}_{\pm \text{0.30}}$ & $\textcolor{blue}{\text{66.29}_{\pm \text{0.82}}}$ & $\text{55.00}_{\pm \text{0.44}}$ \\
\methodvenue{TALR}{ICLR 26}            & $\text{55.03}_{\pm \text{0.50}}$ & $\text{32.87}_{\pm \text{0.64}}$ & $\text{58.16}_{\pm \text{0.39}}$ & $\textcolor{blue}{\text{33.28}_{\pm \text{0.89}}}$ & $\textcolor{blue}{\text{44.84}_{\pm \text{0.32}}}$ & $\text{64.87}_{\pm \text{0.77}}$ & $\text{54.86}_{\pm \text{0.42}}$ \\

\rowcolor{monaPurple}
NADS (Ours) & $\textcolor{red}{\text{55.46}_{\pm \text{0.50}}}$ & $\textcolor{red}{\text{33.47}_{\pm \text{0.61}}}$ & $\textcolor{red}{\text{59.31}_{\pm \text{0.40}}}$ & $\textcolor{red}{\text{34.75}_{\pm \text{0.89}}}$ & $\textcolor{red}{\text{45.75}_{\pm \text{0.31}}}$ & $\textcolor{red}{\text{66.52}_{\pm \text{0.77}}}$ & $\textcolor{red}{\text{56.14}_{\pm \text{0.42}}}$ \\

\bottomrule
\end{tabular}
\end{table}

\section{Limitations}
\label{Limitations}
A main limitation of NADS lies in navigator construction. We obtain the navigator by following the target fine-tuning recipe without preservation regularization, so its deviation from the pretrained model exposes recipe-induced drift. While simple and aligned with the adaptation process, this design makes the selection signal sensitive to the navigator checkpoint and training dynamics, and introduces an additional pre-selection cost before final fine-tuning. A navigator stopped too early may provide weak drift signals, whereas one trained too long may yield noisy or biased utility estimates. Although validation-based checkpoint selection and the cost analysis in Appendix~\ref{Computational} support the practicality of this design, more adaptive and lower-cost navigator construction may further improve robustness across tasks and fine-tuning recipes.

\section*{Broader Impacts}

NADS aims to make LLM fine-tuning more stable by selecting synthetic constraints that help preserve pretrained behavior during new-task adaptation. This may reduce capability degradation and lower the cost of safer model adaptation. However, preserved pretrained behavior and synthetic candidate pools may still contain biases, factual errors, or unsafe content. Deployments in sensitive domains should therefore include task-specific safety evaluation, data-quality review, and misuse-risk assessment. Our experiments use public datasets, benchmarks, and base models, and do not collect user data or involve human-subject studies.

\section*{Reproducibility}

We will release code, configurations, prompt templates, and scripts for reproducing the main results. All datasets, benchmarks, synthetic candidate pools, and base models used in this work are publicly available. The repository will document preprocessing, split construction, navigator training, constraint selection, final fine-tuning, and evaluation commands.

\section*{Licenses and Assets}

We use publicly available datasets, benchmarks, synthetic data sources, and base models under their respective licenses and terms of use. NADS does not introduce a new dataset. Released assets will include code, configurations, scripts, and prompt templates, and will not redistribute third-party datasets or model weights unless permitted by their licenses.



\end{document}